\documentclass[mnsc，nonblindrev]{informs3} 
\OneAndAHalfSpacedXI 
\usepackage{endnotes}
\usepackage{hyperref}
\usepackage{adjustbox}  
\usepackage{array}      
\usepackage{float}      
\usepackage{caption}    
\usepackage[ruled]{algorithm}
\usepackage{algorithmicx,algpseudocode,amsfonts}
\usepackage{graphicx,hyperref,comment,appendix}
\usepackage[capitalize,noabbrev]{cleveref}
\usepackage{tikz, pgfplots}
\usepackage{subfigure,caption,microtype,booktabs} 

\usepackage{amsthm,amsmath,amssymb,soul}
\usepackage{comment,enumitem}
\usepackage{subcaption}
\setlist{leftmargin=*}
\usepackage{booktabs}   
\usepackage{multirow}  
\usepackage{graphicx}
\usepackage{pdflscape}
\usepackage{float}   
\usepackage{graphicx}

\usepackage{subcaption}
\RequirePackage{tgtermes}
\RequirePackage{newtxtext}
\RequirePackage{newtxmath}
\RequirePackage{bm}
\RequirePackage{endnotes}
\usepackage{amssymb}
\usepackage{pifont}
\usepackage{authblk}
\usepackage{csquotes}
\usepackage{xspace}
\OneAndAHalfSpacedXII 
\usepackage{algorithm}
\usepackage{algpseudocode}
\usepackage{tikz}
\usepackage{hyperref}
\usepackage[capitalize]{cleveref}
\usepackage[labelfont=bf]{caption} 
\usepackage[most]{tcolorbox}
\usepackage{longtable}
\usepackage{listings}
\usepackage[dvipsnames]{xcolor}
\usepackage{multirow}

\newtheorem{remark}{Remark}

\newcommand{\blue}{\color{blue}}
\newcommand{\green}{\color{Green}}

\newcommand{\agent}{\texttt{LEAN-LLM-OPT}\xspace}
\newcommand{\agentabbrv}{\texttt{LEAN}\xspace}

\newcommand{\refdata}{\texttt{Ref-Data}\xspace}
\newcommand{\testdata}{\texttt{Large-Scale-OR}\xspace}
\newcommand{\sindata}{\texttt{Air-NRM}\xspace}
\newcommand{\sindataca}{\texttt{Air-NRM-CA}\xspace}

\newcommand{\sindatanp}{\texttt{Air-NRM-NP}\xspace}

\newcommand{\demo}{\text{demo}}

\renewcommand{\arraystretch}{1.2}
\definecolor{boxblue}{HTML}{CC7C71}
\definecolor{boxgrey}{HTML}{6F6F6F}
\definecolor{boxagent2}{HTML}{3E608D}
\newcolumntype{Y}{>{\centering\arraybackslash}X} 

\definecolor{codegreen}{rgb}{0,0.6,0}
\definecolor{codegray}{rgb}{0.5,0.5,0.5}
\definecolor{codepurple}{rgb}{0.58,0,0.82}
\definecolor{backcolour}{rgb}{0.95,0.95,0.92}

\lstdefinestyle{mystyle}{
    backgroundcolor=\color{backcolour},   
    commentstyle=\color{codegreen},       
    keywordstyle=\color{magenta},         
    numberstyle=\tiny\color{codegray},    
    stringstyle=\color{codepurple},       
    basicstyle=\ttfamily\footnotesize,     
    breakatwhitespace=false,              
    breaklines=true,                      
    captionpos=b,                         
    keepspaces=true,                      
    numbers=left,                         
    numbersep=5pt,                        
    showspaces=false,                     
    showstringspaces=false,               
    showtabs=false,                       
    tabsize=2,                            
    language=Python                       
}
\usepackage{natbib}
\bibpunct[, ]{(}{)}{,}{a}{}{,}%
\def\bibfont{\small}%
\usepackage{bm,comment}

\ECRepeatTheorems
\EquationsNumberedThrough    
\allowdisplaybreaks
\begin{document}
\RUNTITLE{LLM for Large-Scale Optimization Model Auto-Formulation}
\TITLE{\Large Large-Scale Optimization Model Auto-Formulation: Harnessing LLM Flexibility via Structured Workflow}

\ARTICLEAUTHORS{\footnotesize
  	\AUTHOR{\textbf{Kuo Liang$^1$, Yuhang Lu $^{2}$, Jianming Mao$^3$, Shuyi Sun$^4$, Chunwei Yang$^5$, Congcong Zeng$^7$\\ Xiao Jin$^{6,9}$, Hanzhang Qin$^{2,6,7}$, Ruihao Zhu$^8$, Chung-Piaw Teo$^{6,7,9}$}}
 	\AFF{
  	$^1$ Antai College of Economics and Management, Shanghai Jiao Tong University\\
  	$^2$ Department of Industrial Systems Engineering and Management, National University of Singapore\\         $^3$ Department of Analytics \& Operations, Imperial Business School \\
    $^4$ Department of Decision Analytics \& Operations, City University of Hong Kong \\
	$^5$ Department of System Engineering, City University of Hong Kong\\
  	$^6$ Institute of Operations Research and Analytics, National University of Singapore\\
         $^7$ Mitch Daniels School of Business, Purdue University \\
  	$^8$ SC Johnson College of Business, Cornell University \\
         $^9$ SIA-NUS Digital Aviation Corporate Lab}
  	\AFF{\textit{
 	cora.liang1116@gmail.com\quad
     adamleo@nus.edu.sg\quad
  	j.mao25@imperial.ac.uk\quad
  	shuyisun6-c@my.cityu.edu
  	chunwyang2-c@my.cityu.edu.hk\quad
         zeng365@purdue.edu\quad
         xiao.j@nus.edu.sg\quad
  	hzgin@nus.edu.sg\quad
  ruihao.zhu@cornell.edu\quad bizteocp@nus.edu.sg }}
 }

\ABSTRACT{%
Large-scale optimization is a key backbone of modern business decision-making. However, building these models is often labor-intensive and time-consuming. We address this by proposing \agent, a \underline{L}ightw\underline{E}ight \underline{A}ge\underline{N}tic workflow construction framework for LLM-assisted large-scale \underline{OPT}imization model auto-formulation. \agent takes as input a problem description together with associated datasets and orchestrates a team of LLM agents to produce an optimization formulation. Specifically, upon receiving a query, two upstream LLM agents dynamically construct a workflow that specifies, step-by-step, how optimization models for similar problems can be formulated; A downstream LLM agent then follows this workflow to generate the final output. The agentic workflow leverages common modeling practices to structure the modeling process into a sequence of sub-tasks, offloading mechanical data-handling operations to auxiliary tools. This reduces the LLM’s burden in planning and data handling, allowing us to exploit its flexibility to address unstructured components. Extensive simulations show that \agent, instantiated with GPT-4.1 and the open source gpt-oss-20B, achieves strong performance on large-scale optimization modeling tasks. In addition, in a Singapore Airlines choice-based revenue management use case, \agent demonstrates practical value by achieving leading performance across a range of scenarios. Along the way, we introduce \testdata and \sindata, the first comprehensive benchmarks for large-scale optimization auto-formulation. The code and data of this work is available at
\href{https://github.com/CoraLiang01/lean-llm-opt}{https://github.com/CoraLiang01/lean-llm-opt}. 
}%

\maketitle

\section{Introduction}\label{sec:Intro}
As modern businesses grow in scale and complexity, organizations increasingly rely on advanced optimization techniques to enhance decision-making and operational efficiency. From supply chain networks to financial planning and resource allocation, many business sectors face the challenge of extracting information from data and formulating intricate optimization problems on top of it \citep{chen2021item,kangJDComImproves2022,dang2021network,deng2023alibaba}. As observed in many cases, in order to apply these techniques, companies and organizations usually rely on highly skilled experts to understand the problems, process data, and formulate the optimization models. Currently, the above works typically require significant manual effort and can be costly, making it a major bottleneck in fully leveraging the power of these techniques. Therefore, researchers start to explore automated modeling of optimization problems in an end-to-end manner, i.e., from the natural language problem descriptions and data to optimization formulations and the corresponding programming codes \citep{ramamonjison2022augmenting,ramamonjison2023nl4opt}. 

Recently, advancements in Large Language Models (LLMs) open a new avenue for automated optimization modeling \citep{levi2025democratizing}. Built on LLMs' remarkable ability in understanding and processing unstructured information and natural languages, several studies have been focusing on utilizing LLMs to formulate and/or solve optimization problems based on text descriptions. This facilitates the application of optimization techniques by automating expert-level modeling, allowing more non-experts to handle tricky optimization problems through interfacing with LLMs. As initial attempts, several works examine how to prompt pretrained LLMs to formulate and/or solve optimization problems directly \citep{xiao2023chain,ahmaditeshnizi2024OptiMUS,li2023synthesizing,chen2024diagnosing,wasserkrug2025decision}. Perhaps not surprisingly, using prompting techniques only without any modification to the LLMs may lead to unsatisfactory accuracy. Hence, researchers turn to fine-tuning approaches, which modify the parameters of open source LLMs based on labeled datasets, to improve their performance \citep{ma2024llamoco,huang2024mamo,huang_orlm_2024,yang2024optibench,jiang2024llmopt,astorga2024autoformulation}. 
 
While fine-tuning approaches significantly improve LLMs' modeling accuracy, prior works primarily focus on small-scale inputs and outputs, where the data can be directly embedded into the problem description and the output only has a few variables. Meanwhile, a practical optimization  model typically involves a large number of variables, and in order to correctly define its objective and constraints, the input usually has a large volume of data that has to be stored separately in  datasets. However, when turning to large-scale problems\footnote{We refer to problem instances whose data has to be stored separately in datasets as large-scale. For the corresponding model, we define the scale based on the number of variables needed (small-size: $<20$ variables, medium-size: $20-99$ variables, large-size: $\geq 100$ variables). In this work, we focus mainly on large-scale inputs but allow the scale of the output to vary. See Section \ref{sec:general} for a summary statistics for the scale of the inputs of our benchmark.}, the modeling accuracies of state-of-the-art fine-tuned models and closed-source commercial models can quickly deteriorate (see, e.g., Table \ref{tab:by-family-transposed} of Section \ref{sec:numerical}). This thus raises a critical question,
\emph{can we design LLM-based methods for efficient large-scale optimization modeling from natural language inputs?} For this, several major challenges persist:
\begin{itemize}
    \item \textbf{Challenge 1. Input Heterogeneity yet Inadequate Training Resource:} In large-scale optimization modeling, LLMs must navigate long inputs that combine diverse problem descriptions with associated datasets, making standardization and automation particularly challenging. A tempting solution for the above is to continue with fine-tuning approaches so that LLMs become even more flexible and adept for a wider range of problems. While this can work for small-scale problems, generalizing the training pipelines to large-scale problems can be unrealistic. This is because large-scale optimization problems can be much more versatile. For example, by altering a small portion of the problem description or the datasets (e.g., split a dataset into multiples, add or delete some rows and/or columns in the datasets), one may get a totally different problem instance and a new labeled may also be needed. Consequently, one may need to construct a prohibitively large set of inputs and labeled outputs to apply fine-tuning approaches. 
      \begin{figure}
    \centering
    \includegraphics[width=0.9\linewidth]{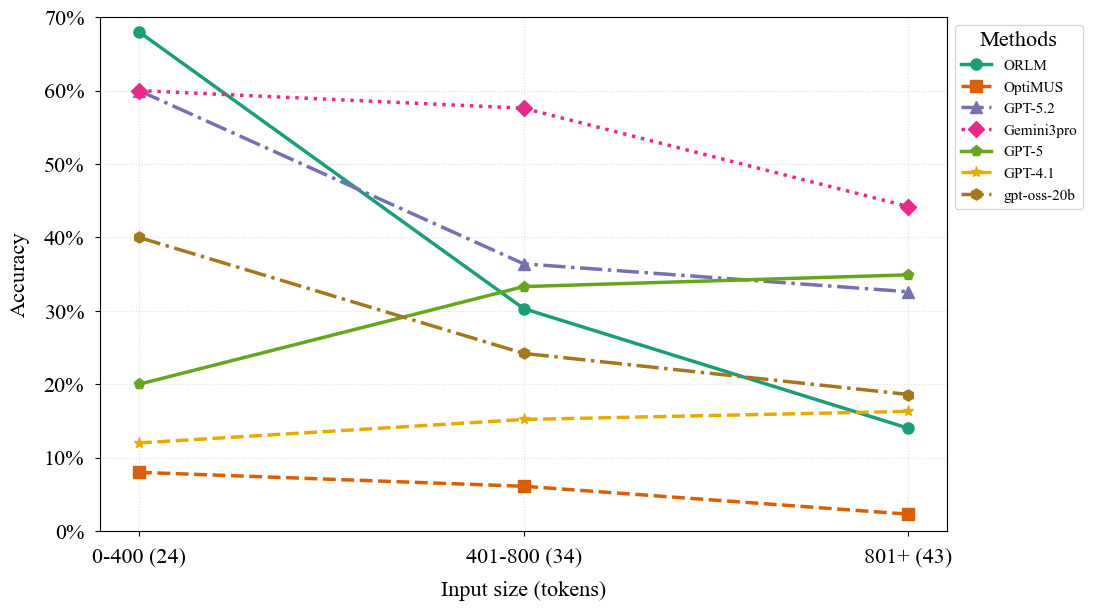}
    \caption{Modeling accuracy of existing methods drops substantially as input size grows}
    \label{fig:accuracy_input_intro}
\end{figure}
    \item \textbf{Challenge 2. Long Input Processing and Reasoning:} From a mechanical perspective, although recent LLMs are designed to handle lengthy and complex reasoning tasks, prior work has reported that reasoning performance can already degrade substantially simply because input length increases \citep{levy2024same,li2025long,li2025context}. To examine whether this phenomenon also arises in the optimization modeling setting, Figure~\ref{fig:accuracy_input_intro} reports the performance of several state-of-the-art models on our curated dataset \testdata (introduced in \cref{sec:dataset}). We observe that the accuracies of most models decrease noticeably as the input size grows. In particular, the modeling accuracies of Gemini~3~Pro and GPT-5.2, the flagship models by Google and OpenAI, both drop below 50\% when the number of input tokens exceeds 800--well below their advertised input limits (on the order of 100{,}000 to 1{,}000{,}000 tokens).
\end{itemize}

\subsection{Main Contributions and Key Messages}
To address the above challenges, we propose \agent, a novel \underline{L}ightw\underline{E}ight \underline{A}ge\underline{N}tic workflow construction framework for LLM-assisted large-scale \underline{OPT}imization model auto-formulation. \agent takes a problem description and associated datasets as input and coordinates
three LLM agents to output the optimization formulation (together with the Python programming code to solve the model). Our main contributions can be summarized as follows:
\begin{enumerate}
     \item \emph{Innovative Agentic Workflow Construction Approach:} For every input, \agent first uses a classification agent to determine the type of the problem (e.g., network revenue management, resource allocation). Then, as the key innovation of \agent (see \cref{fig:framework with csv} for a graphical illustration), a workflow generation agent dynamically constructs a workflow with one or more reference problems to provide an illustrative procedure for applying customized tools and structured reasoning to formulate optimization models for this problem class. The workflow specifies which components of the datasets require focused attention and guides the agent to gradually form better understanding of the target model. Finally, a model generation agent follows this workflow to understand the problem, extract relevant data with our customized tools, and generate the final answer. Through these steps, \agent successfully decomposes a long and complicated reasoning task into a process with multiple structured steps, enabling an LLM to execute one after another with its reasoning power (we provide a more in-depth discussion on its advantages in \cref{remark:pc});
     
    \item \emph{High-Quality Dataset Curation:} As an intermediate step, we turn data acquired from Kaggle \citep{kaggle} into a set of problem instances. Each problem instance contains a user query (problem description and associated datasets) and label (the problem type and optimization model). From these, we build a 101-entry testing dataset \testdata, serving as a benchmark to reflect diverse practical business applications. At the same time, we also create a reference dataset \refdata, whose entries are substantially different from those in \testdata in terms of semantics and scale (see Section \ref{sec:general} for a detailed measurement), for the agents to use as references. Different from existing benchmark datasets that focus mostly on small-size problem instances (less than 20 variables), \testdata puts emphasis on medium-size (20 to 99 variables, constitutes 26\% of the dataset) and large-size (at least 100 variables, constitutes 50\% of the dataset) problem instances. Both \refdata and \testdata span a variety of applications across different domains;

    \item  \emph{Extensive Numerical Simulations:}  We conduct numerical simulations across different datasets to evaluate the performance of \agent built on GPT-4.1 and gpt-oss-20B (an open source model from OpenAI). In \testdata, both of our models achieve an overall accuracy of above 76\%, substantially outperforming other models such as ORLM, Gemini 3 Pro, and GPT-5.2. We also test the performance of our models in other widely-used small-scale benchmarks such as NL4OPT \citep{ramamonjison2023nl4opt}, IndustryOR \citep{huang_orlm_2024}, Mamo (Easy and Complex) \citep{huang2024mamo}. The results show that our models continue to achieve high modeling accuracies;

    \item \emph{Singapore Airlines Use Case and Ablation Study:} Finally, we apply our techniques to a Singapore Airlines' use case that concerns its fare type capacity allocation and flight network planning. We focus on the sales-based linear programming (SBLP) formulation \citep{gallego2015general} using Singapore Airlines' simulated datasets. We start from the fare type capacity allocation problem and show that our approach can achieve leading accuracices in \sindataca, a benchmark that consists of 15 problem instances. Next, we consider the network planning problem (i.e., joint fare type capacity allocation and flight scheduling). Our results show that \agent can consistently achieve a small optimality gap in \sindatanp, a benchmark that consists of 21 problem instances. These outperform the competing methods in almost all instances. Through our ablation study, we can clearly see that both the agentic workflow and the data-handling tools are of great importance for the high modeling accuracy. These findings not only underscore the effectiveness but also showcase the practicality of our design. 

\end{enumerate}

\subsection{Related Works}

Known for their exceptional capabilities in processing natural languages at human level, LLMs have been applied in many different domains, such as market research \citep{brand2023market}, human behavior simulation \citep{chen2025behavioralgenerativeagentsenergy}, and user experience design \citep{li2024user}. Our research focuses primarily on the application of LLMs in the automated modeling of optimization problems. Below, we provide a detailed review of the two main types of methods: prompt-based methods and learning-based methods.

The prompt-based methods use existing LLM interfaces to solve challenging problems. Notable examples in this domain include Chain-of-Thought \citep{wei2022chain}, Tree of Thoughts \citep{yao2024tree}, Graph of Thoughts \citep{besta2024graph}, ProgressiveHint Prompting \citep{zheng2023progressive}, and ReAct \citep{yao2023react}. To tackle complex real-world challenges with implicit constraints, \citet{xiao2023chain} propose a multi-agent framework called Chain-of-Experts (CoE). This is the first LLM-based solution for collaborative problem-solving and iterative modeling optimization, using a forward-through construction and backward reflection mechanism. They also build a benchmark dataset (ComplexOR) from various sources, including academic papers, textbooks, and real-world industry scenarios. Furthermore, to solve problems with lengthy descriptions and avoid excessively long prompts, \citet{ahmaditeshnizi2024OptiMUS} propose OptiMUS, a modular LLM-based agent. It can model and solve linear programming problems by using a connection graph to process each constraint and objective independently. OptiMUS also offers a comprehensive solution by supporting writing and debugging solver code. Additionally, they release a dataset with long and challenging optimization problems called NLP4LP. To improve users' understanding of the optimization models, \citet{chen2024diagnosing} introduce a natural language-based system built on GPT-4 named OptiChat to identify the Irreducible Infeasible Subset (IIS) and offer suggestions to make the model feasible. Unlike the above papers, our framework combines well-crafted step-by-step examples with agent collaboration to tackle complex optimization problems featuring lengthy text descriptions and external data input. This framework also allows for a plug-and-play approach, supporting evolving LLM architectures for various scenarios and different testing datasets. Additionally, the framework's excellent performance is demonstrated in a real-world application with Singapore Airlines.

Learning-based methods primarily rely on fine-tuning to train open-source LLMs. The LLaMoCo framework, introduced by \citep{ma2024llamoco}, fine-tunes general LLMs to generate robust and expert-level optimization code. Moreover, \citet{huang2024mamo} propose the Mamo benchmark to evaluate the mathematical modeling capabilities of LLMs. To improve LLMs' ability to automate optimization modeling, \citet{huang_orlm_2024} introduce a semi-automated data synthesis framework to generate optimization issues for training various open-source LLMs, along with the IndustryOR benchmark featuring practical optimization problems to better assess performance. Similarly, \citet{yang2024optibench} propose the OptiBench benchmark with diverse optimization problems and develop the ReSocratic data synthesis method for supervised fine-tuning of multiple open-source models. \citet{jiang2024llmopt} introduce the LLMOPT framework, which significantly enhances LLMs' ability to model and solve general optimization problems by adopting a five-element formulation, aligning the multi-instruction tuning model, and incorporating a self-correction mechanism. \citet{astorga2024autoformulation} combine LLMs within a Monte-Carlo Tree Search framework, systematically exploring the space of possible formulations by leveraging the hierarchical nature of optimization modeling. They also utilize LLMs as formulation correctness evaluators and introduce a pruning technique to achieve significant efficiency gains. More recently, \cite{zhou2025dp} extend fine-tuning-based approaches to formulate dynamic programs. The aforementioned studies require significant expenses for generating large-scale training data and renting GPUs when fine-tuning open-source LLMs. In contrast, our research employs a more efficient and budget-friendly framework, achieving similar or even better performance.

\section{Dataset Compilation} \label{sec:dataset} 

In this section, we present our method for constructing the reference dataset \refdata and the testing dataset \testdata, which will be utilized in the subsequent sections.

\subsection{Reference Dataset Construction}
The reference dataset \refdata is used throughout the modeling process, enabling \agent to identify the problem type of a query and to learn from them to generate the formulation. We note that this is critical because formulations for different types of problems can be substantially different. Therefore, to improve the accuracy, we need to accurately identify the type of problem and perform the subsequent steps accordingly. To ensure broad applicability across diverse domains and scales, we construct a 96-entry dataset, \refdata, which include 78 small-scale and 18 large-scale instances. We remark that in the current work, we focus on the prevalent linear programming and mixed integer linear programming, but our approach is general and can go well beyond these.

\begin{table}
\centering
\caption{Distribution of problem types in \refdata} 
\label{tab:small_scale_problems}
\resizebox{0.7\textwidth}{!}{
\begin{tabular}{l c c}
\toprule
\textbf{Problem Type}               & \hspace{3cm} & \textbf{Count} \\ 
\midrule
Resource Allocation (RA) & \hspace{3cm} & 24      \\
Mixture  & \hspace{3cm} & 24       \\
Others & \hspace{3cm} & 16      \\ 
Facility Location Problem (FLP) & \hspace{3cm} & 9      \\ 
Assignment Problem  (AP)   & \hspace{3cm} & 8      \\ 
Transportation Problem  (TP)     & \hspace{3cm} & 7     \\
Network Revenue Management (NRM)  & \hspace{3cm} & 5       \\
Sales-Based Linear Programming (SBLP) & \hspace{3cm} & 3     \\ 
\bottomrule
\end{tabular}
}
\end{table}

In \refdata, we include several types of optimization problems, such as network revenue management (NRM), resource allocation (RA), and transportation problem (TP), that are widely studied and used in the literature (see, e.g., \cite{bertsimas1997linear}). In addition, we also include problem instances of the assignment problem (AP), facility location problem (FLP), and other types (e.g., minimum-cost flow problem or a combination of multiple types). We include additional instances of SBLP to be used in our Singapore Airlines use case. Table \ref{tab:small_scale_problems} summarizes the distribution of each problem type. To ensure the problem instances are practically relevant, we start from textbooks such as \cite{bertsimas1997linear} and Kaggle \citep{kaggle}, a widely recognized data science platform with high-quality data across different application domains. Other sources include real-world industry cases and practical studies reported in the literature. We focus on several prevalent operations management scenarios such as retailing, assortment optimization, and logistics \& supply chain (see Section \ref{sec:large-scale} of the appendix for more details). We note that it is possible that some key parameters are missing in the datasets. In this case, we supplement them using reasonable rules and a standardized generation methodology. We detail the codes for generating the missing key parameters for different types of problems in Section \ref{sec:AppendixB} of the appendix. 

Each instance in \refdata can be expressed using a tuple $(q,t,g,f,m)$. The user's query $q$ includes the problem descriptions and the corresponding datasets, the problem type is denoted as $t$, and the optimization model $m$ is generated by our experts. Besides, we also use $g$ and $f$ to additionally capture the related data category (i.e., the type of the data that is relevant to the optimization model) and the details of the relevant data. We note that some of the instances in \refdata can be of small-scale as they are helpful in problem classification. An example of a problem instance in \refdata is given below\footnote{To test the generalization capability of our method and avoid overfitting, we purposely keep the instances in \refdata simple. We shall report the comparison between \refdata and \testdata in terms of length and semantics in Section \ref{sec:general}.}.
\begin{tcolorbox}[colback=white, colframe=boxgrey, coltitle=white, coltext=black,title=Query $q$ (sample instance in \refdata), breakable]
\small
\label{tbox:5}
The data for the store offering several styles of Nike shoes is provided in \enquote{\texttt{NRM\_example/Nike Shoes Sales.csv}}. Through the dataset, the revenue of each shoe is listed in the column `revenue'. The demand for each style is independent. The objective of the store is to maximize the total expected revenue based on the fixed initial inventories of the `Nike x OliviaKim brand', which are detailed in the column `inventory'. During the sales horizon, no replenishment is allowed and there are no in-transit inventories. Customer arrivals, corresponding to demand for different styles of Nike shoes, occur in each period according to a Poisson process, with arrival rates specified in the column `demand'. Moreover, the trade will be started only when the total demand is no less than 100 to ensure trading efficiency. The decision variables $x_i$ represent the number of customer requests the store intends to fulfill for Nike shoe style $i$, with each $x_i$ being a positive integer.
\end{tcolorbox}

\begin{tcolorbox}[colback=white, colframe=boxgrey, coltitle=white, coltext=black,title=Problem Type $t$ (sample instance in \refdata), breakable]
\label{tbox:5}
\small Network Revenue Management.
\end{tcolorbox}

\begin{tcolorbox}[colback=white, colframe=boxgrey, coltitle=white, coltext=black,title=Data Category $g$ (sample instance in \refdata), breakable]
\small
\label{tbox:5}
 \small Nike x OliviaKim brand.
\end{tcolorbox}

\begin{tcolorbox}[colback=white, colframe=boxgrey, coltitle=white, coltext=black,title=Details of the Relevant Data $f$ (sample instance in \refdata), breakable]
\small
 \small 1. Product Name: Nike x Olivia Kim Air Mowabb, Revenue: 11197, Demand: 17, Initial Inventory: 97\\
\small 2. Product Name: Nike x Olivia Kim Air Footscape, Revenue: 9097, Demand: 26, Initial Inventory: 240\\
\small 3. Product Name: Nike x Olivia Kim Air Max 98, Revenue: 11197, Demand: 50, Initial Inventory: 322\\
\small 4. Product Name: Nike x Olivia Kim Air Force 1'07, Revenue: 9995, Demand: 53, Initial Inventory: 281
\end{tcolorbox}

\begin{tcolorbox}[colback=white, colframe=boxgrey, coltitle=white, coltext=black,title=Label $m$ (sample instance in \refdata), breakable]
\small
\textbf{Objective Function:}

$\quad \quad \max \quad \sum_i A_i \cdot x_i$

\textbf{ Constraints:}

1. Inventory Constraints: 

$\quad \quad x_i \leq I_i, \quad \forall i $

2. Demand Constraints: 

$\quad \quad x_i \leq d_i, \quad \forall i $

3. Variable Constraints: 

$\quad \quad x_i \in \mathbb Z_+, \quad \forall i $

\textbf{Retrieved Information:}

$\quad \quad I = [97, 240, 322, 281]$

$\quad \quad A =  [11197, 9097, 11197, 9995]$

$\quad \quad d =  [17, 26, 50, 53]$

\end{tcolorbox}

\subsection{\testdata Testing Dataset}
\label{sec:test_data}
\begin{figure}[!ht]
    \centering
    \includegraphics[width=0.9\linewidth]{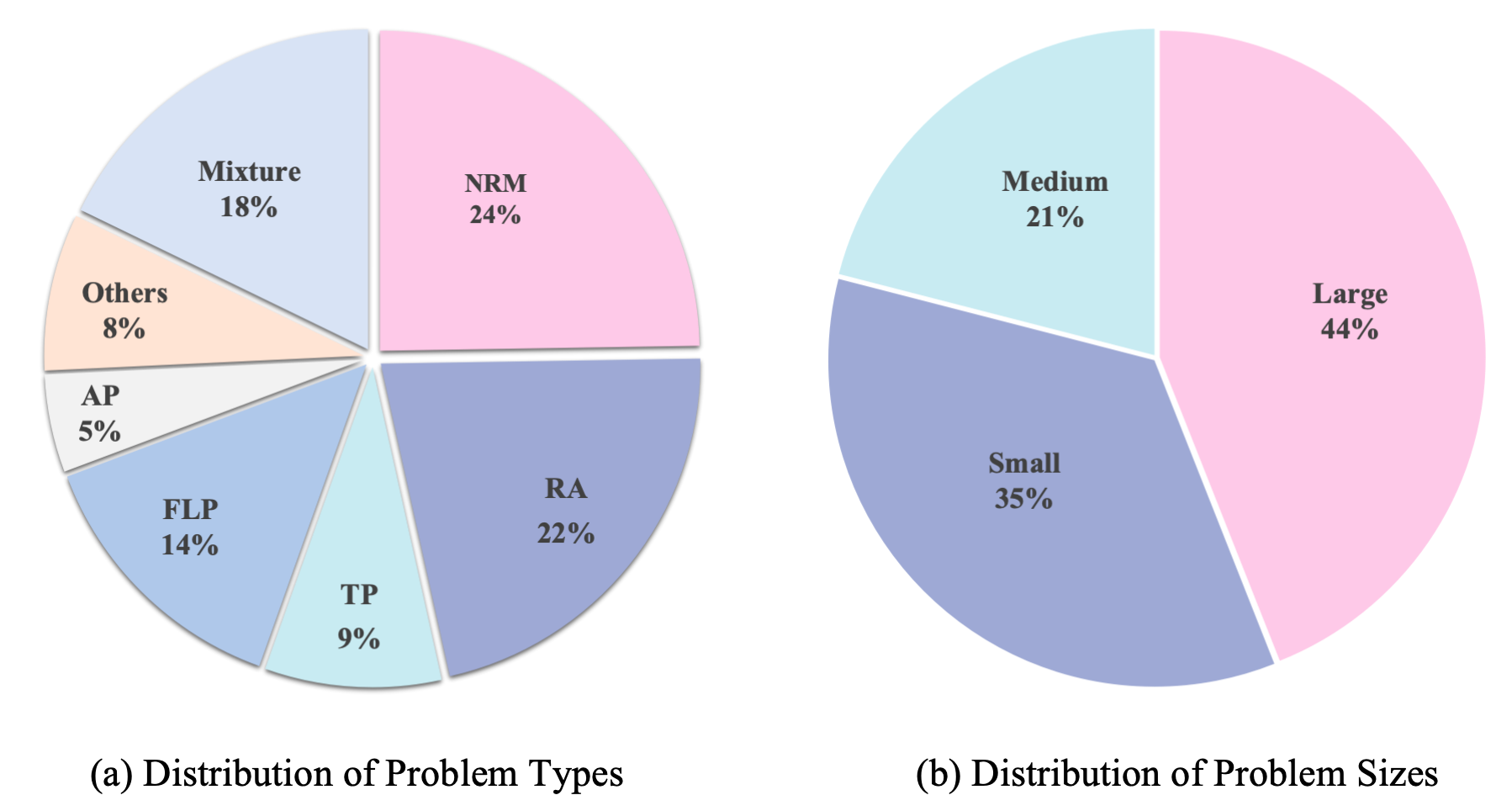}
    \caption{Statistics of \testdata (all problem instances are large-scale)}
    \label{fig:enter-label}
\end{figure}

At the same time, to test the performance of different methods, we construct a 101-entry testing dataset \testdata. Each instance in the \testdata has two key parts: a natural-language query including problem descriptions and corresponding datasets, and a label including a ground-truth mathematical model and the optimal value. To ensure a comprehensive evaluation of \agent's applicability across diverse real-world scenarios, \testdata includes applications spanning medical, e-commerce, and supply chain management. From the perspective of problem types, \testdata consists of problems in common categories such as NRM and RA. It also has problems that are mixture of different types. We provide more details on how we determine the type of a problem in Section \ref{appendix:ex_classification} of the appendix. Among the instances in \testdata, half of them are large-size problems (at least 100 variables). Figure \ref{fig:enter-label} presents the summary statistics of \testdata. Due to limitation of space, we provide an example instance in \testdata in \cref{sec:framework} when introducing \agent and an additional example of problem instances in \testdata in Section \ref{appendix:sample_test} of the appendix.

\subsection{Textual Comparisons between \refdata and \testdata}\label{sec:general}
To mitigate the risk of overfitting or data leakage, we conduct a rigorous textual similarity analysis between \refdata and all the benchmarks we plan to use, including \testdata and other small-scale ones. 

We first consider the semantic aspect. For each benchmark, we report the average semantic similarity between each instance in the benchmark and its closest counterpart in \refdata. For this task, we use the cross-encoder (CE) approach \citep{DBLP:conf/naacl/DevlinCLT19} to serve as our judge. CE computes a single score in $[0,1]$ (higher indicates more similar) for a pair of texts to measure their similarity using a transformer. This method allows for full self-attention between the two texts at every layer, and yields rich interactions that can result in significantly higher accuracy compared to other alternatives \citep{DBLP:conf/iclr/HumeauSLW20}. Specifically, we use the stsb-TinyBERT-L4  CE model \citep{reimers2019sentence-bert}, which is pretrained using the widely recognized Semantic Textual Similarity (STS) benchmark \citep{cer2017semeval}. This averaged score quantifies the expected degree of semantic overlap between benchmarks and \refdata, thereby providing a formal metric for evaluating instance novelty and mitigating risks of data contamination or overfitting in downstream optimization tasks.

In our evaluation, for each query $q$ in the benchmark, we use the CE model to identify the instance in \refdata with the highest similarity score to $q$. We then compute the average of these maximum similarity scores across all queries in the benchmark. The results are presented in Table \ref{tab:similarity}. To facilitate interpretation, we can multiply each score by 5 and map it to the scale shown in Table 1 of \cite{agirre2013sem}. The results validate that the semantic similarities between \refdata and the benchmarks appear to lie toward the lower end of the scale.
\begin{table}[h!]
\caption{Text similarity between \refdata and different benchmarks} \label{tab:similarity} 
\centering
 \begin{tabular}{c c c c c c} 
 \hline
 &\testdata& NL4OPT & MAMO Easy & MAMO Complex & IndustryOR  \\ 
 \hline
 \refdata &0.445& 0.476 & 0.479 & 0.358& 0.413\\
 \hline
 \end{tabular}
\end{table}

Besides, we also compute the average token counts of \refdata, \testdata, and other benchmarks using the $\text{o200k\_base}$ tokenizer from the $\text{tiktoken}$ library and the results is presented in Table \ref{tab:similarity_2}. As we can see, the average token count of the instances in \refdata is much smaller compared to the average token count of the instances in \testdata. The scale of inputs in \testdata is also much larger than those in other benchmarks.
\begin{table}[h!]
\caption{Average token counts of \refdata and the benchmarks} \label{tab:similarity_2} 
\centering
 \begin{tabular}{c c c c c c} 
 \hline
 \refdata&\testdata& NL4OPT & MAMO Easy & MAMO Complex & IndustryOR \\ 
 \hline
 422&47269&121&217&457&266\\
 \hline
 \end{tabular}
\end{table}

Together, we can conclude that the instances in \refdata are substantially different from those in the benchmarks. Also, the instances in \testdata are significantly larger in terms of token counts than those in \refdata. Hence, in order to perform well across all the benchmarks, a method has to be able to generalize well across both the semantic and input scale dimensions.

\section{\agent Framework} \label{sec:framework}
\begin{figure}[h]
    \centering
    \includegraphics[width=
    \linewidth]{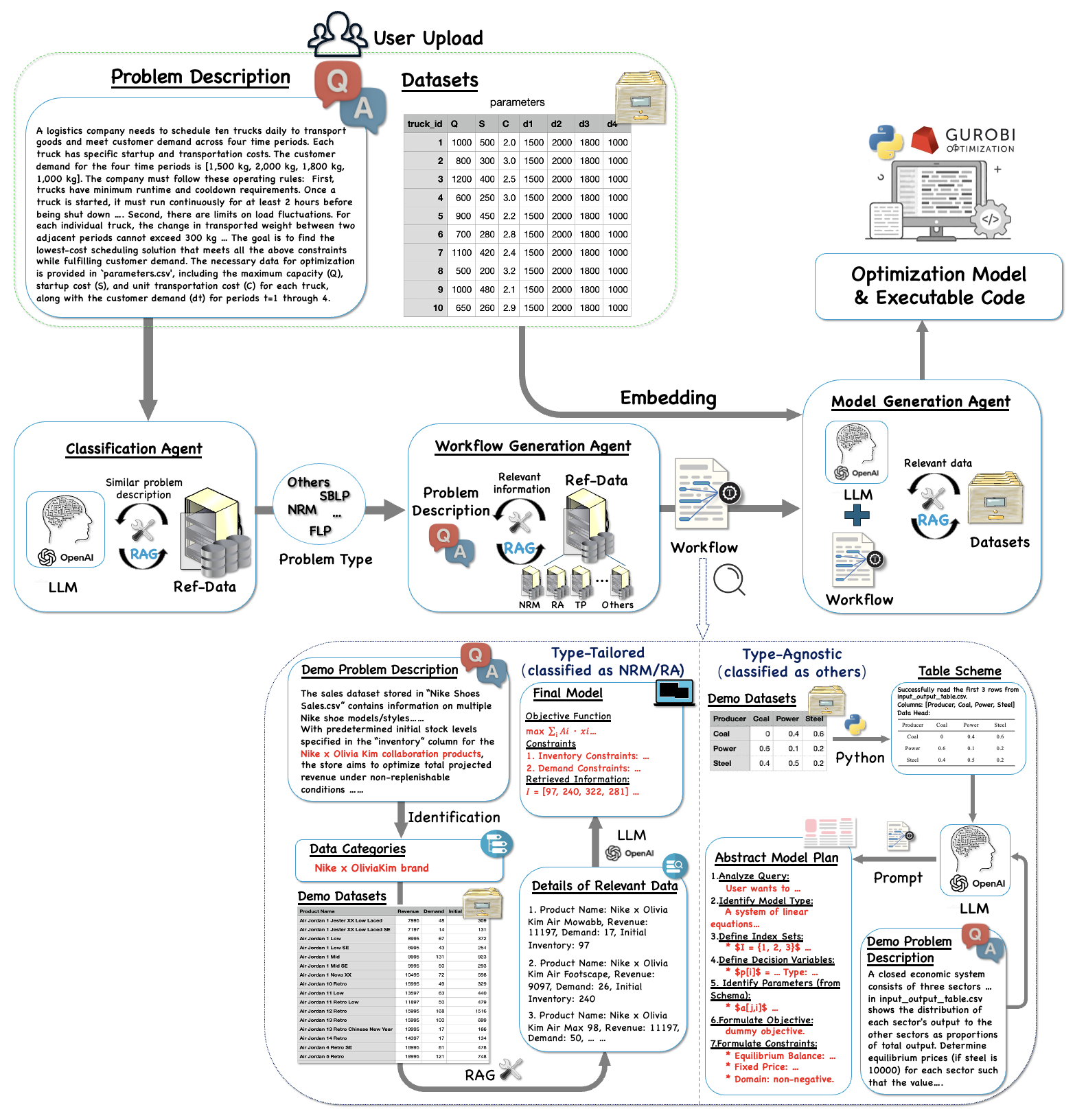}
    \caption{\agent flow-diagram}
    \label{fig:framework with csv}
\end{figure}
We are now ready to present the design details of \agent. \agent builds model for large-scale optimization problems through the collaboration of three LLM agents as illustrated in Figure \ref{fig:framework with csv}. Specifically, after receiving an input, which includes a problem description and the corresponding datasets, the classification agent identifies the problem type and outputs it to the workflow generation agent. Then, the workflow generation agent retrieves the most relevant example(s) of the same type from \refdata, and builds the workflow that demonstrates how to build the model for the example. Subsequently, this workflow is provided to the model generation agent. The latter adheres to the well-defined, step-by-step workflow to generate the optimization formulation. In what follows, we use the below large-scale query as a running example to introduce the three agents.

\begin{tcolorbox}[colback=white, colframe=boxgrey, coltitle=white, coltext=black,title=Input Query $q$ (running example), breakable]
\label{Example of Large-Scale Query}
\small
A logistics company needs to schedule ten trucks daily to transport goods and meet customer demand across four time periods. Each truck has specific startup and transportation costs. The customer demand $d_t$ for the four time periods is [1,500 kg, 2,000 kg, 1,800 kg, 1,000 kg]. The company must follow these operating rules: 

First, trucks have minimum runtime and cooldown requirements. Once a truck is started, it must run continuously for at least 2 hours before being shut down. If turned off, it must remain idle for at least 1 hour before it can be restarted. 

Second, there are limits on load fluctuations. For each individual truck, the change in transported weight between two adjacent periods cannot exceed 300 kg. Additionally, the company must always maintain a 10\% spare capacity buffer for emergency orders. This means the total transported weight in any period must not exceed 90\% of the combined maximum capacity of the trucks active in that period.

The goal is to find the lowest-cost scheduling solution that meets all the above constraints while fulfilling customer demand. The necessary data for optimization is provided in `parameters.csv', including the maximum capacity ($Q$), startup cost ($S$), and unit transportation cost ($C$) for each truck, along with the customer demand $d_t$ for periods $T=[1,\cdots,4]$.
\end{tcolorbox}

\subsection{Problem Classification}
After receiving the query, the  classification agent tries to identify the type of the input problem (e.g., one of those in Table \ref{tab:small_scale_problems}). This is to make sure the subsequent agents can capture the unique characteristics of the problem to generate the workflow and formulate the model. 

At the beginning, the classification agent is first presented with a demo that shows how it should determine the type of a problem step-by-step. Afterwards, the agent follows the same procedure to determine the problem type of the input. Specifically, the demo has the following steps,
\begin{itemize}
    \item \emph{Question:} In this step, a sample problem description (different from the user's one) whose type will be determined following the remaining steps is provided;  
    \item \emph{Thought:} Afterwards, the agent should analyze the query and plan the next steps through stating the objective and reasoning. The demo also instructs the agent to use the retrieval-augmented generation (RAG)\footnote{ RAG is a technique that enables LLMs to generate responses by retrieving relevant information from external datasets. See \cite{gao2023retrieval} for a thorough introduction and survey.} tool FileQA, a critical component of the process. FileQA is a RAG tool that uses similarity search to find problem instances that are closely related to the input problem description from \refdata. In our case, this tool retrieves the top 5 problem instances with the greatest similarity to the input problem and returns their problem descriptions and corresponding problem types;
    
    \item \emph{Action and Action Input:} These two steps kick off the use of the tool FileQA. Specifically, the ``Action" step instructs the agent to use the FileQA tool and ``Action Input" specifies the desired input (the text in the previous ``Question" step) to the FileQA tool;

    \item \emph{Observation, Thought, and Final Answer:} By using the retrieved problem instances, the LLM agent needs to reason the type of the user's input query and provides this in the ``Observation" step. In particular, the agent determines whether the user's input query matches any type from our predefined list as guided by the ``Thought" step. If it does, the answer will be formatted appropriately in the ``Final Answer" step; otherwise, the output would be categorized as ``Others" or ``Mixture". 
\end{itemize} 
\begin{tcolorbox}[colback=white, colframe=boxblue, coltitle=white, coltext=black,title=Demo of the Classification Process, breakable]
\small
{\green \%\% An example that demonstrates how an agent should determine the problem type of the input query step-by-step with the customized RAG tool FileQA}\\
\textbf{Question:} 
What is the problem type in the operation of the text? Please give the answer directly. Text: There are three best-selling items (P1, P2, P3) on Amazon with the profit w\_1,w\_2,w\_3.There is an independent demand stream for each of the products. The objective of the company is to decide which demands to be fulfilled over a ﬁnite sales horizon [0,10] to maximize the total expected revenue from ﬁxed initial inventories. The on-hand inventories for the three items are c\_1,c\_2, and c\_3, respectively. During the sales horizon, replenishment is not allowed and there are no in-transit inventories. Customers who want to purchase P1, P2, P3 arrive at each period according to a Poisson process with a\_1, a\_2, a\_3, the arrival rates, respectively. Decision variables y\_1, y\_2, y\_3 correspond to the number of requests that the firm plans to fulfill for products 1, 2, 3. These variables are all positive integers.\\
\textbf{Thought:} I need to determine the problem type of the content. I'll use the FileQA tool to retrieve the relevant information.\\
\textbf{Action:} FileQA \\
\textbf{Action Input:} ``What is the problem type in the content? content: There are three best-selling items (P1, P2, P3) on Amazon with the profit w\_1, w\_2, w\_3. ..."\\
\textbf{Observation:} The problem type of the content is Network Revenue Management.\\
\textbf{Thought:} The problem type Network Revenue Management is in the allowed list [Network Revenue Management, Resource Allocation, Transportation, Facility Location Problem, Assignment Problem, Sales-Based Linear Programming]. I could get the final answer and finish.

\textbf{Final Answer:} Network Revenue Management.
\end{tcolorbox}

After receiving the input, the classification agent follows the demo to output the problem type by executing each of the steps. For our running example, the type of the input is identified as ``Resource Allocation Problem". As we can see, every step above require certain level of reasoning, but each of them is kept to minimal so that it becomes manageable for the agent.

We point out that the incorporation of this classification agent is not strictly required, yet it can positively impact the modeling accuracy. We shall provide a way to bypass this in the forthcoming Section \ref{sec:agnostic} and demonstrate its performance in Section \ref{sec:numerical}. We also provide a detailed discussion in the forthcoming Remark \ref{remark:classification}.

\subsection{Workflow Generation and Model Generation}\label{sec:workflow}
As a key step, a dedicated workflow generation agent is deployed to build workflow for the downstream model generation agent. This intends to guide the downstream agent to form an initial understanding of the target optimization model. It also instructs the model generation agent to offload mechanical operations, such as data processing, to auxiliary tools, thereby allowing the model generation agent to concentrate its reasoning capacity on the most critical components.

At a high level, the workflow generation agent picks a demo instance from the \refdata that is contextually and structurally similar to the user's input. This is then structured into several components to form the workflow. In this section, we present two kinds of workflows, one type-tailored and one type-agnostic, to accommodate different scenarios. In particular, we use type-tailored workflow for all the problems except for those classified as ``Others" and ``Mixture", and use type-agnostic workflow for these two.

\subsubsection{Type-Tailored Workflow}\label{sec:type_workflow}  In this case, the agent first retrieves from \refdata the most relevant instance (in terms of problem description and datasets) to serve as the demo. This includes its query $q_{\demo}$, data category $g_{\demo}$, details of the relevant data $f_{\demo}$, and the target model $m_{\demo}$ (see Section \ref{appendix:demo} of the appendix for the retrieved details). Then, they are embedded into the workflow that showcases how the optimization model is formulated step-by-step. The workflow includes the following components,
\begin{itemize}
    \item \emph{Question and Thought:} In these two steps, the workflow states clearly the task and the following steps. It also suggests the model generation agent should pay attention to any details such as additional constraints;
    \item \emph{Action and Action Input:} This step invokes the use of the customized RAG tool CSVQA and is a crucial component in data-retrieval\footnote{We note that the tool can be different depending on our base model and the type of the problem.}. With this tool, the input datasets are first converted into vector representations line by line. When a query comes in, it applies similarity search to find lines of data that best match the needs of the problem description. For instance, the demo example $q_{\demo}$ asks for information related to Sony products (see Section \ref{appendix:demo} of the appendix). CSVQA thus retrieves all the products related to the keyword ``Sony" while excluding irrelevant information (e.g., ``Apple" product, which is also brought up in the context, but not used in the final model). Notably, CSVQA supports fuzzy search, which is often required in optimization modeling;
    \item \emph{Observation, Thought, and Final Answer:} These steps first list out all the retrieved data. Then, the ``Thought" step instructs the downstream agent to use the retrieved data to output the label that is of similar form to the ``Final Answer" step. 
\end{itemize} 
In this workflow, several components are type-tailored. For example, through $g_{\demo}$, the downstream agent can be prompted to prioritize dataset components that are likely to be relevant. We also deploy two key type-tailored components:
\begin{enumerate}
    \item \textbf{Type-Specific Output Structure:} 
    The output is organized in the form of ``abstract model + retrieved information'' (see \cref{sec:model_generation_agent}), so the model-generation agent is provided with a type-specific model skeleton in advance. The agent is instructed to produce an output similar to $m_{\demo}$ (with modifications when necessary), which can help it focus on incorporating the retrieved information into the appropriate parts of the formulation;

    \item \textbf{Type-Aware Formatting:} 
    To further facilitate data placement, the retrieved information shown in the ``Observation'' is formatted to support straightforward incorporation into the final output. For example, in our running example the problem type is RA, so each product’s resource consumption and value are presented line by line (see \cref{appendix:demo}), which can make it easier for the downstream agent to associate each numeric entry with the corresponding product. As another example, for TP, demand and supply are provided as separate tables, while transportation costs are provided as a matrix. Additional examples are provided in \cref{appendix:data_format}.
\end{enumerate}

\begin{tcolorbox}[colback=white, colframe=boxagent2, coltitle=white, coltext=black,title=Type-Tailored Workflow for Model Generation,breakable]
\small
{\green \%\% The workflow that demonstrates how an agent should retrieve data and formulate type-tailored optimization model step-by-step with the tool CSVQA} \\
\textbf{Question:} Based on the following problem description and data, please formulate a complete mathematical model using real data from retrieval. \\
\textcolor{blue}{\{$q_{\demo}$: Demo Problem Description\}} \\
\textbf{Thought:} I need to formulate the objective function and constraints of the optimization model based on the user's description and the provided data. I should retrieve the relevant information from the CSV file. Pay attention: 1. If the data to be retrieved is not specified, retrieve the whole dataset instead. 2. I should pay attention to whether there are further detailed constraints in the problem description. If so, I should generate an additional constraint formula. 3. The final expressions should not be simplified or abbreviated.\\
\textbf{Action:} CSVQA \\
\textbf{Action Input:} Retrieve all the product data related to \textcolor{blue}{\{$g_{\demo}$: Data Categories Needed by the Demo Problem\}} to formulate the mathematical model with no simplification or abbreviation. Retrieve the documents in order, row by row. Use the given context to answer the question. If you mention a certain kind of product, retrieve all the relevant product information details based on its product name. If no specific kind of product is mentioned, retrieve all the data instead. Only present the final answer in detail of the row, instead of giving a sheet format. \\
\textbf{Observation:} \textcolor{blue}{\{$f_{\demo}$: Retrieved Data for the Demo Problem\}}\\
\textbf{Thought:} Now that I have the necessary data, construct the objective function and constraints using the retrieved data as parameters of the formula. Ensure to include any additional detailed constraints present in the problem description. Always pay attention to the variable type. If not mentioned, use nonnegative integers. Do NOT include any explanations, notes, or extra text. Format the expressions strictly in markdown format. Following this example. The expressions should not be simplified or abbreviated. \\
\textbf{Final Answer:} \textcolor{blue}{\{$m_{\demo}$: Label of the Demo Problem\}}
\end{tcolorbox}
After the workflow is built, the downstream model generation agent can follow it to generate the target output for the input.
\begin{tcolorbox}[colback=white, colframe=boxagent2, coltitle=white, coltext=black,title=The Prompt for the Model Generation Agent,breakable]
\small
{\green \%\% This is the prompt that guides the model generation agent to follow the few-shot example to generate the final optimization model} \\
You are an assistant who generates a mathematical model based on the user's description and provided CSV data.

Please refer to the following example and generate the answer in the same format:

\textcolor{blue}{\{Type-Tailored Workflow for Model Generation\}}.

When you need to retrieve information from the CSV file, use the provided tool.

\textbf{Question}: What is the mathematical model based on the user's query?\\ 
Content: \textcolor{blue}{\{$q$: Input Query\}}
\end{tcolorbox}

\subsubsection{Type-Agnostic Workflow} \label{sec:agnostic}
Using a type-tailored workflow, the downstream model-generation agent is explicitly instructed to display data and construct the model in a prescribed form. While this approach can substantially reduce the agent’s workload and can thereby improve performance, it depends heavily on classification accuracy and typically does not accommodate complex problems involving mixed types. To support a more robust and general workflow that can operate across a wide range of scenarios, we additionally develop a type-agnostic workflow.

Similar to the type-tailored workflow, the type-agnostic workflow also provides a demonstration; however, it includes only the query $q_{\demo}$ and the label $m_{\demo}$. Different from the type-tailored workflow, however, in the type-agnostic workflow, the emphasis is placed on constructing an abstract model from queries and a CSV schema of the input dataset(s), rather than on following the most relevant problem in \refdata to build a similar model. As such, instead of constraining the type or format of the data to be retrieved, the workflow instructs the downstream model-generation agent to follow an \textcolor{magenta}{abstract model plan}.

Under this plan, the agent first inspects the query and lightweight snapshots of the associated datasets (e.g., the first few rows) to guide subsequent decisions. Specifically, it begins by identifying the type of the target model (e.g., LP or MILP) rather than the type of the problem. It then determines fundamental modeling components such as index sets and decision variables. Next, the agent uses the dataset snapshots to identify relevant rows and columns. Afterwards, the agent examines snapshots of the dataset to determine which parts can be extracted as model parameters, and utilizes Python routines to perform the extraction. Finally, based on the information gathered in the preceding steps, it formulates the objective function and constraints sequentially. Meanwhile, required data are extracted via a set of Python routines. Importantly, the model-generation agent is allowed to decide what information to retrieve and how to incorporate it into the final formulation.
\begin{tcolorbox}[colback=white, colframe=boxagent2, coltitle=white, coltext=black, title=Type-Agnostic Workflow for Model Generation, breakable]
\small
{\green \%\% The type-agnostic workflow that demonstrates how an agent should retrieve data and formulate the optimization model by inspecting the dataset snapshot of the input query}\\
You are an expert optimization modeler. Your task is to create an {\color{magenta}``Abstract Model Plan"} based on the user's query and the CSV schema (data structure). This plan is \textit{not} Gurobi code or mathematical formulas, but a clear, step-by-step reasoning process in English.

\textbf{\color{orange}\normalsize[Example]}

\textbf{Question:} \textcolor{blue}{\{$q_{\demo}$: Demo Problem Description\}} \\
\textbf{Thought:} I need to create an Abstract Model Plan based on the user's query and the dataset snapshot (data structure).

\textbf{Final Answer:} {\blue \{$m_{\demo}$: Label of the Demo Problem\}}

\textbf{\color{orange}{\normalsize [Current Task]}}

\textbf{User Query:} \textcolor{blue}{\{$q$: Input Query\}} 

\textbf{CSV Schema:} \textcolor{blue}{\{Snapshots of datasets of $q$\}}

\textbf{\color{orange}\normalsize{[Your Output]}}

You must strictly follow this format for your {\color{magenta}``Abstract Model Plan"} output:

\centering \textbf{\color{magenta}\normalsize------------Abstract Model Plan Start------------}
\begin{enumerate}[leftmargin=*, label=\arabic*., noitemsep, topsep=5pt]
    \item \textbf{Analyze Query:} The user wants to \{Your analysis\}.
    \item \textbf{Identify Model Type:} Based on the query, this is a \{e.g., LP, MIP, Fixed-Charge, Blending\} problem.
    \item \textbf{Define Index Sets:} The primary indices are \{e.g., Products, Workers, Sources, Destinations\}.
    \item \textbf{Define Decision Variables:}
    \begin{itemize}[noitemsep, topsep=2pt]
        \item \texttt{x[i]} = \{Describe first variable, e.g., `quantity of product i'\}. Type: GRB.CONTINUOUS / GRB.INTEGER.
        \item \texttt{y[i]} = \{Describe second variable, e.g., `if product i is produced'\}. Type: GRB.BINARY.
    \end{itemize}
    \item \textbf{Identify Parameters (from CSV Schema):}
    \begin{itemize}[noitemsep, topsep=2pt]
        \item Objective coefficients (e.g., profit) will come from column(s): \{e.g., `Price', `Production Cost'\}.
        \item Constraint coefficients (e.g., resource use) will come from: \{e.g., `Resource 1', `Available Time'\}.
        \item Constraint RHS (limits) will come from: \{e.g., `Max Demand'\}.
    \end{itemize}
    \item \textbf{Formulate Objective:} Describe objective, e.g., 
    $$\text{Maximize} \sum_i \left((\text{schema}[\text{Price}][i] - \text{schema}[\text{Cost}][i]) \cdot x[i] - \text{schema}[\text{FixedCost}][i] \cdot y[i]\right).$$
    \item \textbf{Formulate Constraints:}
    \begin{itemize}[noitemsep, topsep=2pt]
        \item \textbf{Constraint 1} :Describe constraint 1, e.g., Resource Limit:
        $$\sum_i \text{schema}[\text{Resource 1}][i] \cdot x[i] \leq \text{schema}[\text{Available}].$$
        \item \textbf{Constraint 2}: Describe constraint 2, e.g., Linking:
        $$x[i] \leq M \cdot y[i].$$
        \item \ldots (Other constraints) \ldots
    \end{itemize}
\end{enumerate}
\centering \textbf{\color{magenta}\normalsize------------Abstract Model Plan End------------}
\end{tcolorbox}

\subsubsection{Model Generation Agent}\label{sec:model_generation_agent}
Once the workflow is ready, it is used to guide the model generation agent to output the final mathematical model and optional Python code (see the forthcoming Remark \ref{remark:code}). For our running example, the output is shown below.
\begin{tcolorbox}[colback=white, colframe=boxgrey, coltitle=white, coltext=black,title=Output Optimization Model for the Running Example, breakable]
\small
\textbf{Sets and Parameters}
\begin{itemize}
    \item $I = \{1, \dots, 10\}$: Set of available trucks.
    \item $T = \{1, \dots, 4\}$: Set of time periods.
    \item $d_t$: Customer demand in period $t$ ($\text{kg}$).
    \item $Q_i$: Maximum capacity of truck $i$ ($\text{kg}$).
    \item $S_i$: Startup cost for truck $i$.
    \item $C_i$: Unit transportation cost for truck $i$.
\end{itemize}

\textbf{Decision Variables}
\begin{itemize}
    \item $w_{it} \in \mathbb{R}^+$: Weight transported by truck $i$ in period $t$. (Continuous)
    \item $y_{it} \in \{0, 1\}$: Operating status ($1$ if running, $0$ otherwise). (Binary)
    \item $u_{it} \in \{0, 1\}$: Startup status in period $t$ (defined for $t \in \{2, 3, 4\}$). (Binary)
\end{itemize}

\textbf{Objective Function (Minimize Total Cost)}
\begin{align*}
    \min \quad & \sum_{i \in I} \left( \sum_{t=1}^{4} C_i w_{it} \right) + \sum_{i \in I} \left( S_i y_{i1} + \sum_{t=2}^{4} S_i u_{it} \right)
\end{align*}

\textbf{Constraints}
\begin{enumerate}
    \item \textbf{Demand Satisfaction}
    $$\sum_{i \in I} w_{it} \geq d_t, \quad \forall t \in T$$

    \item \textbf{Capacity and Status Linkage}
    $$w_{it} \leq Q_i y_{it}, \quad \forall i \in I, t \in T$$

    \item \textbf{Spare Capacity Buffer (10\% Reserve)}
    $$\sum_{i \in I} w_{it} \leq 0.9 \sum_{i \in I} Q_i y_{it}, \quad \forall t \in T$$

    \item \textbf{Startup Variable Definition}
    \begin{align*}
        u_{it} & \geq y_{it} - y_{i,t-1}, & \forall i \in I, t \in \{2, 3, 4\} \\
        u_{it} & \leq y_{it}, & \forall i \in I, t \in \{2, 3, 4\}\\
        u_{it} & \leq 1 - y_{i,t-1}, & \forall i \in I, t \in \{2, 3, 4\}
    \end{align*}

    \item \textbf{Minimum Runtime (2 Periods)}
    \begin{align*}
        y_{i1} & \leq y_{i2}, & \forall i \in I\\
        y_{it} + y_{i,t+1} & \geq 2 u_{it}, & \forall i \in I, t \in \{2, 3\}\\
        u_{i4} & = 0, & \forall i \in I
    \end{align*}

    \item \textbf{Cooldown Constraints (1 Period Off)}
    $$y_{i,t-1} - y_{it} \leq 1 - y_{i,t+1}, \quad \forall i \in I, t \in \{2, 3\}$$

    \item \textbf{Load Fluctuation ($\pm 300 \text{ kg}$)}
    \begin{align*}
        w_{it} - w_{i,t-1} & \leq 300, & \forall i \in I, t \in \{2, 3, 4\}\\
        w_{i,t-1} - w_{it} & \leq 300, & \forall i \in I, t \in \{2, 3, 4\}
    \end{align*}
    \item \textbf{Variable Domains}
    \begin{align*}
        w_{it} & \geq 0, & \forall i \in I, t \in T \\
        y_{it} & \in \{0, 1\}, & \forall i \in I, t \in T\\
        u_{it} & \in \{0, 1\}, & \forall i \in I, t \in \{2, 3, 4\}
    \end{align*}
    \item \textbf{Retrieved Information}
    
    For $t = \in \{1,2,3,4\}$:
    \begin{align*}
        d = [1500, 2000, 1800, 1000]
    \end{align*}
    For $i \in \{1, \cdots,10\}$:
    \begin{align*}
        Q &= [1000, 800, 1200, 600, 900, 700, 1100, 500, 1000, 650]  \\
      S & = [500, 300, 400, 250, 450, 280, 420, 200, 480, 260]\\
        C &= [2.0, 3.0, 2.5, 3.0, 2.2, 2.8, 2.4, 3.2, 2.1, 2.9]
    \end{align*}
\end{enumerate}

\end{tcolorbox}

\begin{remark}[Code Generation]\label{remark:code}
   In practical applications, users usually not only want to have the optimization formulation, but also the corresponding optimal solution. To meet this need, our framework includes a dedicated component that automatically translates the generated complete optimization model—or an abstract model plan together with its corresponding CSV schema—into executable code compatible with optimization solvers such as Gurobi. The implementation details of this component are provided in Section \ref{sec:AppendixC} of the appendix. 
\end{remark}

\begin{remark}[The Benefit of Classification and Our Design]\label{remark:classification}
We note that the classification agent is not strictly required as \agent can perform well even without it (see the numerical experiment  results in Table \ref{tab:by-family-transposed}). However, using a classification agent can help customize the subsequent workflow for commonly encountered problem types, and hence, freeing up reasoning bandwidth for the downstream agents to further improve performance. Two of our design choices worth further discussions.
\begin{itemize}
    \item First, one may consider directly using RAG without LLM reasoning. However, as we show in Section \ref{sec:AppendixR} of the appendix, the classification accuracy can be negatively impacted if so. This underscores the importance of leveraging the reasoning capability of the LLMs;
    \item Moreover, given the large input token limits of modern LLMs, one might argue that we could simply include the entirety of \refdata and allow the agents to decide autonomously which parts are relevant. However, as noted in Challenge~1 of \cref{sec:Intro}, the performance of state-of-the-art LLMs can degrade substantially once the input length exceeds roughly 800 tokens. Since \refdata contains more than 40{,}000 tokens, directly placing it all into the prompt would likely exacerbate this issue and could further hinder the model’s ability to generate accurate formulations.

\end{itemize}

\end{remark}

\subsection{Novelties in Our Design and Key Messages}
\label{remark:pc}
\agent presents a fundamentally different design principle compared to prior fine-tuning approaches and we provide discussion here.

\vspace{2mm}
\noindent\textbf{Harness LLM Flexibility via Structured Workflow:} As discussed earlier, a central challenge in applying LLMs to large-scale optimization model auto-formulation is reasoning over long, heterogeneous, and often unstructured token sequences, given that it is infeasible to collect sufficient data to cover all possible cases. To mitigate this challenge, \agent follows common practices to design workflows that decompose a complex optimization modeling task into a sequence of structured sub-tasks, facilitating partial standardization. Moreover, given the versatility of optimization modeling, the workflows are constructed dynamically and leverage LLMs’ natural language processing capabilities. Specifically,
\begin{itemize}
    \item Inspired by common human modeling practices, both workflows explicitly define a sequence of steps for the downstream agent to follow and offload token-intensive yet largely mechanical data-handling operations to dedicated tools. Together, these design choices reduce the downstream agent’s burden, in terms of token consumption and reasoning, associated with planning and data processing; and
    \item Leveraging LLMs’ flexible language understanding capabilities, both workflows are instantiated around a reference problem, helping the downstream agent identify which parts of the datasets merit closer examination and form an initial hypothesis of the target optimization model, rather than reasoning entirely from scratch. LLMs' flexibility also helps to handle the variability in the inputs (e.g., different column names of similar problems), and minimizes the needs for human intervention.
\end{itemize}
Together, the workflows help direct the model-generation agent’s attention to the most challenging components of the modeling task (e.g., interpreting the query intent, aligning coefficients across the objective and constraints, and incorporating subtle but critical constraints), enabling the optimization model to be constructed more incrementally and effectively. 

\vspace{2mm}
\noindent\textbf{Resource Efficiency and Potential for Generalization:} As shown in Table~\ref{tab:contribution}, compared to prior approaches to optimization modeling, our method is distinguished by integrating LLMs’ ability to process unstructured information into a structured workflow. This design supports handling large-scale inputs without requiring task-specific fine-tuning, which can help mitigate the substantial resource costs---especially labeled data requirements---associated with fine-tuning-based approaches. We also emphasize that, as discussed in \cref{sec:general}, the data in \refdata can differ substantially from the benchmark instances in both semantics and length.

Despite these gaps, our framework can still utilize \refdata effectively, suggesting potential for generalization as detailed below.
\begin{itemize}
    \item By including different problem instances in \refdata, one can apply \agent to domains and scenarios that are completely different than the ones would be considered in the current work; 

    \item Moreover, our plug-and-play strategy allows \agent to be readily paired with more advanced LLMs as they become available. This can make sure we can always leverage the continuously evolving reasoning power of the latest LLMs.
\end{itemize}

\begin{table}[h]
\centering
\caption{Resource consumption comparison} 
\label{tab:contribution}
{\begin{tabular}{lcccc}
\toprule
& Training dataset size& Support large-scale input\\ 
\midrule
ORLM \citep{huang_orlm_2024}& 30,000&$\times$\\
OptiBench \citep{yang2024optibench}& 29,000  & $\times$                                    \\
LLMOPT \citep{jiang2024llmopt} & 79,391                & $\times$                                    \\
\agent (Ours)                                                        & 96                   & $\checkmark$                                   \\ \bottomrule
\end{tabular}}
\end{table}

\section{Numerical Simulations} \label{sec:numerical}
In this section, we present numerical simulations to assess the effectiveness of \agent. To test its effectiveness, we instantiate \agent on less capable LLM models, such as GPT-4.1, a old version GPT model, and gpt-oss-20B, a open source model released by OpenAI. We note that gpt-oss-20B is a much smaller model in terms of number of parameters and using it as the base model for \agent can be viewed as a challenging test for our framework.

\noindent\textbf{Baselines:} For our competing methods, we include Gemini 3 Pro and GPT-5.2, the latest flagship models from Google and OpenAI, which are known to be very capable in mathematics and reasoning. We also include ORLM \citep{huang_orlm_2024} and OptiMUS \citep{ahmaditeshnizi2024OptiMUS}. Older versions GPT models such as GPT-5, GPT-4.1, and gpt-oss-20B. We note that \cite{huang_orlm_2024} also report results from several other works for small-scale benchmarks, and we include them here for ease of comparisons as well.

\vspace{2mm}
\noindent\textbf{Benchmarks:} To the best of our knowledge, \testdata appears to be the only benchmark for large-scale optimization modeling (i.e., problem instances whose data has to be stored separately in datasets). We thus use it to test each model's performance on large-scale problems.

For completeness, we also include benchmarks for small-scale optimization problems. These include IndustryOR \citep{huang_orlm_2024}, NL4OPT benchmark \citep{ramamonjison2023nl4opt},  MAMO Easy, and MAMO Complex \citep{huang2024mamo}. The IndustryOR benchmark consists of 100 real-world operations research problems from eight industries, covering five types of optimization models (linear programming, non-linear programming, integer programming, mixed-integer programming, and others) across three difficulty levels. The NL4OPT benchmark extracts and adapts linear programming problems from a variety of real-world application scenarios, covering a wide range of domains such as supply chain management, production planning, and resource allocation. Each problem is described in natural language with a corresponding optimal value. The MAMO benchmarks also include a variety of linear programming and mixed-integer linear programming problems with MAMO complex being a more challenging set of problem instances. In this work, we use the most up-to-date versions of these benchmarks provided by \cite{chen2025sirl}, with known errors corrected.

\vspace{2mm}
\noindent\textbf{Metrics and Solver Configurations:} 
For all benchmarks, we first follow the convention in \cite{huang_orlm_2024} to use the computed optimal value of the output model as our evaluation metric. In particular, the modeling accuracy for a benchmark is defined as 
\begin{align}
    \text{Modeling accuracy} = \frac{\#\text{resulted optimal value that matches the ground truth optimal value}}{\#\text{problems}}.
\end{align} All evaluations are conducted using Gurobi Optimizer 13.0 \citep{gurobi} with the optimality tolerance ($\epsilon$-gap) fixed at $10^{-4}$. Crucially, we verify that every problem instance in our dataset admits a unique optimal solution. This property ensures that the evaluation is invariant to solver algorithmic choices (e.g., Simplex vs. Interior Point) or constraint permutations, providing a deterministic ground truth for correctness verification.

For \testdata, we additionally conduct a manual check on the generated formulations, and compare each of them against the ground truth formulation. This evaluates if the generated formulation is an exact match (EM) to the ground truth except for naming of variables. Formally,
\begin{align}
    \text{Modeling accuracy (EM)} = \frac{\#\text{generated formulation that matches the ground truth formulation}}{\#\text{problems}}.
\end{align}
In our results on \testdata, we shall show that modeling accuracy is indeed a good proxy for modeling accuracy (EM). 

\begin{table}[htbp]
\centering
\small
\setlength{\tabcolsep}{3pt}
\renewcommand{\arraystretch}{1.5}
\caption{\textbf{Modeling accuracy (EM) by problem type. \texttt{LEAN} is the abbreviation of \agent. The best performing model in each column is boldfaced and the second best is underlined }}\label{tab:by-family-transposed_em}
\begin{tabular}{@{}l*{8}{c}@{}}
\toprule
\textbf{Model} & \textbf{NRM (25)} & \textbf{RA (22)} & \textbf{TP (9)} & \textbf{FLP (14)} & \textbf{AP (5)} & \textbf{Others (8)} & \textbf{Mixture (18)} & \textbf{Overall} \\
\midrule
ORLM                    & 28\%   & 45.5\% & 0\%& 0\% & 60\%  & 0\%& 5.6\%  & 19.8\% \\
OptiMUS                 & 0\%    & 18.2\%     & 0\%    & 0\% & 0\%  & 12.5\%    & 0\%  & 5\%  \\
\midrule
Gemini 3 Pro             & 4\%    & 63.6\% & \textbf{100\%}    & 50\% & \underline{80\%}  & \underline{75\%}   & 55.6\% & 50.5\% \\
GPT-5.2                  & 0\%   & \underline{68.2\%} & 77.8\%& 35.7\% & 40\%  & 50\%& 33.3\% & 38.6\% \\
GPT-5                   & 40\%   & 13.6\% & 33.3\%& 35.7\% & \underline{80\%}  & 12.5\%& 27.8\% & 30.7\% \\
GPT-4.1                 & 20\%   & 18.2\% & 22.2\%& 21.4\% & 0\%   & 0\%    & 5.6\%  & 14.9\% \\
gpt-oss-20B             & 0\%    & 45.5\% & 0\%    & 28.6\% & 40\%  & 25\%   & 44.4\% & 25.7\% \\
\midrule
\agentabbrv (GPT-4.1)        & \underline{72\%}   & \textbf{86.4\%} & \textbf{100\%}  & \textbf{78.6\%} & \textbf{100\%} & \textbf{100\%}  & \textbf{77.8\%} & \textbf{83.2\%} \\
\agentabbrv (gpt-oss-20B)    & \textbf{80\%}   & \textbf{86.4\%} & \underline{88.9\%}  & \underline{64.3\%} & 60\%  & \underline{75\%}  & \underline{72.2\%} & \underline{77.2\%} \\
\agentabbrv (GPT-4.1-Agnostic)             & 20\%    & 45.5\% & 44.4\% & 7.1\%  & \underline{80\%}  & \textbf{100\%} & \textbf{77.8\%}& 45.5\% \\
\agentabbrv (gpt-oss-Agnostic)             & 12\%    & 54.6\% & \underline{88.9\%} & 28.6\%  & 40\%   & \underline{75\%} & \underline{72.2\%} & 47.5\% \\
\bottomrule
\end{tabular}
\end{table}

\begin{table}[htbp]
\centering
\small
\setlength{\tabcolsep}{3pt}
\renewcommand{\arraystretch}{1.5}
\caption{\textbf{Modeling accuracy by problem type. \texttt{LEAN} is the abbreviation of \agent. The best performing model in each column is boldfaced and the second best is underlined }}\label{tab:by-family-transposed}
\begin{tabular}{@{}l*{8}{c}@{}}
\toprule
\textbf{Model} & \textbf{NRM (25)} & \textbf{RA (22)} & \textbf{TP (9)} & \textbf{FLP (14)} & \textbf{AP (5)} & \textbf{Others (8)} & \textbf{Mixture (18)} & \textbf{Overall} \\
\midrule
ORLM                    & 32\%   & 68.2\% & 33.3\%& 14.3\% & 60\%  & 12.5\%& 5.6\%  & 32.7\% \\
OptiMUS                 & 0\%    & 18.2\%     & 0\%    & 0\% & 0\%  & 12.5\%    & 0\%  & 5\%  \\
\midrule
Gemini 3 Pro             & 12\%    & 63.6\% & \textbf{100\%}    & 50\% & \underline{80\%}  & \underline{75\%}   & 55.6\% & 52.5\% \\
GPT-5.2                  & 4\%   & 68.2\% & 77.8\%& 35.7\% & 40\%  & 50\%& 38.9\% & 40.6\% \\
GPT-5                   & 40\%   & 13.6\% & 33.3\%& 35.7\% & \underline{80\%}  & 12.5\%& 27.8\% & 30.7\% \\
GPT-4.1                 & 20\%   & 18.2\% & 22.2\%& 21.4\% & 0\%   & 0\%    & 5.6\%  & 14.9\% \\
gpt-oss-20B             & 0\%    & 45.5\% & 0\%    & 28.6\% & 40\%  & 25\%   & 44.4\% & 25.7\% \\
\midrule
\agentabbrv (GPT-4.1)        & \underline{72\%}   & \underline{86.4\%} & \textbf{100\%}  & \textbf{85.7\%} & \textbf{100\%} & \textbf{100\%}  & \underline{83.3\%} & \textbf{85.1\%} \\
\agentabbrv (gpt-oss-20B)    & \textbf{80\%}   & \textbf{90.9\%} & \textbf{100\%}  & \underline{64.3\%} & 60\%  & \underline{75\%}  & 77.8\% & \underline{80.2\%} \\
\agentabbrv (GPT-4.1-Agnostic)             & 20\%    & 68.2\% & 77.8\% & 7.1\%  & \underline{80\%}  & \textbf{100\%} & \textbf{100\%}& 54.5\% \\
\agentabbrv (gpt-oss-Agnostic)             & 12\%    & 59.1\% & \underline{88.9\%} & 28.6\%  & 60\%   & \underline{75\%} & 77.8\% & 50.5\% \\
\bottomrule
\end{tabular}
\end{table}

\vspace{2mm}
\noindent\textbf{Results on \testdata:} Table~\ref{tab:by-family-transposed_em} summarizes the modeling accuracy (EM) of different models on \testdata (\texttt{LEAN} is the abbreviation of \agent). We report the overall results as well as the detailed results on different problem types. \agent (GPT-4.1) and \agent (gpt-oss-20B) achieve highest overall accuracies, substantially exceeding the competing methods' modeling accuracies. It is worth mentioning that their respective base models, GPT-4.1 and gpt-oss-20B, could only achieve overall accuracies of less than 26\%. This means \agent is able to dramatically improve their performance by stimulating their reasoning capabilities with the agentic workflows. 

Among the competing methods, Gemini 3 Pro has the best modeling accuracy (EM) of 50.5\%. The fine-tuned model ORLM has an overall modeling accuracy (EM) of 19.8\% while OptiMUS is only able to achieve a 5\% modeling accuracy (EM). After a careful review, we found that a key factor affecting OptiMUS's performance is its requirement that each problem's datasets adhere to a predefined format. Despite our best efforts to convert our datasets to meet this specification, mismatches between the expected schema and the actual data still occur frequently, which ultimately leads to errors. In contrast, \agent does not impose any constraints on the datasets' structures. Instead, it only provides a problem-relevant workflow together with data handling tools to the model generation agent. This enables the latter to freely generate the details of the model in accordance with the input problem.

Recall that in \agent, only problems classified as ``Others" and ``Mixture" would be handled by the type-agnostic workflow. To perform a robustness check, we manually route all problems to the type-agnostic workflow. The results are shown under \agent (GPT-4.1-Agnostic) and \agent (gpt-oss-Agnostic). As we can see, they perform worse than \agent (GPT-4.1) and \agent (gpt-oss-20B), but still manage to outperform almost all other competing methods. These showcase the effectiveness of our type-tailored workflow and also the robustness of our type-agnostic workflow.

\begin{figure}[h]
    \centering
    \includegraphics[width=0.9\linewidth]{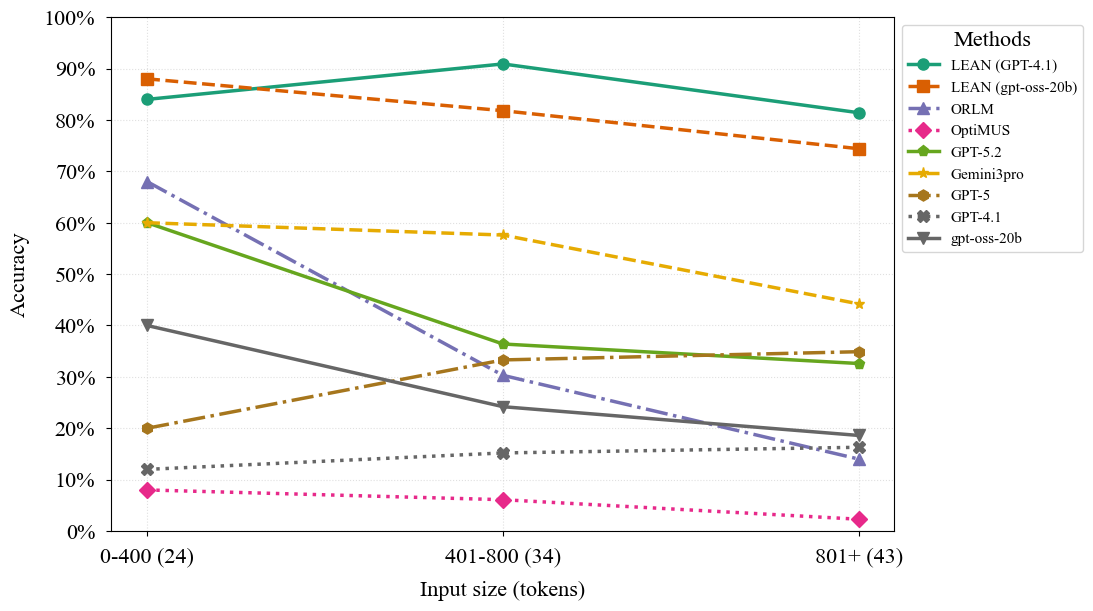}
    \caption{Modeling accuracy (EM) vs. input size}
    \label{fig:acuracy_input}
\end{figure}

\begin{figure}[h]
    \centering
    \includegraphics[width=0.9\linewidth]{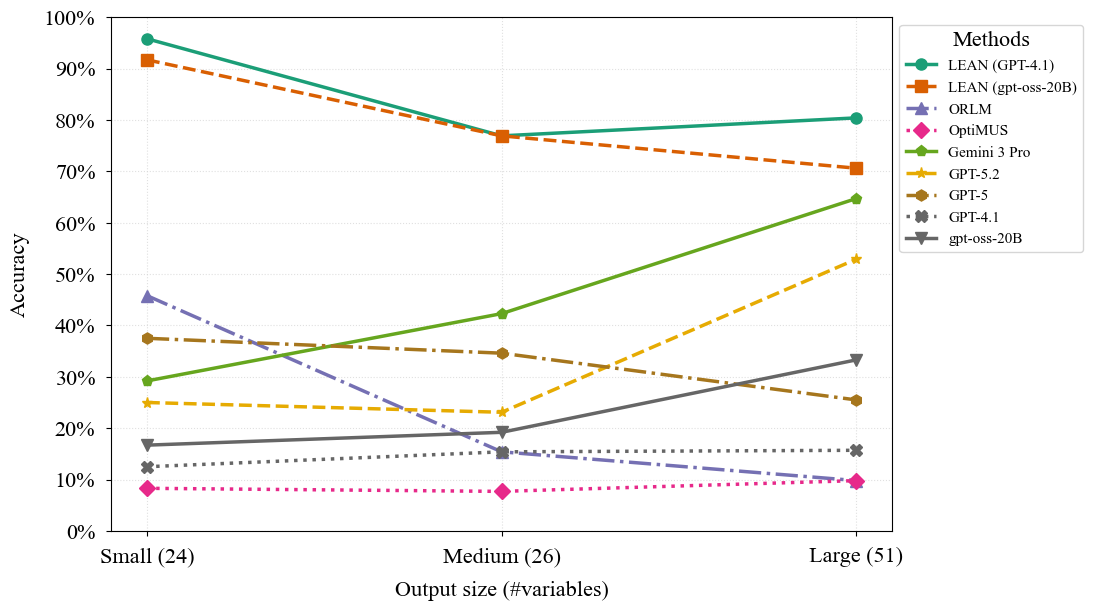}
    \caption{Modeling accuracy (EM)  vs. output size}
    \label{fig:accuracy_output}
\end{figure}

By inspecting the relationship between modeling accuracy (EM) and length of input (see Figure \ref{fig:acuracy_input}), we find that one of the key reasons for the excellent performance of \agent comes from its capability to sustain longer inputs. We can also find from Figure \ref{fig:accuracy_output} that \agent's performance drops as the number of variables in the output increases. This is expected because as the number of variables increases, the formulation's complexity also increases, which leaves \agent a more challenging reasoning task (e.g., figure out the placement of more coefficients). But somewhat surprisingly, several methods' performance can increase as the number of variables of the output increases. This suggests 1) similar to other reasoning tasks, the difficulty in large-scale optimization modeling may also come from the size of the input; 2) there might be better ways to handle the placement of coefficients, which can further improve the performance of \agent as well.

Table \ref{tab:by-family-transposed} also provides the modeling accuracy based on the resulted optimal values. By comparing against Table \ref{tab:by-family-transposed_em}, we find that it is indeed possible for a method to generate an incorrect formulation that can result in a correct optimal value. We provide several examples in Section \ref{sec:em_obj} of the appendix. But as we can see, the discrepancy between the two metrics are not so significant for most of the methods considered. This confirms that modeling accuracy can indeed be viewed as a good proxy for modeling accuracy (EM).

\begin{table}[ht]
\centering
\caption{Performance comparison of models on benchmarks. Results labeled with ``*" are acquired from \cite{chen2025sirl}. The best performing model in each column is boldfaced and the second best is underlined}
\small
\label{tab:method-accuracy}
\setlength{\tabcolsep}{5pt}
\renewcommand{\arraystretch}{1.5}
\begin{tabular}{lcccc}
\toprule
\textbf{Model} & \textbf{NL4OPT}&\textbf{MAMO Easy}&\textbf{MAMO Complex}&\textbf{IndustryOR} \\
\midrule
ORLM & 85.7\%* & 82.3\%* & 37.4\%* & 24.0\%* \\
OptiMUS & 78.8\%* & 77.2\%* & 43.6\%* & 31.0\%* \\
\midrule
Gemini 3 Pro & 87.3\% & 90.3\%& \underline{70.4\%} & \textbf{65.0\%} \\
GPT-5.2 & 78.8\%& \underline{92.8\%}& 67.5\%& 56.0\% \\
GPT-5 & 78.8\%& 92.3\%& 54.4\%& 58.0\% \\
GPT-4.1 & 72.4\%& 89.4\%& 56.8\%& 54.0\% \\
gpt-oss-20B & 50.2\% & 68.9\%& 29.1\%& 42.0\% \\
\midrule
\agent (GPT-4.1) & \textbf{94.7\%}& \textbf{93.5\%}& \textbf{71.4\%} & \textbf{65.0\%} \\
\agent (gpt-oss-20B) & \underline{93.5\%}& 91.4\%& 56.2\%& \underline{59.0\%} \\
\bottomrule
\end{tabular}
\vspace{1ex}\small
\end{table}

\vspace{2mm}
\noindent\textbf{Results on Small-Scale Benchmarks:} Table~\ref{tab:method-accuracy} summarizes the results on the small-scale benchmarks. Comparing \agent (GPT-4.1) and \agent (gpt-oss-20B) against the base models, we can again see that \agent can bring substantial improvements in modeling capabilities. In terms of absolute modeling accuracies, \agent (GPT-4.1) performs the best in all the benchmarks; while Gemini 3 Pro makes it a tie in IndustryOR. \agent (gpt-oss-20B) achieves the second best in NL4OPT and IndustryOR and GPT-5.2 achieves the second best in MAMO Easy. Compared with ORLM and OptiMUS, both \agent (GPT-4.1) and \agent (gpt-oss-20B) achieve higher performance across all benchmarks. Through these results, \agent demonstrate its ability to handle small-scale optimization modeling tasks.

In summary, the above results showcase \agent's modeling capability on a wide range of scenarios. Moreover, by cross-checking the performance of Gemini~3~Pro, GPT-5.2, GPT-5 on \testdata and the small-scale benchmarks, we observe that larger inputs tend to pose greater challenges for these models. This trend is particularly notable given that many instances in \testdata fall into commonly studied problem types (e.g., NRM, RA, and FLP).

\section{Case Study: Singapore Airlines Choice-Based Revenue Management} \label{sec:case} 

In this section, we demonstrate the practicality of \agent using a case study on Singapore Airlines' choice-based revenue management and we consider a generalized SBLP formulation \citep{gallego2015general}. We are interested in examining \agent's ability to earn revenue in this setting. In this case study, our model faces a more challenging task because it has to handle multiple datasets and the target model is more complicated than previous ones. In this section, we include Gemini 3 Pro and GPT-5.2 as our competitors because they are the best performing ones in the previous rounds of simulations. Because SBLP is a relatively less common model, we provide our competing methods with a single example that includes the problem description, the correct formulation, the corresponding Python code, and the original paper by \cite{gallego2015general}. 

Along the way, we also conduct an ablation study to assess the role of different components in performance improvement. For this purpose, we introduce two other methods, namely the tools-only approach and the workflow-only approach. The tools-only approach follows the same design of \agent, but the model generation agent is not provided with the workflow to demonstrate the processes of model generation. Instead, only the abstract form of the target model is given. In the workflow-only approach, the model generation agent can only use built-in functions rather than the customized tools to retrieve data from the input datasets. Our ablation study is only conducted for \agent (GPT-4.1). This is because \agent (gpt-oss-20B) would only perform worse during the ablation study.

\subsection{Background Information and Sales-Based Linear Programming}
Singapore Airlines is the flagship carrier of Singapore and is considered to be one of the best carriers worldwide. Singapore Airlines is ranked as a 5-star airline as well as ranked as the world's best airline by Skytrax five times \citep{wikipedia_SQ}. To improve the revenue of its flight network, Singapore Airlines applies large-scale NRM techniques. In this study, we apply \agent to help Singapore Airlines automate the optimization modeling and solution workflow for SIN, HKG and MNL three-city network. We only consider direct itinerary options as one-stop sales are marginal in reality. Specifically, we consider the (generalized) SBLP framework \citet{gallego2015general}, which can critically capture the important ``upsell demand" (i.e., customers may opt for a higher fare on the same flight if the lower-priced fare is unavailable) and ``cross-flight recapture" (i.e., customers may purchase tickets for a similar flight provided by a competing company when a preferred flight from Singapore Airlines isn't available) in airline revenue management.   

SBLP works under the General Attraction Model (GAM). In GAM, when a set of products \( S \subset N \) is offered, the demand for each product $h \in S$ is captured by the attraction value \(v_h = e^{\rho u_h}\), where \( u_h \) denotes product $h$'s utility and \( \rho > 0 \) is a parameter inversely related to the variance of the Gumbel distribution. At the same time, the utility of the substitute of product $h$ is represented by a shadow attraction value $w_h\in[0,v_h]$ for all $h\in N$. The probability of a customer selecting product \( h \in S \) or not making a purchase is thus
\begin{align}\label{eq:pi}
\pi_h(S) = \frac{v_h}{\tilde{v}_0 + \tilde{V}(S)} \quad \forall h \in S \qquad \text{and}\qquad
\pi_0(S) = 1 - \pi_S(S),
\end{align}
where $\tilde{v}_0 = v_0 + \sum_{h\in N} w_h$, $\tilde{v}_h = v_h - w_h$, $\tilde{V}(S) = \sum_{h \in S} \tilde{v}_h$ and $\pi_S(S) = \sum_{h \in S} \pi_h(S)$. We define customers interested in a flight with a specific origin-destination pair within a given day as a market segment $l$, and use $\mathcal{L}$ to denote all market segments. We let $N_l$ with $|N_l|=n_l$ be the set of flight departure times offered to market segment $l\in\mathcal L$. We also use $J$ to denote the set of fare types for each flight. We remark that in our setting, a product (i.e., the flight) is indexed by both its departure time and the fare type. Following the notations in \cite{gallego2015general}, the aggregate demand of segment $l$ is \( \Lambda_l \). We let $p_{lkj}~(l \in \mathcal{L}, k \in N_l, j \in J)$ be the unit price of a product with departure time $k$ and fare type $j$ in market segment $l$ and $A_{lkj}~(l \in \mathcal{L}, k \in N_l, j \in J)$ be the resource consumption of a product with departure time $k$ and fare type $j$ in market segment $l$.  Using \( x_{lkj} ~(l \in \mathcal{L}, k \in N_l, j \in J)\) as the decision variable that denotes the sales of a product with departure time \( k\) and fare type $j $  in market segment \( l \), the SBLP formulation is,
 \begin{align}
 \label{eq:sblp}
\text{Maximize} \quad & \sum_{l \in \mathcal{L}} \sum_{k \in N_l} \sum_{j \in J} p_{lkj} x_{lkj} \\
\nonumber\text{Subject to} \quad 
&  \sum_{j\in J} A_{lkj} x_{lkj} \leq c, \quad \forall l \in \mathcal{L}, ~\forall k \in N_l \quad \text{(capacity constraint)} \\
\nonumber& \frac{\tilde{v}_{l0}}{v_{l0}} x_{l0} + \sum_{k \in N_l} \sum_{j \in J} \frac{\tilde{v}_{lkj}}{v_{lkj}} x_{lkj} = \Lambda_l,\quad \forall l \in \mathcal{L} \quad \text{(balance constraint)} \\
\nonumber& \frac{x_{lkj}}{v_{lkj}} - \frac{x_{l0}}{v_{l0}} \leq 0,  \quad \forall l \in \mathcal{L}, ~\forall k \in N_l, ~\forall j \in J \quad \text{(scale constraint)} \\
\nonumber& x_{lkj} \geq 0, \quad \forall l \in \mathcal{L}, ~ \forall k \in N_l, ~\forall j \in J \quad \text{(non-negativity constraint)}
\end{align}
SBLP aims at providing the optimal solution for NRM under the GAM by optimizing sales of each product across different market segments. Capacity constraints limit the total allocations (i.e., the total number of seats available). Balance constraints ensure that the total (re-scaled) sales match total demand. Together with the scale constraints, they recover all the essential characteristics of GAM in a compact manner.

The canonical SBLP formulation can help Singapore Airlines find out the optimal fare type capacity allocation (i.e., the number of seats offered for a fare type) for different flights. In this case study, we consider a duopoly setting where substitute products are all offered by a single competing airline. Due to the incorporation of upsell demand and cross-flight recapture, we assume that there is no additional demand spillover across segments. However, we emphasize that this assumption is only for ease of construction of problem instances in our testing dataset (to be described in the forthcoming Section \ref{sec:air-nrm}). In fact, due to its generality, it is evident that \agent can easily accommodate optimization formulations that generalize SBLP and/or relax this assumption. For instance, in the forthcoming sections, we also consider variants of the SBLP which further take flight scheduling into account.

\subsection{Available Data and Testing Dataset \sindata} \label{sec:air-nrm}
A booking simulator trained by Singapore Airlines' real sales data that simulates each customer's choice behavior is provided, and its populated datasets are applied to fit the attractiveness of each product in the network. We simulate the whole 355 days of the selling season of flights that depart within 14 days. A real network planning schedule is simulated. The user's input datasets consist of four parts: ticket choice data/sales data generated by the simulator, demand data aggregated from the simulation, attraction values trained by the sales data, and adjusted attraction values.

Table \ref{tab:od_demand} demonstrates the average demand volume for each market segment $l$, where Avg Pax refers to the average number of both purchasing and non-purchasing passengers within a given market segment.

\begin{longtable}{>{\centering\arraybackslash}p{6cm} >{\centering\arraybackslash}p{6cm}} 
\caption{\textbf{OD pairs' demands}} \label{tab:od_demand} \\
\toprule
\textbf{OD} & \textbf{Avg Pax} \\
\midrule
\endfirsthead
\toprule
\textbf{OD} & \textbf{Avg Pax} \\
\midrule
\endhead
\bottomrule
\endfoot
(A,B) & 38965.86 \\
(A,C) & 4812.5 \\
(B,A) & 33210.71 \\
(C,A) & 4807.43 \\
\end{longtable}

The ticket choice dataset includes 19 unique combinations of origin-destination (OD) pairs and departure times, along with detailed information about each flight, including the key attributes shown below,
\begin{itemize}
    \item \emph{Origin-Destination pair (OD):} The departure and arrival cities of the flights. In this study, these cities could be Singapore, Hong Kong, and Manila, denoted A, B, and C, respectively. 
    \item \emph{Departure time:} The scheduled takeoff time of the flight. 
    \item \emph{Fare type:} The airfare subclass of the flight. Here, we consider `Eco-lite', which denotes the basic economy class, and `Eco-flexi', which denotes the flexible economy class.
    \item \emph{Average price (Avg Price):} The average price of a fare type for a flight.
\end{itemize}

We refer to Table \ref{tab:ticket_choice} for a sample of one flight. Table \ref{tab:attraction_values} exhibits the attraction value of each flight ticket \( v_{lk} \) trained from the simulated sales data, and Table \ref{tab:shadow_attraction_values} shows the shadow attraction value $w_{lk}$ for different $l$'s and $k$'s. As a prerequisite, we estimate the parameters \( v_{lk} \), \( w_{lk} \) and \(\Lambda_l\) of the generalized attraction model (GAM) by employing the expectation maximization (EM) method developed in \cite{vulcano2012estimating,gallego2015general}. 

\begin{longtable}{>{\centering\arraybackslash}p{3cm} >{\centering\arraybackslash}p{3cm} >{\centering\arraybackslash}p{4cm} >{\centering\arraybackslash}p{3cm}} 
\caption{\textbf{Sample ticket choice}} \label{tab:ticket_choice} \\
\toprule
\textbf{OD} & \textbf{Departure Time} & \textbf{Fare type} & \textbf{Avg Price} \\
\midrule
\endfirsthead
\toprule
\textbf{OD} & \textbf{Departure Time} & \textbf{Fare type} & \textbf{Avg Price} \\
\midrule
\endhead
\bottomrule
\endfoot

(A,B) & 11:20 & Eco-flexi & 1140.3  \\
(A,B) & 11:20 & Eco-lite  & 429.26    \\
\end{longtable}

\begin{longtable}{>{\centering\arraybackslash}p{2cm} >{\centering\arraybackslash}p{4cm} >{\centering\arraybackslash}p{1cm} >{\centering\arraybackslash}p{4cm} >{\centering\arraybackslash}p{3cm}}
\caption{\textbf{Attraction values}} \label{tab:attraction_values} \\

\toprule
\textbf{OD Pairs} & \textbf{Eco-flexi* (12-6pm)} & $\cdots$ & \textbf{Eco-lite* (8am-12pm)} & \textbf{No Purchase} \\ 
\midrule
\endfirsthead

\toprule
\textbf{OD Pairs} & \textbf{Eco-flexi* (12-6pm)} & $\cdots$ & \textbf{Eco-lite* (8am-12pm)} & \textbf{No Purchase} \\ 
\midrule
\endhead

\bottomrule
\endfoot

\bottomrule
\endlastfoot

('A', 'B') & 0.063743 & $\cdots$ & 0.24677  & 1 \\ 
('A', 'C') & 0.395947 & $\cdots$ & 0.0625   & 1 \\ 
('B', 'A') & 0.177309 & $\cdots$ & 0.0625   & 1 \\ 
('C', 'A') & 0.089563 & $\cdots$ & 0.108023 & 1 \\ 

\end{longtable}

\begin{longtable}{>{\centering\arraybackslash}p{2cm} >{\centering\arraybackslash}p{4cm} >{\centering\arraybackslash}p{1cm} >{\centering\arraybackslash}p{4cm} >{\centering\arraybackslash}p{4cm}}
\caption{\textbf{Shadow attraction values}} \label{tab:shadow_attraction_values} \\

\toprule
\textbf{OD Pairs} & \textbf{Eco-flexi* (12-6pm)} & $\cdots$ & \textbf{Eco-lite* (10pm-8am)} & \textbf{Eco-lite* (8am-12pm)} \\ 
\midrule
\endfirsthead

\toprule
\textbf{OD Pairs} & \textbf{Eco-flexi* (12-6pm)} & $\cdots$ & \textbf{Eco-lite* (10pm-8am)} & \textbf{Eco-lite* (8am-12pm)} \\ 
\midrule
\endhead

\bottomrule
\endfoot

\bottomrule
\endlastfoot

('A', 'B') & 0 & $\cdots$ & 0.104684 & 0 \\ 
('A', 'C') & 0.197974 & $\cdots$ & 0 & 0 \\ 
('B', 'A') & 0 & $\cdots$ & 0.044375 & 0.044375 \\ 
('C', 'A') & 0.026869 & $\cdots$ & 0.016357 & 0.011883 \\ 

\end{longtable}

For this case study, we create a comprehensive testing dataset \sindata. \sindata consists of two parts, \sindataca, a dataset of 15 medium-sized problem instances on fare type capacity allocation, and \sindatanp, a dataset of 21 medium-sized problem instances on network planning (i.e., joint fare type capacity allocation and flight scheduling). Each problem instance includes a query, i.e., problem description and four input datasets as described above, and its label, i.e., the (generalized) SBLP optimization formulation and the optimal value.  

In \sindataca, one is asked to formulate the SBLP \eqref{eq:sblp} that optimizes fare type capacity for all flights chosen by the query (see Section \ref{sec:case_supp} for an example query). In \sindatanp, it is further required that the optimization formulation should pick a subset of flights (e.g., 3 flights among all those available) that maximize revenue and ensure the flow conservation constraint holds. Below is an example query in \sindatanp, 
\begin{tcolorbox}[colback=white, colframe=boxgrey, coltitle=white, coltext=black,title=Query (Joint Fare Type Capacity Allocation and Flight Scheduling),breakable]
\small

Based on flight ticket options provided in the table, along with their average passengers (Avg Pax), average prices (Avg Price), and capacity coefficients (Flex Cpy Coef), considering that each Eco-flexi ticket consumes 3.1 units of flight capacity and each Eco-lite ticket consumes 1 unit of capacity, while enforcing flow conservation constraints at each airport for a long-term planning model, develop a Sales-Based Linear Programming (SBLP) model. The goal of this model is to recommend the optimal 9 flights that maximize total ticket sale revenue, specifically among flights where the origin-destination pairs are: (OD = (`B', `A') AND Departure Time=`06:25'), (OD = (`A', `B') AND Departure Time=`06:40'), (OD = (`C', `A') AND Departure Time=`07:40'), (OD = (`A', `B') AND Departure Time=`07:55'), (OD = (`C', `A') AND Departure Time=`08:15'), (OD = (`B', `A') AND Departure Time=`09:05'), (OD = (`A', `C') AND Departure Time=`09:45'), (OD = (`A', `B') AND Departure Time=`11:20'), (OD = (`B', `A') AND Departure Time=`15:40'), (OD = (`C', `A') AND Departure Time=`16:55'), (OD = (`A', `B') AND Departure Time=`17:05'), (OD = (`A', `C') AND Departure Time=`17:25'), (OD = (`C', `A') AND Departure Time=`18:30'), (OD = (`B', `A') AND Departure Time=`18:50'), (OD = (`A', `C') AND Departure Time=`19:05'),(OD = (`A', `B') AND Departure Time=`19:10'), (OD = (`B', `A') AND Departure Time=`20:25').

\end{tcolorbox}
For this query, its label is a generalized SBLP as below (we omit the detailed values of the coefficients for conciseness),

\begin{tcolorbox}[colback=white, colframe=boxgrey, coltitle=white, coltext=black,title=Label $m$, breakable]
\small\vspace{-5mm}
 \begin{align}
 \text{Maximize} \quad & \sum_{l \in \mathcal{L}} \sum_{k \in N_l} \sum_{j \in J} p_{lkj} x_{lkj} \label{eq:sblp_np}\\
\nonumber\text{Subject to} \quad 
&  \sum_{j\in J} A_{lkj} x_{lkj} \leq c, \quad \forall l \in \mathcal{L}, ~\forall k \in N_l \quad \text{(capacity constraint)} \\
\nonumber& \frac{\tilde{v}_{l0}}{v_{l0}} x_{l0} + \sum_{k \in N_l} \sum_{j \in J} \frac{\tilde{v}_{lkj}}{v_{lkj}} x_{lkj} = \Lambda_l,\quad \forall l \in \mathcal{L} \quad \text{(balance constraint)} \\
\nonumber& \frac{x_
{lkj}}{v_{lkj}} - \frac{x_{l0}}{v_{l0}} \leq 0,\quad \forall l \in \mathcal{L}, ~\forall k \in N_l,  \forall j \in J \quad \text{(scale constraint)} \\
& x_{lkj} \leq M \cdot y_{lk}, \quad \forall k \in N_l,  ~\forall l \in \mathcal{L} \quad \text{(big $M$ constraint)} \label{eq:sblp_M_const}\\
& \sum_{l \in \mathcal{L}} \sum_{k \in N_l} y_{lk} \leq Z \quad \text{(cardinality constraint)} \label{eq:sblp_card_const}\\
& \sum_{l \in \sigma^+_o} \sum_{k \in N_l} y_{lk} = \sum_{l \in \sigma^-_o} \sum_{k \in N_l} y_{lk}, \quad \forall o \in O \quad \text{(flow conservation constraint)} \label{eq:sblp_flow_const}\\
\nonumber& x_{lkj} \geq 0,  \quad \forall l \in \mathcal{L},~\forall k \in N_l, ~\forall j \in J \quad \text{(non-negativity constraint)}\\
\nonumber& y_{lk} \in \{0,1\}, \quad \forall l \in \mathcal{L}, ~\forall k \in N_l \quad \text{(binary constraint)}
\end{align}
\end{tcolorbox}
Here, in the flow conservation constraint \eqref{eq:sblp_flow_const}, the set $O = \{A, B, C\}$ represents the three airport locations, with $\sigma^+_o$ and $\sigma^-_o$ denoting the sets of inbound and outbound flights 
for airport $o \in O$, respectively. This constraint makes sure the number of inbound and outbound flights of any location is equal. The cardinality constraint \eqref{eq:sblp_card_const} limits the number of flights one can choose. The big $M$ constraint \eqref{eq:sblp_M_const} ensures we only count the revenue from the chosen flights. By solving the optimization formulation \eqref{eq:sblp_np}, we could acquire the $Z$ most profitable flights together with the optimal fare type capacity allocation among all the provided flights. 

\begin{figure}[!h]  
    \centering
    \includegraphics[width=\linewidth]{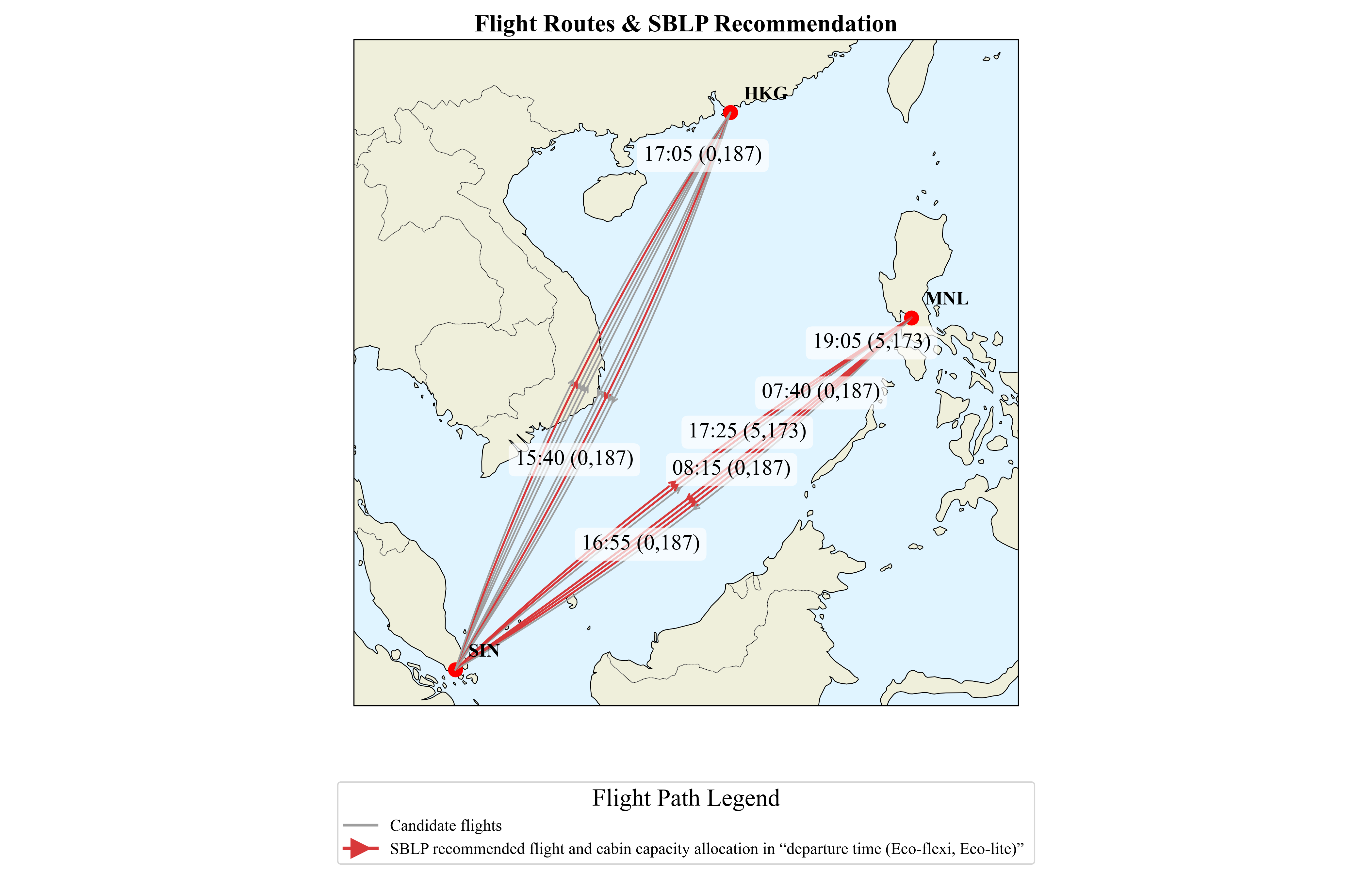}
    \caption{Flight routes \& the recommended flights from SBLP generated by \agent}
\label{fig:SBLP_result} 
\end{figure}
Among the 21 problem instances in \sindatanp, they impose the flow conservation constraints \eqref{eq:sblp_flow_const} to mimic a closed-loop network planning scenario. Based on the existing datasets, the problem size increases from 25 to a maximum of 61 variables. Figure \ref{fig:SBLP_result} shows the candidate flights and the flights recommended by the SBLP \eqref{eq:sblp_np} for the example query. From the results, Singapore Airlines is not only presented with the most profitable flights, but is also provided with the fare type capacity configuration that achieves the optimal revenue.

\subsection{Performance on Fare Type Capacity Allocation and Ablation Study}\label{sec:case_ca}
\begin{table}[!h]
\centering
\caption{Modeling accuracy of different approaches on \sindataca} 
\label{tab:accuracy-comparison}
{\begin{tabular}{lc}
\toprule
\textbf{Model}& \textbf{Modeling accuracy}\\ 

\midrule
Gemini 3 Pro & 80\%\\
GPT-5.2 & 26.7\% \\ 
\midrule
\agent (GPT-4.1) & 93.3\%   \\
\agent (gpt-oss-20B) & 86.7\%  \\
\agent (GPT-4.1, tools-only) & 20\% \\
\agent (GPT-4.1, workflow-only) & 0\% \\
\agent (GPT-4.1, base model-only) &0\%\\\bottomrule
\end{tabular}}
\end{table}

In this section, we report the performance of \agent on the fare type capacity allocation dataset \sindataca. Table \ref{tab:accuracy-comparison} presents the modeling accuracy of various approaches. In this table, \agent (GPT-4.1) and \agent (gpt-oss-20B) exhibit leading performance compared to the other methods. Between the competing methods, we can also see that Gemini 3 Pro has a much stronger performance when compared to GPT-5.2. 

GPT-4.1, the base model of \agent, cannot provide the correct model and code for any of the instances. Inspecting GPT-4.1's outputs, we find that it is only capable of generating some of the non-negativity constraints correctly, which do not require access to the datasets, but not the other constraints. This can be attributed to 1) not knowing it should retrieve information from the datasets; and 2) nor does it have the tools to retrieve relevant data from the input datasets. 

The tools-only approach has an modeling accuracy of 20\%. With access to the data-handling tools, it can retrieve relevant data, such as flight capacities, average passenger counts, ticket prices, and attraction values from the input datasets. However, this approach often struggles with interpreting the retrieved data and placing the coefficients correctly, especially when the problem size grows larger. We reason that this is because the model generation agent is not provided with a step-by-step guide on data usage. Note that in this use case, the query comes with multiple datasets; it is thus important for the agent to correctly interpret the retrieved data so that it knows which equation every piece of data belongs to.

The modeling accuracy of the workflow-only approach is 0\%. Although this approach can adeptly interpret the retrieved data, a critical drawback is that it can hardly extract all the required data even when the size of the problem instance is small. This is because it could only use Python built-in functions rather than the customized tools. This drastically degrades the capability in data handling. Therefore, it can only correctly generate the abstract model without correct and complete parameters. 

\subsection{Performance on Network Planning}\label{sec:case_np}
\begin{table}[!h]
\centering
\caption{Modeling accuracy of different approaches on \sindatanp} 
\label{tab:accuracy-comparison-from-image}
{\begin{tabular}{lc}
\toprule
\textbf{Model}& \textbf{Modeling accuracy}\\ 

\midrule
Gemini 3 Pro & 85.7\%\\
GPT-5.2 & 55.6\% \\ 
\midrule
\agent (GPT-4.1) & 92.1\%   \\
\agent (gpt-oss-20B) & 82.5\%  \\
\agent (GPT-4.1, tools-only) & 30.2\% \\
\agent (GPT-4.1, workflow-only) & 0\% \\
\agent (GPT-4.1, base model-only) &0\%\\\bottomrule
\end{tabular}}
\end{table}
We next turn to \sindatanp and examine the modeling accuracy and revenue loss incurred when deploying \agent to plan the airline network. In \sindatanp, each instance requires formulating and solving an optimization model that jointly determines the number of seats allocated to each fare class and selects a subset of flights to maximize revenue. The modeling accuracies of different methods are reported in Table~\ref{tab:accuracy-comparison-from-image}. We observe that \agent (GPT-4.1) continues to achieve the highest modeling accuracy, with Gemini~3~Pro performing second best. Moreover, \agent (gpt-oss-20B) attains modeling accuracy comparable to Gemini~3~Pro. Given that gpt-oss-20B has substantially fewer parameters than the other models considered, these results suggest that \agent can help elicit and steer the underlying model’s reasoning in this setting.

\begin{figure}[h]
    \centering
    \includegraphics[width=0.9\linewidth]{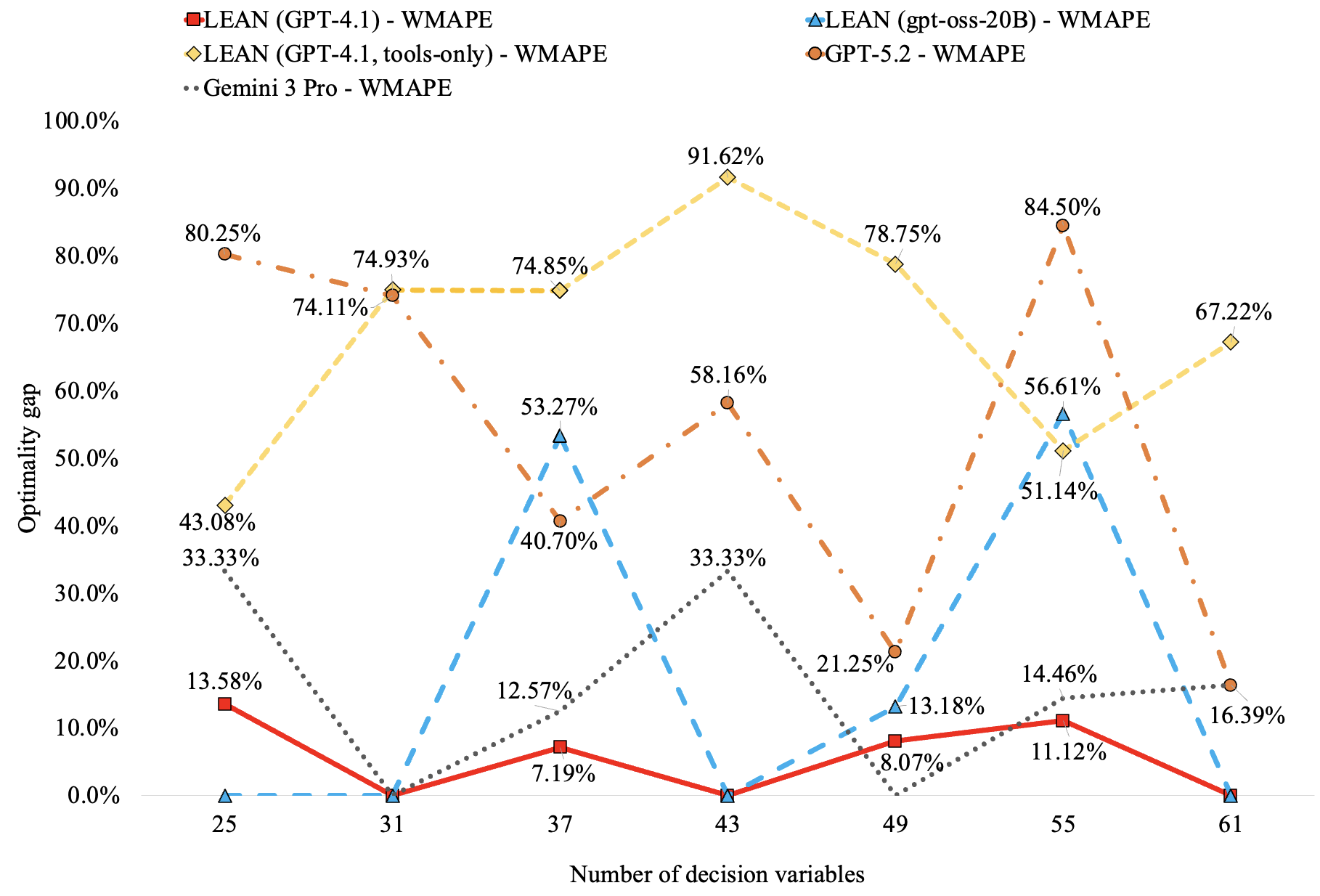}
    \caption{Performance comparison on \sindatanp}
    \label{fig:opt_gap_rag}
\end{figure}

To systematically assess the impact of execution error to revenue, we input the query of each instance in \sindatanp into the models, and then find the optimal solutions of the respective generated Python program codes. We use the notion of optimality gap measured by weighted mean absolute percentage error (WMAPE) to evaluate the performance. In particular, denoting $\text{OPT}_{i}$ as the ground-truth optimal revenue and $\text{REV}_{i}$ as the revenue obtained by following the solution resulting from the generated formulation of instance $i$\footnote{If a method fails to generate a complete and feasible formulation, the corresponding revenue is 0.}, WMAPE are defined as
\begin{align*}
&\text{WMAPE}=\frac{\sum_{i=1}^n(\text{OPT}_i-\text{REV}_{i})}{\sum_{i=1}^n\text{OPT}_i},\qquad 
\end{align*}
where $n$ is the number of problem instances with the same number of decision variables considered. Here, for stability, we average the results from 3 iterations.

Figure~\ref{fig:opt_gap_rag} reports the optimality gaps of different methods on \sindatanp as the problem size varies. As shown, the optimality gap of \agent (GPT-4.1) remains consistently below 14\%, whereas the tools-only counterpart typically yields gaps exceeding 55\% for most instances. The optimality gap of \agent (gpt-oss-20B) fluctuates but stays below 24\% in all but two cases. This trend is similar to Gemini~3~Pro, whose optimality gap remains below 17\% except for two cases. In contrast, GPT-5.2 often exhibit optimality gaps above 40\%, and in some cases exceed 80\%.

Overall, these results suggest that \agent can achieve strong revenue performance under the SBLP framework and its variants, and that its performance remains relatively stable as problem size and complexity increase.

\section{Concluding Remarks} \label{sec:conclusion}
Increasing adoption of large-scale optimization techniques raises the need for automated optimization model formulation and solution. While meaningful attempts are made, they focus exclusively on small-scale problems where data can be embedded into the problem description and the target optimization model is small. Because LLMs may struggle with long-form reasoning, it remains unclear if we can apply LLMs to formulate large-scale optimization models. In this work, we provide an affirmative answer by developing \agent. 

\agent uses an agentic workflow construction framework that can alleviate the LLMs' burden associated with planning and data handling, and allows it to focus on the parts that cannot be easily standardized. Our plug-and-play design also greatly enhances the practicality and portability. Our results offer several novel insights and reveal several interesting future directions. 
\begin{itemize} 
\item \textbf{Generalizability of the Agentic Workflow Approach:} Although our main focus is optimization modeling, the idea of agentic workflow may extend beyond it. In many domains, valuable and structured expertise has accumulated over time. While such expertise is rarely one-size-fits-all, LLMs provide a powerful mechanism for making these practices more customizable and flexible. By translating domain expertise into (the skeleton of the) workflows while allowing LLMs to handle inherently unstructured components, we may be able to improve both effectiveness and efficiency across a broad range of applications.

\item \textbf{Stimulate LLM Capabilities:} Next, our results suggest that older-generation models and less powerful open-source models may still have considerable potential for solving challenging problems, provided that their reasoning is appropriately elicited and guided. Given that such models often offer lower cost and/or privacy advantages, developing systematic methods for steering them toward competitive performance warrants further study.

\item \textbf{Enforceable and Controllable Structure:} In our work, we introduce structure and workflow primarily through semantic instructions. However, it has been widely-recognized that LLM's behavior is inherently probabilistic and can make the system's behavior unstable. It is thus interesting to explore how to design more enforceable and controllable mechanisms to further improve performance and reliability systematically. One promising direction is to follow the assured autonomy framework proposed in \cite{dai2025assured}, which ensures formal structure, constraints satisfaction, accountability, and tail-risk control through an operations research lens. 

\end{itemize}

\ACKNOWLEDGMENT{The authors would like to thank seminar and conference participants at the INFORMS MSOM Conference, INFORMS International Conference, POMS-HK International Conference, Chinese University of Hong Kong, Hong Kong Polytechnic University, Hong Kong University,  Cornell University, Georgetown University, New York University, University of Maryland, and University of Washington for feedback and suggestions.}
\SingleSpacedXI
\bibliographystyle{ormsv080}
\bibliography{sample}

@article{chen2025sirl,
    title = {Solver-Informed RL: Grounding Large Language
Models for Authentic Optimization Modeling},
    author = {Yitian Chen and Jingfan Xia and Siyu Shao and DongDong Ge and Yinyu Ye},
    journal = {Proceedings of the 39th Conference on Neural Information Processing Systems},
    year = {2025},
}

@article{dai2025assured,
    title = {Assured Autonomy: How Operations Research Powers and Orchestrates Generative AI Systems},
    author = {Tinglong Dai and David Simchi-Levi and Michelle Xiao Wu and Yao Xie},
    journal = {arXiv:2512.23978 [cs.LG]},
    year = {2025},
}

@article{levi2025democratizing,
    title = {Democratizing Optimization with Generative AI},
    author = {David Simchi-Levi and Tinglong Dai and Ishai Menache and Michelle Xiao Wu},
    journal = {SSRN 5511218},
    year = {2025},
}

@article{cer2017semeval,
    title = {{S}em{E}val-2017 Task 1: Semantic Textual Similarity Multilingual and Crosslingual Focused Evaluation},
    author = {Cer, Daniel  and
      Diab, Mona  and
      Agirre, Eneko  and
      Lopez-Gazpio, I{\~n}igo  and
      Specia, Lucia},
    journal = {Proceedings of the 11th International Workshop on Semantic Evaluation ({S}em{E}val-2017)},
    year = {2017},
}

@inproceedings{agirre2013sem,
    title = {*{SEM} 2013 shared task: Semantic Textual Similarity},
    author = {Agirre, Eneko  and
      Cer, Daniel  and
      Diab, Mona  and
      Gonzalez-Agirre, Aitor  and
      Guo, Weiwei},
    booktitle = {Second Joint Conference on Lexical and Computational Semantics (*{SEM}), Volume 1: Proceedings of the Main Conference and the Shared Task: Semantic Textual Similarity},
    year = {2013},
}

@article{reimers2019sentence-bert,
  title = {Sentence-BERT: Sentence Embeddings using Siamese BERT-Networks},
  author = {Reimers, Nils and Gurevych, Iryna},
  journal = {Proceedings of the 2019 Conference on Empirical Methods in Natural Language Processing},
  year = {2019},
  publisher = {Association for Computational Linguistics},
}

@article{DBLP:conf/naacl/DevlinCLT19,
  author       = {Jacob Devlin and
                  Ming{-}Wei Chang and
                  Kenton Lee and
                  Kristina Toutanova},
  title        = {{BERT:} Pre-training of Deep Bidirectional Transformers for Language
                  Understanding},
  journal    = {Proceedings of the 2019 Conference of the North American Chapter of
                  the Association for Computational Linguistics: Human Language Technologies,
                  {NAACL-HLT}},
  pages        = {4171--4186},
  year         = {2019},
}

@article{levy2024same,
  title =	 {Same Task, More Tokens: the Impact of Input Length on
the Reasoning Performance of Large Language Models},
  author = {Mosh Levy and Alon Jacoby and Yoav Goldberg},
  journal =	 {Proceedings of the 62nd Annual Meeting of the Association for Computational Linguistics (Volume 1: Long Papers)},
  year =	 {2024}
}

@article{li2025context,
  title =	 {Context Length Alone Hurts LLM Performance Despite Perfect Retrieval},
  author = {Yufeng Du and Minyang Tian and Srikanth Ronanki and Subendhu Rongali and Sravan Babu Bodapati and Aram Galstyan and Azton Wells and Roy Schwartz and Eliu A Huerta and Hao Peng},
  journal =	 {Findings of the Association for Computational Linguistics: EMNLP},
  year =	 {2025}
}

@article{li2025long,
  title =	 {Long-context LLMs Struggle with Long In-context Learning},
  author = {Tianle Li and Ge Zhang and Quy Duc Do and Xiang Yue and Wenhu Chen},
  journal =	 {Transactions on Machine Learning Research},
  year =	 {2025}
}

@article{wasserkrug2025decision,
  title =	 {Enhancing decision making through the integration of large language models and operations research optimization},
  author =	 {Segev Wasserkrug and Leonard Boussioux and Dick Den Hertog and Farzaneh Mirzazadeh and  S. Ilker Birbil and Jannis Kurtz and Donato Maragno},
  journal =	 {Proceedings of the 39th AAAI Conference on Artificial Intelligence (AAAI)},
  year =	 {2025}
}

@article{zhou2025dp,
  title =	 {Auto-Formulating Dynamic Programming Problems with Large Language Models},
  author =	 {Chenyu Zhou and Jingyuan Yang and Linwei Xin and Yitian Chen and Ziyan He and Dongdong Ge},
  journal =	 {SSRN 5150161},
  year =	 {2025}
}

@book{bertsimas1997linear,
  author = {Dimitris Bertsimas and John N. Tsitsiklis },
  title = {Introduction to Linear Optimization },
  publisher = {Athena Scientific},
  year = {1997},
}

@misc{kaggle,
  author       = {Kaggle},
  year         = {2025},
  url = {https://www.kaggle.com/},
  note         = {Accessed: 2025-03-06}
}

@article{yao2023react,
  title =	 {{ReAct}: Synergizing Reasoning and Acting in Language Models},
  author =	 {Shunyu Yao and Jeffrey Zhao and Dian Yu and Nan Du and Izhak Shafran and Karthik R Narasimhan and Yuan Cao},
  journal =	 {Proceedings of the 11th International Conference on Learning Representations (ICLR)},
  year =	 {2023}
}

@article{brand2023market,
  title =	 {Using LLMs for Market Research},
  author =	 {James Brand and Ayelet Israeli and Donald Ngwe},
  journal =	 {SSRN 4395751},
  year =	 {2023}
}

@article{li2024user,
    title = {User Experience Design Professionals' Perceptions of Generative Artificial Intelligence},
    author = {Jie Li and Hancheng Cao and Laura Lin and Youyang Hou and Ruihao Zhu and Abdallah El Ali},
    journal = {Proceedings of the 2024 CHI Conference on Human Factors in Computing Systems},
    year = {2024},
    pages = {1--18},
}

@article{ramamonjison2022augmenting,
    title = {Augmenting Operations Research with Auto-Formulation of Optimization Models From Problem Descriptions},
    author = {Rindra Ramamonjison and      Haley Li and
      Timothy Yu and Shiqi
      He and Vishnu Rengan and
      Amin Banitalebi-dehkordi and
      Zirui Zhou and
      Yong Zhang},
    journal = {Proceedings of the 2022 Conference on Empirical Methods in Natural Language Processing: Industry Track},
    year = {2022},
    pages = {29--62},
}

@article{chen2021item,
  title={Item aggregation and column generation for online-retail inventory placement},
  author={Chen, Annie I and Graves, Stephen C},
  journal={Manufacturing \& Service Operations Management},
  volume={23},
  number={5},
  pages={1062--1076},
  year={2021},
  publisher={INFORMS}
}

@inproceedings{ramamonjison2023nl4opt,
  title={Nl4opt competition: Formulating optimization problems based on their natural language descriptions},
  author={Ramamonjison, Rindranirina and Yu, Timothy and Li, Raymond and Li, Haley and Carenini, Giuseppe and Ghaddar, Bissan and He, Shiqi and Mostajabdaveh, Mahdi and Banitalebi-Dehkordi, Amin and Zhou, Zirui and others},
  booktitle={NeurIPS 2022 Competition Track},
  pages={189--203},
  year={2023},
  organization={PMLR}
}

@inproceedings{xiao2023chain,
  title={Chain-of-Experts: When LLMs Meet Complex Operations Research Problems},
  author={Xiao, Ziyang and Zhang, Dongxiang and Wu, Yangjun and Xu, Lilin and Wang, Yuan Jessica and Han, Xiongwei and Fu, Xiaojin and Zhong, Tao and Zeng, Jia and Song, Mingli and others},
  booktitle={The Twelfth International Conference on Learning Representations},
  year={2023},
  pages = {},
}

@article{ahmaditeshnizi2024optimus,
  title={OptiMUS: Scalable Optimization Modeling with (MI) LP Solvers and Large Language Models},
  author={AhmadiTeshnizi, Ali and Gao, Wenzhi and Udell, Madeleine},
  journal={arXiv preprint arXiv:2402.10172},
  year={2024}
}

@article{li2023synthesizing,
  title={Synthesizing mixed-integer linear programming models from natural language descriptions},
  author={Li, Qingyang and Zhang, Lele and Mak-Hau, Vicky},
  journal={arXiv preprint arXiv:2311.15271},
  year={2023}
}

@article{chen2024diagnosing,
  title={Diagnosing infeasible optimization problems using large language models},
  author={Chen, Hao and Constante-Flores, Gonzalo E and Li, Can},
  journal={INFOR: Information Systems and Operational Research},
  volume={62},
  number={4},
  pages={573--587},
  year={2024},
  publisher={Taylor \& Francis}
}

@article{huang_orlm_2024,
author = {Huang, Chenyu and Tang, Zhengyang and Hu, Shixi and Jiang, Ruoqing and Zheng, Xin and Ge, Dongdong and Wang, Benyou and Wang, Zizhuo},
title = {ORLM: A Customizable Framework in Training Large Models for Automated Optimization Modeling},
journal = {Operations Research},
year = {2025},
note = {Article in Advance}
}

@article{ma2024llamoco,
  title={LLaMoCo: Instruction Tuning of Large Language Models for Optimization Code Generation},
  author={Ma, Zeyuan and Guo, Hongshu and Chen, Jiacheng and Peng, Guojun and Cao, Zhiguang and Ma, Yining and Gong, Yue-Jiao},
  journal={arXiv preprint arXiv:2403.01131},
  year={2024}
}

@article{huang2024mamo,
  title={Mamo: a Mathematical Modeling Benchmark with Solvers},
  author={Huang, Xuhan and Shen, Qingning and Hu, Yan and Gao, Anningzhe and Wang, Benyou},
  journal={arXiv preprint arXiv:2405.13144},
  year={2024}
}

@article{jiang2024llmopt,
  title={LLMOPT: Learning to Define and Solve General Optimization Problems from Scratch},
  author={Jiang, Caigao and Shu, Xiang and Qian, Hong and Lu, Xingyu and Zhou, Jun and Zhou, Aimin and Yu, Yang},
  journal={Proceedings of the 13th International Conference on Learning Representations (ICLR)},
  year={2025}
}

@article{astorga2024autoformulation,
  title={Autoformulation of Mathematical Optimization Models Using LLMs},
  author={Astorga, Nicol{\'a}s and Liu, Tennison and Xiao, Yuanzhang and van der Schaar, Mihaela},
  journal={arXiv preprint arXiv:2411.01679},
  year={2024}
}

@article{yang2024optibench,
  title={OptiBench Meets ReSocratic: Measure and Improve LLMs for Optimization Modeling},
  author={Yang, Zhicheng and Wang, Yiwei and Huang, Yinya and Guo, Zhijiang and Shi, Wei and Han, Xiongwei and Feng, Liang and Song, Linqi and Liang, Xiaodan and Tang, Jing},
  journal={arXiv preprint arXiv:2407.09887},
  year={2024}
}

@article{kangJDComImproves2022,
  title   = {{JD.Com improves delivery networks by a multiperiod facility location model}},
  author  = {Kang, Ningxuan and Shen, Hao and Xu, Ye},
  year    = {2022},
  month   = mar,
  journal = {INFORMS Journal on Applied Analytics},
  volume  = {52},
  number  = {2},
  pages   = {133--148},
  issn    = {2644-0865, 2644-0873},
  doi     = {10.1287/inte.2021.1077},
  urldate = {2023-07-16},
  langid  = {english}
}

@article{dang2021network,
  title={Network mode optimization for the DHL supply chain},
  author={Dang, Yibo and Singh, Manjeet and Allen, Theodore T},
  journal={INFORMS Journal on Applied Analytics},
  volume={51},
  number={3},
  pages={179--199},
  year={2021},
  publisher={INFORMS}
}

@article{deng2023alibaba,
  title={Alibaba realizes millions in cost savings through integrated demand forecasting, inventory management, price optimization, and product recommendations},
  author={Deng, Yuming and Zhang, Xinhui and Wang, Tong and Wang, Lin and Zhang, Yidong and Wang, Xiaoqing and Zhao, Su and Qi, Yunwei and Yang, Guangyao and Peng, Xuezheng},
  journal={INFORMS Journal on Applied Analytics},
  volume={53},
  number={1},
  pages={32--46},
  year={2023},
  publisher={INFORMS}
}

@article{gallego2015general,
  title={A general attraction model and sales-based linear program for network revenue management under customer choice},
  author={Gallego, Guillermo and Ratliff, Richard and Shebalov, Sergey},
  journal={Operations Research},
  volume={63},
  number={1},
  pages={212--232},
  year={2015},
  publisher={INFORMS}
}

@article{wei2022chain,
  title={Chain-of-thought prompting elicits reasoning in large language models},
  author={Wei, Jason and Wang, Xuezhi and Schuurmans, Dale and Bosma, Maarten and Xia, Fei and Chi, Ed and Le, Quoc V and Zhou, Denny and others},
  journal={Advances in neural information processing systems},
  volume={35},
  pages={24824--24837},
  year={2022}
}

@article{yao2024tree,
  title={Tree of thoughts: Deliberate problem solving with large language models},
  author={Yao, Shunyu and Yu, Dian and Zhao, Jeffrey and Shafran, Izhak and Griffiths, Tom and Cao, Yuan and Narasimhan, Karthik},
  journal={Advances in Neural Information Processing Systems},
  volume={36},
  year={2024}
}

@article{vulcano2012estimating,
  title={Estimating primary demand for substitutable products from sales transaction data},
  author={Vulcano, Gustavo and Van Ryzin, Garrett and Ratliff, Richard},
  journal={Operations Research},
  volume={60},
  number={2},
  pages={313--334},
  year={2012},
  publisher={INFORMS}
}

@inproceedings{besta2024graph,
  title={Graph of thoughts: Solving elaborate problems with large language models},
  author={Besta, Maciej and Blach, Nils and Kubicek, Ales and Gerstenberger, Robert and Podstawski, Michal and Gianinazzi, Lukas and Gajda, Joanna and Lehmann, Tomasz and Niewiadomski, Hubert and Nyczyk, Piotr and others},
  booktitle={Proceedings of the AAAI Conference on Artificial Intelligence},
  volume={38},
  number={16},
  pages={17682--17690},
  year={2024}
}

@article{zheng2023progressive,
  title={Progressive-hint prompting improves reasoning in large language models},
  author={Zheng, Chuanyang and Liu, Zhengying and Xie, Enze and Li, Zhenguo and Li, Yu},
  journal={arXiv preprint arXiv:2304.09797},
  year={2023}
}

@article{gao2023retrieval,
  title={Retrieval-augmented generation for large language models: A survey},
  author={Gao, Yunfan and Xiong, Yun and Gao, Xinyu and Jia, Kangxiang and Pan, Jinliu and Bi, Yuxi and Dai, Yi and Sun, Jiawei and Wang, Haofen},
  journal={arXiv preprint arXiv:2312.10997},
  year={2023}
}

@misc{wikipedia_SQ,
  author       = {Skytrax},
  title        = {{Singapore Airlines}},
  year         = {2025},
  howpublished = {\url{https://skytraxratings.com/airlines/singapore-airlines-rating}},
  note         = {Accessed: 2025-03-28},
}

@book{talluri2006theory,
  title={The theory and practice of revenue management},
  author={Talluri, Kalyan T and Van Ryzin, Garrett J},
  volume={68},
  year={2006},
  publisher={Springer Science \& Business Media}
}

@article{daskin1997network,
  title={Network and discrete location: models, algorithms and applications},
  author={Daskin, Mark},
  journal={Journal of the Operational Research Society},
  volume={48},
  number={7},
  pages={763--764},
  year={1997},
  publisher={Taylor \& Francis}
}

@book{winston2004operations,
  title={Operations research: applications and algorithm},
  author={Winston, Wayne L},
  year={2004},
  publisher={Thomson Learning, Inc.}
}

@inproceedings{pinedo1992scheduling,
  title={Scheduling: theory, algorithms and systems development},
  author={Pinedo, Michael and Hadavi, Khosrow},
  booktitle={Operations research proceedings 1991: Papers of the 20th annual meeting/vortr{\"a}ge der 20. Jahrestagung},
  pages={35--42},
  year={1992},
  organization={Springer}
}

@incollection{pisinger1998knapsack,
  title={Knapsack problems},
  author={Pisinger, David and Toth, Paolo},
  booktitle={Handbook of Combinatorial Optimization: Volume1--3},
  pages={299--428},
  year={1998},
  publisher={Springer}
}

@inproceedings{DBLP:conf/iclr/HumeauSLW20,
author       = {Samuel Humeau and
Kurt Shuster and
Marie{-}Anne Lachaux and
Jason Weston},
title        = {Poly-encoders: Architectures and Pre-training Strategies for Fast
and Accurate Multi-sentence Scoring},
booktitle    = {8th International Conference on Learning Representations, {ICLR} 2020,
Addis Ababa, Ethiopia, April 26-30, 2020},
publisher    = {OpenReview.net},
year         = {2020},
url          = {https://openreview.net/forum?id=SkxgnnNFvH},
bibsource    = {dblp computer science bibliography, https://dblp.org}
}

@misc{gurobi,
  author = {{Gurobi Optimization, LLC}},
  title  = {{Gurobi optimizer reference manual}},
  year   = 2026,
  url    = {https://www.gurobi.com}
}
\newpage
\begin{appendices}{\large \noindent\textbf{Appendix}}
\OneAndAHalfSpacedXI

\section{Supplementary Details for \refdata and \testdata}
\subsection{Application Domains of the Problem Instances}\label{sec:large-scale}
In this section, we describe the application domains of the large-scale problem instances acquired from Kaggle in Tables \ref{tab:categorization} - \ref{table:cat_ap}. We also use the following rule to determine the type of a problem:
\begin{itemize}
    \item \textbf{Network Revenue Management}: If the datasets contain information on \textit{demand}, \textit{price}, \textit{product name}, or similar information, this problem is classified into NRM.
    \item \textbf{Resource Allocation}: If the datasets provide data on \textit{resource capacity}, \textit{product value}, \textit{resource consumption}, or similar information, this problem is classified into RA.
    \item \textbf{Transportation Problem}: If the datasets show details about \textit{customer demand}, \textit{transportation costs}, \textit{supply capacity} or similar information, this problem is classified as TP.
    \item \textbf{Facility Location Problem}: If the dataset includes \textit{a list of potential facility locations with associated fixed opening costs}, \textit{customer demand quantities}, and \textit{transportation costs from facilities to customers}, this problem can be classified as FLP.
    \item \textbf{Assignment Problem}: If the dataset consists of a cost matrix where \textit{rows represent resources} (e.g., workers, machines) and \textit{columns represent tasks} (e.g., jobs, projects), and each cell indicates the \textit{cost of assigning a resource to a task}, this problem is classified as AP.
\end{itemize}

\begin{longtable}{p{4.5cm}p{1.5cm}p{8.5cm}}
\caption{Categorization of Network Revenue Management} \label{tab:categorization} \\
\toprule
\textbf{Category} & \textbf{Count} & \textbf{Description} \\
\midrule
\endfirsthead

\multicolumn{3}{l}{\textit{Continued from previous page}} \\
\toprule
\textbf{Category} & \textbf{Count} & \textbf{Description} \\
\midrule
\endhead

\multicolumn{3}{r}{\textit{Continued on next page}} \\
\bottomrule
\endfoot

\bottomrule
\endlastfoot

General Retail Scenarios & 18 & Includes supermarkets, restaurants, pizzerias, bakery shops, and dairy goods stores. \\
E-commerce Platforms & 4 & Covers online platforms like Amazon and other e-commerce sites managing furniture, clothing, etc. \\
High-Value or Entertainment Products & 3 & Focuses on high-value items, such as real estate properties, as well as entertainment-related products. \\
\end{longtable}

\begin{longtable}{p{4.5cm}p{1.5cm}p{8.5cm}}
\caption{Categorization of Resource Allocation} \label{table:cat_ra}\\
\toprule
\textbf{Category} & \textbf{Count} & \textbf{Description} \\
\midrule
\endfirsthead

\multicolumn{3}{l}{\textit{Continued from previous page}} \\
\toprule
\textbf{Category} & \textbf{Count} & \textbf{Description} \\
\midrule
\endhead

\multicolumn{3}{r}{\textit{Continued on next page}} \\
\bottomrule
\endfoot

\bottomrule
\endlastfoot
Retail Product Allocation & 12 & Allocates products (e.g., coffee, bread, books) to shelves or cabinets with space constraints. \\
Digital Game Allocation & 2 & Allocates game types (e.g., racing, sports) to platforms with memory limitations, maximizing value. \\
High-Value Product Display & 8 & Displays high-value products (e.g., boats, cars) to maximize value within space or capacity limits. \\
\end{longtable}

\begin{longtable}{p{4.5cm}p{1.5cm}p{8.5cm}}
\caption{Categorization of Transportation Problem} \label{table:cat_tp}\\
\toprule
\textbf{Category} & \textbf{Count} & \textbf{Description} \\
\midrule
\endfirsthead

\multicolumn{3}{l}{\textit{Continued from previous page}} \\
\toprule
\textbf{Category} & \textbf{Count} & \textbf{Description} \\
\midrule
\endhead

\multicolumn{3}{r}{\textit{Continued on next page}} \\
\bottomrule
\endfoot

\bottomrule
\endlastfoot
Retail Distribution & 4 & Distributes products from centers to customers or store locations (e.g., Amazon, Walmart). \\
Logistics and Supply Chain & 3 & Manages complex supply chains for efficient distribution of goods (e.g., logistics companies, retail chains). \\
Specialized Product Distribution & 2 & Distributes niche products (e.g., apparel, beverages) from centers to retail outlets (e.g., FashionHub, BrewCo). \\
\end{longtable}

\begin{longtable}{p{4.5cm}p{1.5cm}p{8.5cm}}
\caption{Categorization of Facility Location Problem} \label{table:cat_uflp} \\
\toprule
\textbf{Category} & \textbf{Count} & \textbf{Description} \\
\midrule
\endfirsthead

\multicolumn{3}{l}{\textit{Continued from previous page}} \\
\toprule
\textbf{Category} & \textbf{Count} & \textbf{Description} \\
\midrule
\endhead

\multicolumn{3}{r}{\textit{Continued on next page}} \\
\bottomrule
\endfoot

\bottomrule
\endlastfoot
Consumer Goods Supply & 7 & Focuses on distributing products like groceries, beverages, household goods, and CDs from suppliers to retail stores while minimizing total logistics and setup costs. \\
Retail Chain Inventory Planning & 3 & Includes multinational retailers like Walmart, where numerous stores require replenishment from upstream suppliers, involving fixed and transportation costs. \\
Motor Vehicle Distribution & 1 & Involves vehicle delivery from suppliers to dealerships (e.g., Colorado Motor Vehicle Sales) with fixed opening costs and distance-based delivery costs. \\
Manufacturing \& Service Network & 3 & Optimizes the placement of factories or service centers to satisfy market demand or customer assignments within capacity limits. \\
\end{longtable}

\begin{longtable}{p{4.5cm}p{1.5cm}p{8.5cm}}
\caption{Categorization of Assignment Problem} \label{table:cat_ap} \\
\toprule
\textbf{Category} & \textbf{Count} & \textbf{Description} \\
\midrule
\endfirsthead

\multicolumn{3}{l}{\textit{Continued from previous page}} \\
\toprule
\textbf{Category} & \textbf{Count} & \textbf{Description} \\
\midrule
\endhead

\multicolumn{3}{r}{\textit{Continued on next page}} \\
\bottomrule
\endfoot

\bottomrule
\endlastfoot

Construction Project Assignment & 4 & Involves assigning construction managers to various tasks, with the objective of ensuring a one-to-one match between projects and managers to minimize total cost. \\
Engineering Task Allocation & 1 &  Assigns technical engineers to specific projects, with each engineer's cost reflecting their task-specific expertise or efficiency. The objective is to achieve an optimal one-to-one matching that minimizes total assignment cost. \\
\end{longtable}

Furthermore, instances classified as Others and Mixture are constructed from real-world applications, textbook exercises, and meaningful problems in the literature, such as those studied in \citet{bonami2009exact, byrd1978application, eisenberg2001systemic, ernst2004staff,
franklin1973computed, hillier2005introduction, manne1958programming,
pisinger1998knapsack, pochet2006production, stigler1945cost,
waissi1994network, weingartner1916mathematical, wolsey1999integer, wood2014power}. The application domains for these two categories of instances are summarized in \Cref{table:cat_others} and \Cref{table:cat_mixture_detailed}.

\begin{longtable}{p{4.5cm}p{1.5cm}p{8.5cm}}
\caption{Categorization of Others Problems} \label{table:cat_others} \\
\toprule
\textbf{Category} & \textbf{Count} & \textbf{Description} \\
\midrule
\endfirsthead

\multicolumn{3}{l}{\textit{Continued from previous page}} \\
\toprule
\textbf{Category} & \textbf{Count} & \textbf{Description} \\
\midrule
\endhead

\multicolumn{3}{r}{\textit{Continued on next page}} \\
\bottomrule
\endfoot

\bottomrule
\endlastfoot

Workforce Scheduling & 2 & Focuses on determining the minimum number of personnel (e.g., bus drivers or restaurant waitstaff) required to meet varying demand over 24-hour periods. \\
Traveling Salesman Problem & 2 & Involves finding the shortest possible route for a vehicle or salesperson to visit a set of locations and return to the origin. \\
Knapsack Problem & 1 & Aims to select items with specific values and weights to maximize total value while staying within a fixed capacity limit. \\
Diet Problem & 1 & Optimizes food selection to minimize total cost while strictly adhering to multiple nutritional requirements like calories, protein, and vitamins. \\
Parallel-Machine Scheduling & 1 & Deals with assigning a set of tasks to multiple processors (CPUs) with different frequencies to optimize processing efficiency. \\
Flexible Flow Shop Scheduling & 1 & Manages the arrangement of production across multiple stages (rough/fine processing, packaging) to maximize total output value within workshop time limits. \\
\end{longtable}

\begin{longtable}{p{4.5cm}p{1.5cm}p{8.5cm}}
\caption{Categorization of Mixture Problems} \label{table:cat_mixture_detailed} \\
\toprule
\textbf{Category} & \textbf{Count} & \textbf{Description} \\
\midrule
\endfirsthead

\multicolumn{3}{l}{\textit{Continued from previous page}} \\
\toprule
\textbf{Category} & \textbf{Count} & \textbf{Description} \\
\midrule
\endhead

\multicolumn{3}{r}{\textit{Continued on next page}} \\
\bottomrule
\endfoot

\bottomrule
\endlastfoot

Multi-Period Production \& Inventory & 4 & Optimizes production levels across multiple time buckets to meet seasonal demand while balancing holding and backorder costs. \\
Integrated Supply Chain Logistics & 4 & Combines manufacturing site selection with distribution channel optimization to minimize total network costs. \\
Industrial Resource Blending & 3 & Focuses on the chemical or physical mixing of raw materials (e.g., oil, ores, chemicals) to achieve target output specifications at minimum cost. \\
Capacity Expansion \& Investment & 3 & Addresses strategic long-term decisions regarding facility scaling, equipment upgrades, and budget allocation for growth. \\
Production-Distribution Coupling & 2 & Simultaneously manages shop-floor production schedules and outbound transportation to ensure timely delivery to customers. \\
Process Industry Operational Planning & 2 & Involves complex multi-stage processing where raw material quality directly impacts final product yields and operational efficiency. \\
\end{longtable}
\subsection{Parameters Imputation}
\label{sec:AppendixB}
In this section, we provide more details on how we impute the missing parameters in our dataset. 

\vspace{2mm}
\noindent\textbf{Network Revenue Management:} The following code generates data for the NRM. It takes in sales data, generates product demand, and computes initial inventory for each product.

\begin{lstlisting}[language=Python, caption=Code for Generating Network Revenue Management Data]
import pandas as pd
import numpy as np
import math
import random

def generate_nrm_csv(input_csv: str, output_csv: str, seed_value: int = 42):
    random.seed(seed_value)
    np.random.seed(seed_value)
    df = pd.read_csv(input_csv)
    df["Product Name"] = df["Product id"]
    df["Revenue"] = df["Unit sellingPrice"]
    
    def calc_demand(u):
        k = random.uniform(1.2, 1.5)
        return math.ceil(u * k)
    
    df["Demand"] = df["Units sold"].apply(calc_demand)
    df["Initial Inventory"] = df["Units sold"] * 10
    df["Initial Inventory"] = df["Initial Inventory"].apply(round_up_to_multiple_of_10)
    
    df_out = df[["Product Name", "Revenue", "Demand", "Initial Inventory"]]
    df_out.to_csv(output_csv, index=False)
    print(output_csv, "generated.")

def round_up_to_multiple_of_10(x):
    return int(math.ceil(x / 10.0) * 10)

if __name__ == "__main__":
    generate_nrm_csv("source_data.csv", "NRM_example.csv", 42)
\end{lstlisting}

\begin{itemize}
    \item \texttt{generate\_nrm\_csv}: A function that reads an input CSV file, calculates demand based on the sales data, and generates a new CSV file with product names, revenue, demand, and initial inventory.
    \item \texttt{round\_up\_to\_multiple\_of\_10}: A helper function that rounds the inventory to the nearest multiple of 10.
    \item \texttt{df\_out}: A dataframe contains the final output data, including product name, revenue, demand, and initial inventory for each product.
\end{itemize}

This function is used to simulate the input data required for solving NRM. The generated data can be used for further optimization models in the NRM context.

\vspace{2mm}
\noindent\textbf{Transportation Problem:} The following code generates the transportation cost matrix for the TP. This matrix represents the transportation costs between suppliers and customers based on geographic distances.

\begin{lstlisting}[language=Python, caption=Code for Generating the Transportation Cost Matrix]
import pandas as pd
import random
from geopy.distance import geodesic

def GetDis3(df_raw, coordinates, cost_per_mile, seed_value=42):
    # Set the random seed for reproducibility
    random.seed(seed_value)
    np.random.seed(seed_value)

    # Shuffle and reset the index
    df_raw2 = df_raw.sample(frac=1).reset_index(drop=True)
    
    # Split data into suppliers and customers
    half = len(df_raw2) // 2
    suppliers = df_raw2[:half]
    customers = df_raw2[half:]

    # Create an empty DataFrame to store transportation costs
    df_tc = pd.DataFrame(index=suppliers.iloc[:, 0].values, columns=customers.iloc[:, 0].values)

    # Calculate transportation costs based on geographic distances
    for i in range(len(suppliers)):
        for j in range(len(customers)):
            distance = geodesic(coordinates[suppliers.iloc[i, 0]], coordinates[customers.iloc[j, 0]]).miles
            unit_cost = random.choice(cost_per_mile)
            df_tc.iloc[i, j] = distance * unit_cost
    return df_tc
\end{lstlisting}

\begin{itemize}
    \item \texttt{df\_raw}: A dataframe containing details of suppliers and customers, with their respective locations.
    \item \texttt{coordinates}: A dictionary mapping each region to its geographic coordinates (latitude and longitude).
    \item \texttt{cost\_per\_mile}: A list of possible transportation costs per mile.
    \item \texttt{df\_tc}: The resulting transportation cost matrix, where each entry represents the cost of transporting goods between a supplier and a customer.
\end{itemize}

This function calculates the transportation cost matrix by using geographic coordinates and a fixed transportation cost per mile. We also use the following code to generate the customer demand for each region by filtering data from the original dataframe.

\begin{lstlisting}[language=Python, caption=Code for Generating Customer Demand]
# Assuming the transportation cost matrix 'df_tc' is already generated
countries = df_tc.columns.to_list()

# Filter customer demand based on the regions in 'df_tc'
df_ctrydemand = df_raw[df_raw['region'].isin(countries)].set_index('region')
df_ctrydemand = df_ctrydemand.loc[df_tc.columns]
\end{lstlisting}

\vspace{2mm}
\noindent\textbf{Facility Location Problem:} The following code generates the facility cost matrix for the FLP, calculating the costs associated with opening and operating facilities.

\begin{lstlisting}[language=Python, caption=Code for Generating Facility Cost (FC)]
def get_capital_coordinates(df_raw):
    """
    This function converts the facility location data to geographic coordinates
    (latitude, longitude) for each facility or customer location.
    """
    coordinates = {}
    for _, row in df_raw.iterrows():
        coordinates[row['region']] = (row['latitude'], row['longitude'])
    return coordinates

# Assuming df_raw contains facility and customer data
coordinates = get_capital_coordinates(df_raw)

# Generate transportation cost matrix for FLP
df_tc = GetDis3(df_raw, coordinates, cost_per_mile)
\end{lstlisting}

\begin{itemize}
    \item \texttt{get\_capital\_coordinates}: A function that extracts the latitude and longitude of each facility or customer from the original data and returns a dictionary of coordinates.
    \item \texttt{df\_raw}: The raw input dataframe containing the facility and customer data (including regions and geographic locations).
    \item \texttt{df\_tc}: The transportation cost matrix generated for the FLP, based on the geographic distances between facilities and customers.
\end{itemize}

In FLP, facility setup costs are often inferred using a predefined distribution rule. The following code simulates the fixed setup costs for the facilities.

\begin{lstlisting}[language=Python, caption=Code for Simulating Facility Setup Costs]
def simulate_setup_costs(n, cost_range=(10000, 50000)):
    """
    Simulate fixed setup costs for facilities based on a predefined cost range.
    """
    return [random.randint(cost_range[0], cost_range[1]) for _ in range(n)]

# Simulate setup costs for each facility
facility_setup_costs = simulate_setup_costs(len(df_raw))
\end{lstlisting}

\begin{itemize}
    \item \texttt{simulate\_setup\_costs}: A function that generates random setup costs for facilities within a specified range.
    \item \texttt{facility\_setup\_costs}: A list of simulated setup costs for each facility.
\end{itemize}

These costs are used in the Facility Location Problem to minimize the total cost of opening and operating facilities.

\subsection{Problem Classification Rules}\label{appendix:ex_classification}
To systematically evaluate model performance, we have constructed a benchmark suite that classifies optimization problems into the following seven groups based on their mathematical formulation:

\begin{itemize}
    \item \textbf{Network Revenue Management}: Managing resources across a network to maximize revenue \citep{talluri2006theory}.
    \item \textbf{Resource Allocation}: Distributing and managing limited resources among competing activities to optimize an objective \citep{winston2004operations}.
    \item \textbf{Transportation Problem}: Minimizing the cost of shipping goods from a set of sources to destinations \citep{winston2004operations}.
    \item \textbf{Facility Location Problem}: Determining the optimal placement of facilities to minimize costs while meeting demands from clients \citep{daskin1997network}.
    \item \textbf{Assignment Problem}: Optimally assigning agents to tasks on a one-to-one basis \citep{winston2004operations}.
    \item \textbf{Others}: A collection of well-known archetypes not fitting the canonical models, e.g., the Diet, Blending, Knapsack, Traveling Salesperson, and various scheduling problems \citep{winston2004operations, pisinger1998knapsack, pinedo1992scheduling}.
    \item \textbf{Mixture}: A category for problems that are hybrids of standard types or possess unique, non-standard constraints.
\end{itemize}

\subsection{Sample Problem Instance in \testdata}
\label{appendix:sample_test}
In this section, we present a mixture problem in \testdata.
\begin{tcolorbox}[colback=white, colframe=boxagent2, coltitle=white, coltext=black,title=Problem Description,breakable]
\small
A precision manufacturing facility produces a broad portfolio of aerospace-grade widgets: Widget1 through Widget141. The product\_resources.csv file details the essential production parameters for each widget, specifying the labor hours required per unit, the amounts of Material A and Material B consumed per unit, and the base profit generated per unit sold. During the manufacturing process, Widget3 generates CatalystX as a valuable byproduct at a rate of 5 kg per unit produced. This CatalystX has significant market value and can be sold to pharmaceutical partners at \$300 per kilogram, though monthly sales are capped at 1500 kg due to market demand limitations. Any unsold CatalystX requires specialized hazardous-waste disposal costing \$200 per kilogram to comply with environmental regulations. Critical production constraints are defined in the resource\_limits.csv file: monthly labor hours cannot exceed 5,000 hours, Material A consumption is limited to 24,000 kg, and Material B usage is capped at 15,000 kg. The optimization challenge involves determining the ideal production quantities for each widget—together with the amount of CatalystX sold—to maximize total profits.
\end{tcolorbox}

\begin{tcolorbox}[colback=white, colframe=boxagent2, coltitle=white, coltext=black,title=Corresponding Datasets,breakable]
\small
\begin{longtable}{l c c c r}
\caption{Product Resources Data (product\_resources.csv)} \label{table:product_resources} \\
\toprule
\textbf{Product} & \textbf{LaborHours} & \textbf{MaterialA} & \textbf{MaterialB} & \textbf{Profit} \\
\midrule
\endfirsthead

\multicolumn{5}{l}{\textit{Continued from previous page}} \\
\toprule
\textbf{Product} & \textbf{Labor (hrs)} & \textbf{Material A (kg)} & \textbf{Material B (kg)} & \textbf{Profit (\$)} \\
\midrule
\endhead

\multicolumn{5}{r}{\textit{Continued on next page}} \\
\midrule
\endfoot

\bottomrule
\endlastfoot

Widget1  & 1.6 & 24 & 14 & 525 \\
Widget2  & 2.0 & 20 & 10 & 678 \\
Widget3  & 2.5 & 12 & 18 & 812 \\
Widget4  & 1.9 & 21 & 15 & 769 \\
Widget5  & 0.0 & 15 & 26 & 952 \\
Widget6  & 0.1 & 24 & 17 & 987 \\
Widget7  & 1.2 & 15 & 30 & 644 \\
Widget8  & 1.3 & 21 & 24 & 795 \\
Widget9  & 0.4 & 20 & 30 & 829 \\
Widget10 & 0.9 & 18 & 27 & 574 \\
\addlinespace[2mm]
\multicolumn{5}{c}{\vdots} \\
\addlinespace[2mm]
Widget141 & 1.2 & 11 & 16 & 593 \\
\bottomrule
\end{longtable}

\begin{longtable}{
    >{\raggedright\arraybackslash}p{4cm}  
    >{\raggedleft\arraybackslash}p{3cm}   
}
\caption{Resource Limits Data (resource\_limits.csv)} \label{table:resource_limits} \\
\toprule
\textbf{Resource} & \textbf{MonthlyLimit} \\
\midrule
\endfirsthead

\multicolumn{2}{l}{\textit{Continued from previous page}} \\
\toprule
\textbf{Resource Type} & \textbf{Monthly Availability} \\
\midrule
\endhead

\multicolumn{2}{r}{\textit{Continued on next page}} \\
\midrule
\endfoot

\bottomrule
\endlastfoot

Labor Hours & 5,000 \\
Material A (kg) & 24,000 \\
Material B (kg) & 15,000 \\
\bottomrule
\end{longtable}

\end{tcolorbox}

\begin{tcolorbox}[
    colback=white, 
    colframe=boxagent2, 
    coltitle=white, 
    coltext=black,
    title=Label (Ground-Truth Mathematical Model),
    breakable
]
\small

\subsection*{Decision Variables}
\begin{itemize}
    \item $x_i$: Units of Widget $i$ produced ($i = 1, 2, \dots, 141$)
    \item $s$: kg of CatalystX sold
\end{itemize}

\subsection*{Objective Function}
Maximize total profit:
\begin{align*}
\max Z = &\ 525x_1 + 678x_2 + (-188)x_3 + 769x_4 + 952x_5 + 987x_6 + 644x_7 + 795x_8 + 829x_9 + 574x_{10} \\
        &+ 723x_{11} + 775x_{12} + 992x_{13} + 568x_{14} + 895x_{15} + 838x_{16} + 883x_{17} + 567x_{18} + 829x_{19} + 821x_{20} \\
        &+ 893x_{21} + 977x_{22} + 536x_{23} + 797x_{24} + 793x_{25} + 754x_{26} + 819x_{27} + 947x_{28} + 857x_{29} + 886x_{30} \\
        &+ \cdots + 593x_{141} + 500s
\end{align*}

\subsection*{Constraints}

\subsubsection*{1. Labor Hours Constraint}
\[
\begin{aligned}
&1.6x_1 + 2x_2 + 2.5x_3 + 1.9x_4 + 0x_5 + 0.1x_6 + 1.2x_7 + 1.3x_8 + 0.4x_9 + 0.9x_{10} + 0.2x_{11} + 1.6x_{12} \\
&+ 0.1x_{13} + 0.7x_{14} + 1.1x_{15} + 0.7x_{16} + 1x_{17} + 1.8x_{18} + 0.5x_{19} + 1.2x_{20} + 0.2x_{21} + 1.5x_{22} \\
&+ 1.1x_{23} + 0.6x_{24} + 1.8x_{25} + 2x_{26} + 1.4x_{27} + 1.3x_{28} + 1.2x_{29} + 0.3x_{30} + 1.6x_{31} + 0.9x_{32} \\
&+ \cdots + 1.2x_{141} \leq 5000
\end{aligned}
\]

\subsubsection*{2. Material A Constraint}
\[
\begin{aligned}
&24x_1 + 20x_2 + 12x_3 + 21x_4 + 15x_5 + 24x_6 + 15x_7 + 21x_8 + 20x_9 + 18x_{10} + 24x_{11} + 20x_{12} \\
&+ 20x_{13} + 12x_{14} + 11x_{15} + 15x_{16} + 14x_{17} + 18x_{18} + 23x_{19} + 17x_{20} + 20x_{21} + 10x_{22} \\
&+ 12x_{23} + 12x_{24} + 10x_{25} + 15x_{26} + 17x_{27} + 22x_{28} + 11x_{29} + 13x_{30} + 17x_{31} + 24x_{32} \\
&+ \cdots + 11x_{141} \leq 24000
\end{aligned}
\]

\subsubsection*{3. Material B Constraint}
\[
\begin{aligned}
&14x_1 + 10x_2 + 18x_3 + 15x_4 + 26x_5 + 17x_6 + 30x_7 + 24x_8 + 30x_9 + 27x_{10} + 21x_{11} + 20x_{12} \\
&+ 22x_{13} + 19x_{14} + 28x_{15} + 25x_{16} + 27x_{17} + 15x_{18} + 25x_{19} + 29x_{20} + 17x_{21} + 22x_{22} \\
&+ 21x_{23} + 24x_{24} + 25x_{25} + 23x_{26} + 21x_{27} + 30x_{28} + 17x_{29} + 29x_{30} + 15x_{31} + 19x_{32} \\
&+ \cdots + 16x_{141} \leq 15000
\end{aligned}
\]

\subsubsection*{4. CatalystX Constraints}
\[
\begin{aligned}
& s \leq 1500 \\
& s \leq 5x_3
\end{aligned}
\]

\subsubsection*{5. Non-negativity Constraints}
\[
x_i \geq 0 \quad (i = 1, 2, \dots, 141), \quad s \geq 0
\]

\end{tcolorbox}

\begin{tcolorbox}[colback=white, colframe=boxagent2, coltitle=white, coltext=black,title=Label (Optimal Value),breakable]
\small
1344480.
\end{tcolorbox}

\section{On the Comparison between the Classification Agent and RAG}
\label{sec:AppendixR}
\begin{figure}[h]
    \centering
    \includegraphics[width=0.8\linewidth]{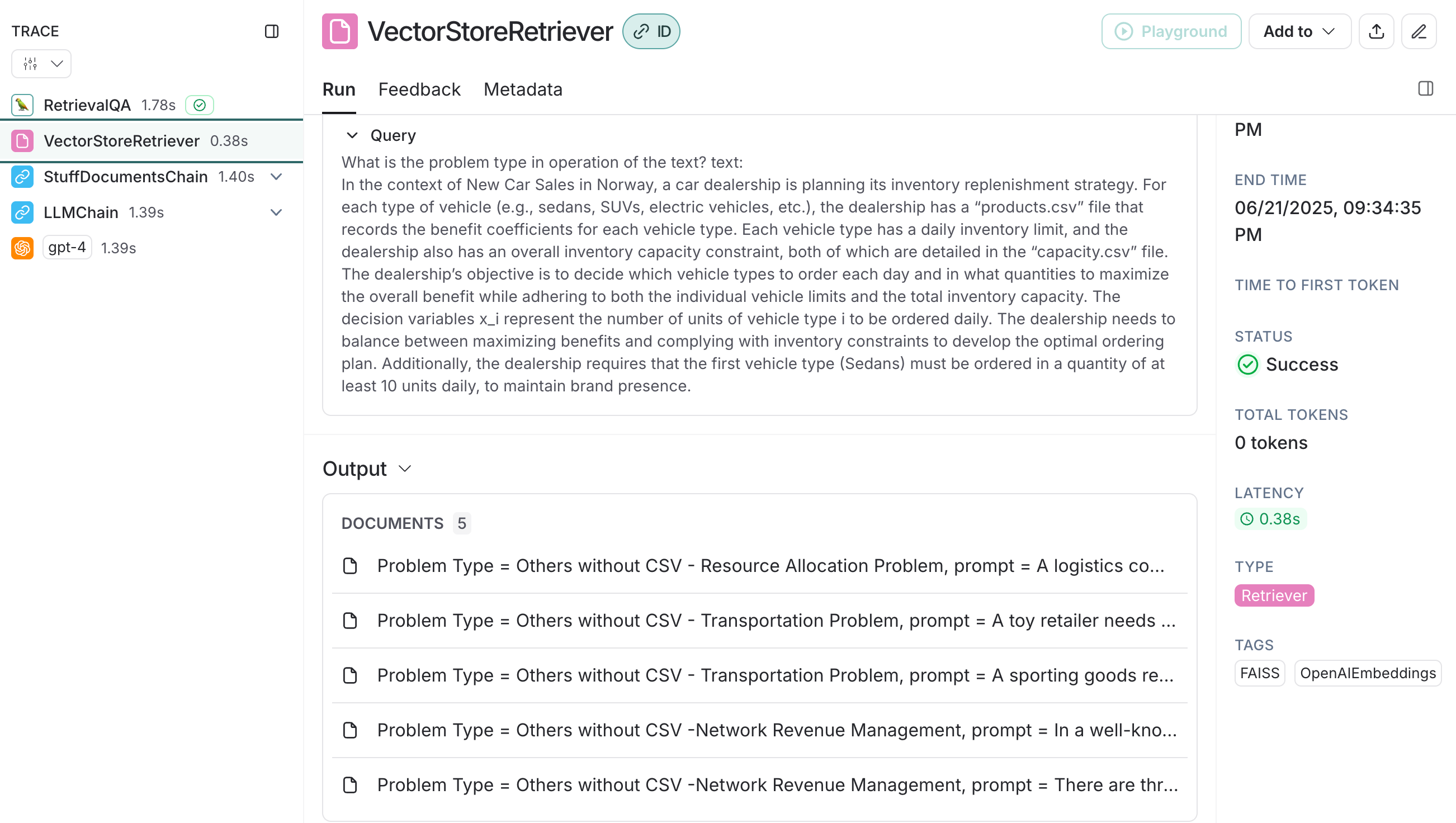}
    \caption{Top-5 problem types retrieved by the RAG framework}
    \label{rag_retrieve}
\end{figure}

In this section, we demonstrate the advantage of our problem classification agent over the simple RAG framework. Consider the query below, 
\begin{tcolorbox}[colback=white, colframe=boxgrey, coltitle=white, coltext=black,title=Query,breakable]
\small
 In the context of New Car Sales in Norway, a car dealership is planning its inventory replenishment strategy. For each type of vehicle (e.g., sedans, SUVs, electric vehicles, etc.), the dealership has a products.csv file that records the benefit coefficients for each vehicle type. Each vehicle type has a daily inventory limit, and the dealership also has an overall inventory capacity constraint, both of which are detailed in the capacity.csv file. The objective is to decide which vehicle types to order each day and in what quantities to maximize the overall benefit while adhering to both the individual vehicle limits and the total inventory capacity. The decision variables $x_i$ represent the number of units of vehicle type $i$ to be ordered daily. The dealership needs to balance between maximizing benefits and complying with inventory constraints to develop the optimal ordering plan.
\end{tcolorbox}
For this query, the retrieved similar problem instances from both the problem classification agent and the simple RAG framework are identical, as demonstrated in \ref{rag_retrieve}.

\begin{figure}[!h]
    \centering
    \includegraphics[width=0.8\linewidth]{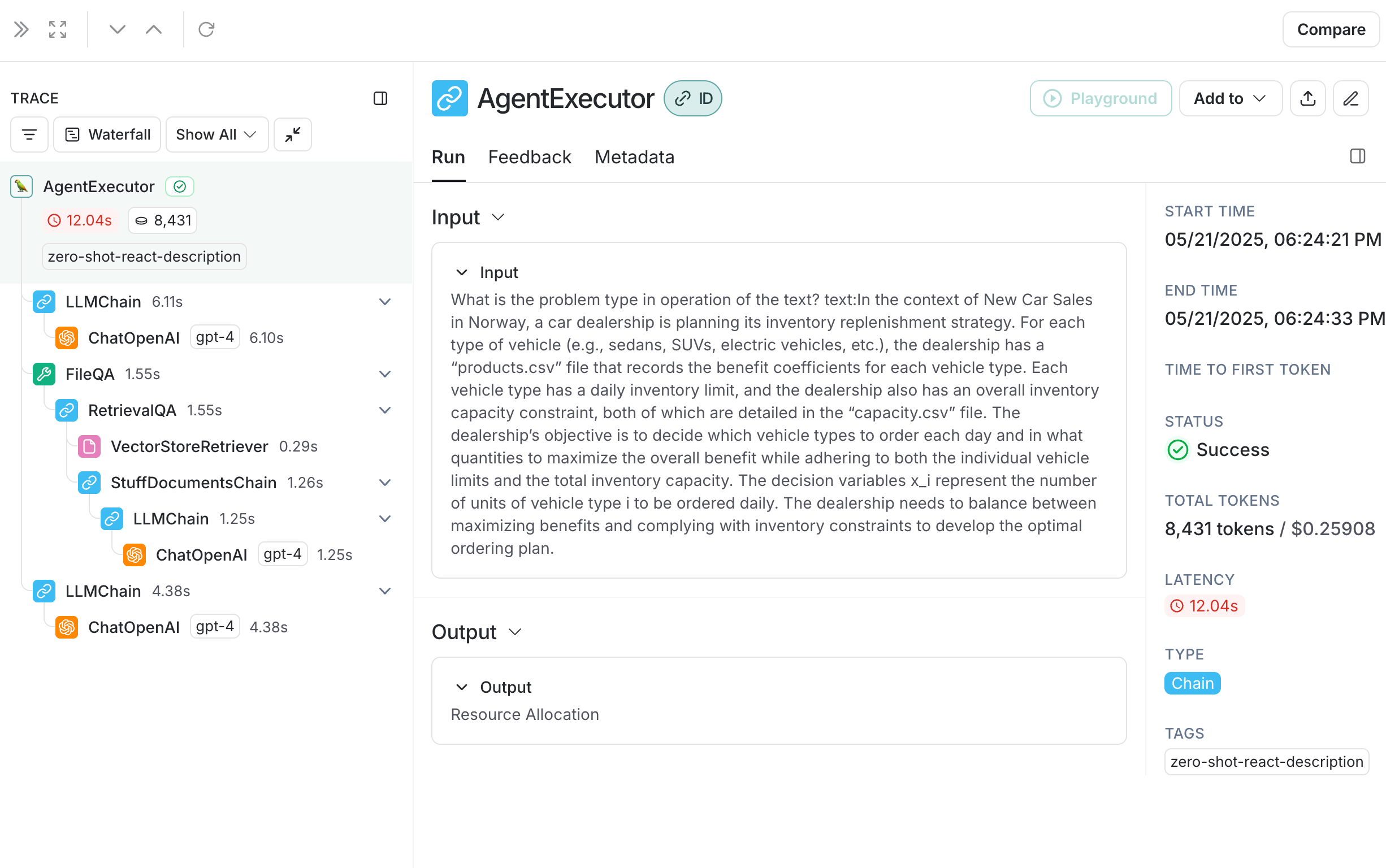}
    \caption{Identified problem type by the classification agent}
    \label{agent_answer}
\end{figure}

\begin{figure}[!h]
    \centering
    \includegraphics[width=0.8\linewidth]{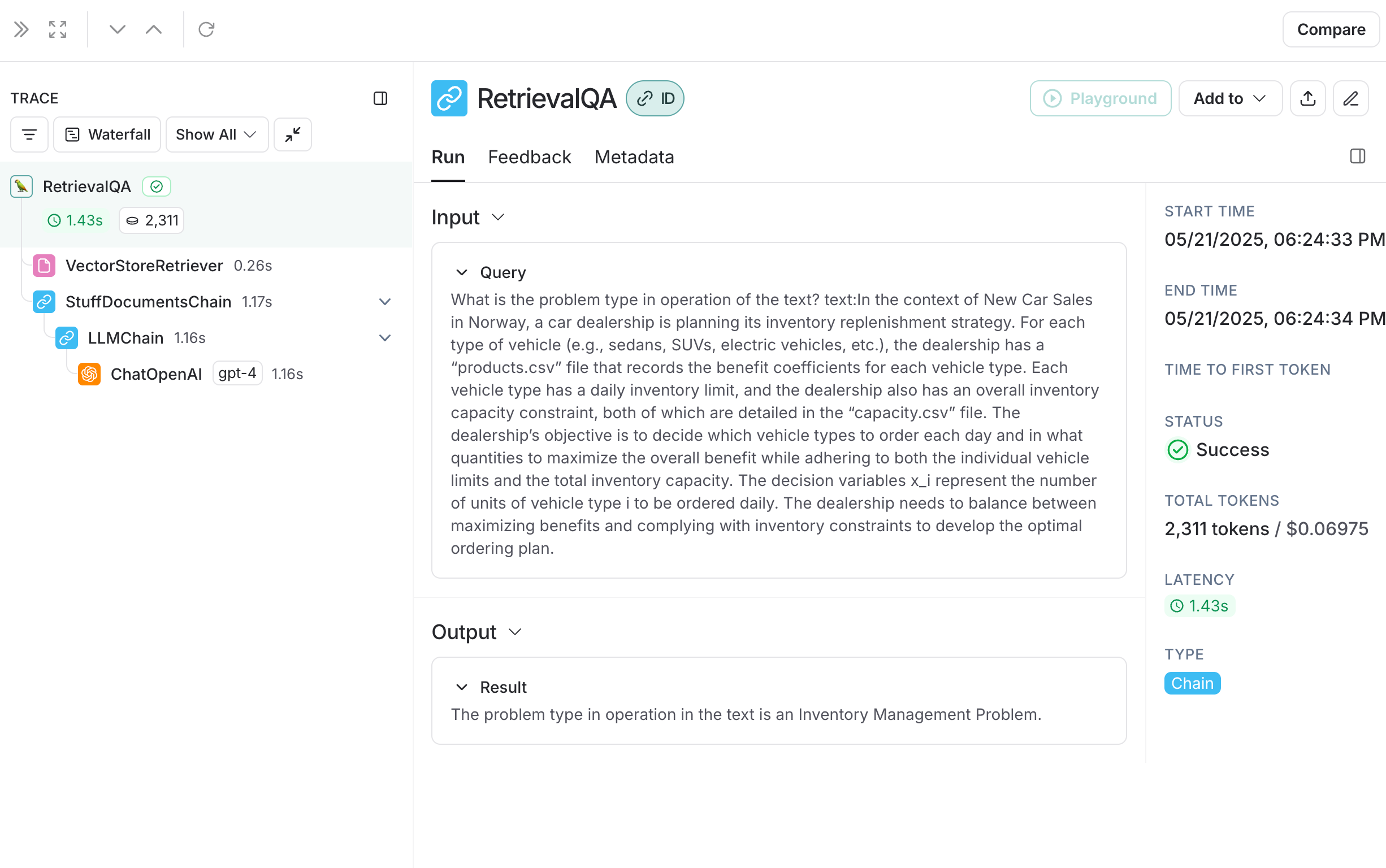}
    \caption{Identified problem type by the RAG framework}
    \label{rag_answer}
\end{figure}

However, the problem types returned by the problem classification agent and the RAG framework are \enquote{Resource Allocation} and \enquote{Inventory Management Problem} respectively, as shown in Figures \ref{agent_answer} and \ref{rag_answer}. Meanwhile, the correct problem type is \enquote{Resource Allocation}. The above example demonstrates the problem classification agent's ability to accurately identify the problem type and underscores the importance of the LLMs reasoning capability.

\section{Demo Problem Used in Workflow Generation}\label{appendix:demo}
We provide the demo problem instance used in generating the workflow presented in Section \ref{sec:type_workflow}. Below is the $(q,g,f,m)$ information of the instance in \refdata that is semantically and structurally closest to the input query.
\begin{tcolorbox}[colback=white, colframe=boxagent2, coltitle=white, coltext=black,title=Demo Query $q_{\demo}$,breakable]
\small
{\green \%\% The problem description of the most similar example in \refdata}\\
A supermarket needs to allocate various products, including high-demand items like the Sony Alpha Refrigerator, Sony Bravia XR, and Sony PlayStation 5, across different retail shelves. The product values and space requirements are provided in the \enquote{Products.csv} dataset. Additionally, the store has multiple shelves, each with a total space limit and specific space constraints for Sony and Apple products, as outlined in the \enquote{Capacity.csv} file. The goal is to determine the optimal number of units of each Sony product to place on each shelf to maximize total value while ensuring that the space used by Sony products on each shelf does not exceed the brand-specific limits. The decision variables $x_{ij}$ represent the number of units of product $i$ to be placed on shelf $j$.

\end{tcolorbox}

\begin{tcolorbox}[colback=white, colframe=boxagent2, coltitle=white, coltext=black,title=Demo Data Categories $g_{\demo}$,breakable]
\small
{\green \%\% The categories of data needed for model formulation}\\
 \small Sony.
\end{tcolorbox}

\begin{tcolorbox}[colback=white, colframe=boxagent2, coltitle=white, coltext=black,title=Retrieved Data for the Demo Problem $f_{\demo}$, breakable]
\small
{\green \%\% All related data from the dataset}\\
----------------DataFrame 1 - products.csv:---------------- \\
\small 1. Product: Sony Alpha Refrigerator, Value: 1818, SpaceRequirement: 400\\
\small 2. Product: Sony Bravia XR, Value: 1609, SpaceRequirement: 200\\
\small 3. Product: Sony HT-A7000 Soundbar, Value: 509, SpaceRequirement: 40\\
\small 4. Product: Sony PlayStation 5, Value: 1808, SpaceRequirement: 60\\
\small 5. Product:Sony PlayStation Soundbar, Value: 528, SpaceRequirement: 50\\
----------------DataFrame 2 - capacity.csv:----------------\\
\small 1. PlatformID: 1, Capacity: 1200\\
\small 2. PlatformID: 2, Capacity: 1374\\
\small 3. PlatformID: 3, Capacity: 800\\
\small 4. PlatformID: 4, Capacity: 2042\\
\small 5. PlatformID: 5, Capacity: 1000\\
\small 6. PlatformID: 6, Capacity: 1800
\end{tcolorbox}

\begin{tcolorbox}[colback=white, colframe=boxagent2, coltitle=white, coltext=black,title=Label of the Demo Problem $m_{\demo}$,breakable]
\small
{\green \%\% The target optimization model of the demo example} \\
\textbf{Objective Function:}

$\quad \quad \max \sum_i \sum_j p_i \cdot x_{ij}$

\textbf{ Constraints:}

1. Capacity Constraints: 

$\quad \quad \sum_i a_i x_{ij} \leq c_j, \quad \forall j$

2. Non-negativity Constraints: 

$\quad \quad  x_{ij} \geq 0, \quad \forall i,j $

\textbf{Retrieved Information:}

$\quad \quad p = [1818, 1609, 509, 1808, 528]$

$\quad \quad a = [400, 200, 40, 60, 50]$

$\quad \quad c = [1200, 1374, 800, 2042, 1000, 1800]$

\end{tcolorbox}

\section{Type-Tailored Retrieved Data Format}\label{appendix:data_format}
In this section, we present several examples for the retrieved data for one AP, one TP, and one FLP to complement the type-tailored workflow in Section \ref{sec:type_workflow}. The first is an AP where the data can be most efficiently stored as a matrix. 
\begin{tcolorbox}[colback=white, colframe=boxagent2, coltitle=white, coltext=black,title=Retrieved Data for AP,breakable]
\small
Here is all the data from cost.csv as provided:

\begin{equation*}
C = \left[ \begin{array}{@{}*{12}{c}@{}}
167.4 &  98.6 & 189.4 & 119.6 & 182.0 & 145.1 & 185.4 &  94.8 & 122.3 & 123.3 &  96.1 &  90.3 \\
156.2 &  88.7 & 187.3 & 124.7 & 173.2 & 144.3 & 179.0 &  91.5 & 115.1 & 119.5 & 100.1 &  88.6 \\
184.3 & 121.0 & 216.6 & 140.0 & 196.2 & 168.8 & 205.6 & 114.2 & 133.3 & 144.5 & 116.0 & 107.7 \\
157.9 &  92.9 & 185.1 & 120.3 & 175.1 & 146.2 & 180.8 &  86.3 & 111.6 & 115.9 &  98.1 &  91.1 \\
175.6 & 103.6 & 204.5 & 130.0 & 192.8 & 157.5 & 194.2 & 106.9 & 129.9 & 134.9 & 105.8 &  98.6 \\
166.8 & 107.0 & 199.2 & 130.4 & 183.6 & 159.5 & 187.0 &  98.2 & 121.3 & 126.2 & 105.9 & 101.8 \\
159.7 &  93.2 & 183.8 & 113.0 & 171.9 & 139.1 & 169.6 &  85.1 & 110.0 & 116.7 &  90.6 &  85.2 \\
184.8 & 115.9 & 205.1 & 138.6 & 195.4 & 160.1 & 200.2 & 108.5 & 136.9 & 140.0 & 114.6 & 103.9 \\
157.3 &  86.2 & 186.0 & 113.9 & 166.2 & 136.8 & 167.5 &  78.8 & 107.4 & 114.5 &  87.2 &  78.6 \\
164.8 &  97.8 & 200.9 & 125.8 & 188.9 & 151.2 & 187.7 &  99.5 & 119.5 & 132.1 & 101.1 &  98.4 \\
164.0 &  92.2 & 186.2 & 115.7 & 174.5 & 143.0 & 175.9 &  92.3 & 114.0 & 121.2 &  93.7 &  91.2 \\
151.7 &  76.7 & 179.5 & 109.5 & 160.6 & 128.4 & 170.2 &  74.4 & 103.7 & 110.4 &  83.7 &  75.2
\end{array} \right]
\end{equation*}
This is the complete cost matrix \( C \) with no simplification or abbreviation.
\end{tcolorbox}

The next example is the data format for TP where the demand an supply data are stored in a table and the cost data between demand and supply locations is stored in a matrix.
\begin{tcolorbox}[colback=white, colframe=boxagent2, coltitle=white, coltext=black,title=Retrieved Data for TP, breakable]
\small
\textbf{1. Supply Data for S1--S10}

\begin{table}[H]
\centering
\caption{Supply Data}
\begin{tabular}{lc}
\toprule
\textbf{Source ID} & \textbf{Supply Units} \\
\midrule
S1 & 103 \\
S2 & 87 \\
S3 & 95 \\
S4 & 112 \\
S5 & 97 \\
S6 & 103 \\
S7 & 101 \\
S8 & 94 \\
S9 & 102 \\
S10 & 106 \\
\bottomrule
\end{tabular}
\end{table}

\textbf{2. Demand Data for D1--D20}

\begin{table}[H]
\centering
\caption{Demand Data}
\begin{tabular}{lc}
\toprule
\textbf{Destination ID} & \textbf{Demand Units} \\
\midrule
D1 & 61 \\
D2 & 54 \\
D3 & 56 \\
D4 & 54 \\
D5 & 53 \\
D6 & 47 \\
D7 & 56 \\
D8 & 57 \\
D9 & 56 \\
D10 & 34 \\
D11 & 55 \\
D12 & 53 \\
D13 & 37 \\
D14 & 31 \\
D15 & 62 \\
D16 & 58 \\
D17 & 39 \\
D18 & 32 \\
D19 & 38 \\
D20 & 67 \\
\bottomrule
\end{tabular}
\end{table}

\textbf{3. Full Cost Matrix $C$}

\begin{table}[H]
\centering
\caption{Cost Matrix $C$ (Sources S1--S10 vs. Destinations D1--D20)}
\label{tab:cost_matrix}
\begin{adjustbox}{max width=\textwidth, center}
\begin{tabular}{l|cccccccccccccccccccc}
\toprule
\textbf{Source ID} & \textbf{D1} & \textbf{D2} & \textbf{D3} & \textbf{D4} & \textbf{D5} & \textbf{D6} & \textbf{D7} & \textbf{D8} & \textbf{D9} & \textbf{D10} & \textbf{D11} & \textbf{D12} & \textbf{D13} & \textbf{D14} & \textbf{D15} & \textbf{D16} & \textbf{D17} & \textbf{D18} & \textbf{D19} & \textbf{D20} \\
\midrule
S1  & 2.75 & 2.60 & 2.90 & 1.70 & 1.85 & 1.98 & 2.74 & 6.20 & 5.75 & 6.44 & 5.20 & 4.54 & 5.39 & 4.34 & 8.28 & 8.87 & 9.03 & 8.49 & 9.66 & 10.91 \\
S2  & 5.24 & 4.87 & 4.72 & 4.12 & 4.26 & 4.55 & 4.45 & 3.88 & 3.24 & 4.24 & 3.13 & 3.16 & 3.67 & 2.43 & 5.83 & 6.51 & 6.68 & 6.10 & 7.20 & 8.40 \\
S3  & 5.54 & 5.16 & 5.09 & 4.37 & 4.57 & 4.78 & 4.89 & 3.84 & 3.08 & 4.42 & 3.37 & 3.53 & 3.98 & 2.70 & 5.84 & 6.48 & 6.68 & 6.14 & 7.24 & 8.31 \\
S4  & 4.51 & 4.17 & 4.01 & 3.36 & 3.59 & 3.73 & 3.68 & 4.39 & 3.93 & 4.71 & 3.55 & 3.15 & 3.84 & 2.60 & 6.51 & 7.03 & 7.27 & 6.69 & 7.89 & 9.04 \\
S5  & 4.84 & 4.74 & 4.77 & 3.80 & 3.96 & 4.10 & 4.52 & 4.82 & 4.03 & 5.35 & 4.25 & 4.14 & 4.73 & 3.46 & 6.89 & 7.50 & 7.67 & 7.20 & 8.20 & 9.35 \\
S6  & 9.13 & 8.32 & 7.78 & 7.90 & 8.13 & 8.45 & 7.47 & 3.04 & 4.25 & 2.19 & 3.34 & 4.07 & 3.30 & 4.24 & 2.77 & 2.63 & 2.97 & 2.56 & 3.32 & 5.09 \\
S7  & 8.64 & 7.86 & 7.34 & 7.32 & 7.51 & 7.83 & 7.03 & 1.90 & 3.14 & 1.43 & 2.64 & 3.64 & 2.92 & 3.51 & 2.41 & 2.74 & 2.98 & 2.46 & 3.51 & 5.08 \\
S8  & 9.25 & 8.41 & 7.81 & 8.13 & 8.32 & 8.60 & 7.49 & 3.71 & 5.00 & 2.88 & 3.78 & 4.27 & 3.49 & 4.63 & 3.62 & 3.34 & 3.59 & 3.29 & 3.79 & 5.54 \\
S9  & 10.30 & 9.58 & 8.89 & 9.11 & 9.32 & 9.69 & 8.57 & 4.11 & 5.29 & 3.49 & 4.59 & 5.28 & 4.47 & 5.51 & 3.22 & 2.59 & 2.72 & 2.68 & 2.82 & 4.47 \\
S10 & 7.85 & 7.08 & 6.57 & 6.72 & 6.88 & 7.18 & 6.21 & 2.34 & 3.56 & 1.52 & 2.18 & 2.78 & 2.01 & 3.00 & 3.47 & 3.62 & 3.89 & 3.41 & 4.38 & 5.98 \\
\bottomrule
\end{tabular}
\end{adjustbox}
\end{table}
\end{tcolorbox}

Finally is an example of the retrieved data for FLP, where the warehouse operating cost and capacity and the store demand is stored in table. The transportation cost between a pair of warehouse and a store is formatted into a matrix.
\begin{tcolorbox}[colback=white, colframe=boxagent2, coltitle=white, coltext=black,title=Retrieved Data for FLP, breakable,top=3pt, bottom=3pt, before=\vspace{4pt}, after=\vspace{-1pt}]
\small
\textbf{1. Warehouse Data}

\begin{table}[H]
\centering
\caption{Warehouse Opening Costs and Capacities}
\begin{tabular}{lcc}
\toprule
\textbf{Warehouse (i)} & \textbf{Opening Cost ($f_i$)} & \textbf{Capacity (units)} \\
\midrule
1 & 3000 & 180 \\
2 & 3200 & 160 \\
3 & 3100 & 200 \\
4 & 2800 & 150 \\
5 & 3500 & 170 \\
6 & 2700 & 190 \\
7 & 2900 & 160 \\
8 & 3050 & 175 \\
9 & 3100 & 170 \\
10 & 2200 & 180 \\
11 & 2890 & 190 \\
\bottomrule
\end{tabular}
\end{table}

\textbf{2. Store Data}

\begin{table}[H]
\centering
\caption{Store Demands}
\begin{tabular}{lc}
\toprule
\textbf{Store (j)} & \textbf{Demand ($d_j$, units)} \\
\midrule
1 & 30 \\
2 & 40 \\
3 & 20 \\
4 & 35 \\
5 & 20 \\
6 & 25 \\
7 & 45 \\
8 & 38 \\
9 & 32 \\
10 & 41 \\
11 & 44 \\
\bottomrule
\end{tabular}
\end{table}

\textbf{3. Transportation Cost Matrix $A$}

\begin{table}[H]
\centering
\caption{Transportation Cost Matrix $A_{ij}$ (Warehouse $i$ to Store $j$)}
\label{tab:transportation_cost_matrix}
\small
\begin{tabular}{l|ccccccccccc}
\toprule
W/S & 1 & 2 & 3 & 4 & 5 & 6 & 7 & 8 & 9 & 10 & 11 \\
\midrule
1 & 12 & 17 & 13 & 18 & 10 & 15 & 14 & 19 & 17 & 14 & 15 \\
2 & 11 & 19 & 14 & 16 & 13 & 12 & 13 & 16 & 18 & 13 & 13 \\
3 & 14 & 15 & 12 & 17 & 12 & 14 & 15 & 18 & 12 & 15 & 16 \\
4 & 15 & 20 & 14 & 13 & 19 & 16 & 17 & 20 & 14 & 17 & 17 \\
5 & 17 & 18 & 16 & 18 & 15 & 13 & 12 & 17 & 16 & 16 & 11 \\
6 & 13 & 14 & 15 & 17 & 11 & 17 & 13 & 19 & 15 & 18 & 13 \\
7 & 12 & 17 & 11 & 14 & 12 & 16 & 14 & 16 & 14 & 14 & 14 \\
8 & 16 & 15 & 14 & 19 & 14 & 16 & 15 & 18 & 17 & 19 & 15 \\
9 & 16 & 13 & 16 & 16 & 12 & 14 & 12 & 15 & 21 & 15 & 19 \\
10 & 14 & 15 & 18 & 13 & 15 & 18 & 16 & 15 & 15 & 17 & 21 \\
11 & 15 & 16 & 17 & 18 & 17 & 19 & 14 & 18 & 18 & 19 & 13 \\
\bottomrule
\end{tabular}
\end{table}
\end{tcolorbox}
\section{Generate Python Program Codes from the Optimization Models}
\label{sec:AppendixC} 
Our framework enables one to translate the formulated optimization model to programming codes executable by solvers such as Gurobi to acquire the corresponding solution. We thus provide the Python code for this. We note that the details of code generation may vary depending on the problem type. To illustrate this, we present two representative code-generation examples below, corresponding to problems categorized as NRM and as Others, respectively.

\begin{lstlisting}[language=Python, caption=Programming Code Generation for NRM]
import openai
from langchain.schema import HumanMessage
from langchain_openai import OpenAIEmbeddings, ChatOpenAI

llm = ChatOpenAI(temperature=0.0, model_name="gpt-4.1", top_p=1, n = 1, openai_api_key=user_api_key)

prompt = f"""
You are an expert in mathematical optimization and Python programming. Your task is to write Python code to solve the provided mathematical optimization model using the Gurobi library. The code should include the definition of the objective function, constraints, and decision variables. Please don't add additional explanations. Below is the provided mathematical optimization model:

Mathematical Optimization Model:
{model}

For example, here is a simple instance for reference:

Mathematical Optimization Model:

Objective Function:
$\quad \quad \max \quad \sum_i A_i \cdot x_i$
Constraints
1. Inventory Constraints:
$\quad \quad x_i \leq I_i, \quad \forall i$
2. Demand Constraints:
$x_i \leq d_i, \quad \forall i$
3. Startup Constraint:
$\sum_i x_i \geq s$
Retrieved Information
$\small I = [7550, 6244]$
$\small A = [149, 389]$
$\small d = [15057, 12474]$
$\small s = 100$

The corresponding Python code for this instance is as follows:

import gurobipy as gp
from gurobipy import GRB

# Create the model
m = gp.Model("Product_Optimization")

# Decision variables for the number of units of each product
x_1 = m.addVar(vtype=GRB.INTEGER, name="x_1") # Number of units of product 1
x_2 = m.addVar(vtype=GRB.INTEGER, name="x_2") # Number of units of product 2

# Objective function: Maximize 149 x_1 + 389 x_2
m.setObjective(149 * x_1 + 389 * x_2, GRB.MAXIMIZE)

# Constraints
m.addConstr(x_1 <= 7550, name="inventory_constraint_1")
m.addConstr(x_2 <= 6244, name="inventory_constraint_2")
m.addConstr(x_1 <= 15057, name="demand_constraint_1")
m.addConstr(x_2 <= 12474, name="demand_constraint_2")

# Non-negativity constraints are implicitly handled by the integer constraints (x_1, x_2 >= 0)

# Solve the model
m.optimize()
"""

messages = [
    HumanMessage(content=prompt)  
]

response = llm(messages)

print(response.content)

file_name = "optimization_script.py"  
with open(file_name, "w") as file:
    file.write(response.content)

!python optimization_script.py
print(f"Code has been saved to {file_name}")
\end{lstlisting}

\begin{lstlisting}[language=Python, caption=Programming Code Generation for Others]
from langchain_openai import OpenAIEmbeddings, ChatOpenAI
from langchain_classic.chains import LLMChain
from langchain_core.prompts import PromptTemplate

llm = ChatOpenAI(
    temperature=0.0, model_name="gpt-4.1", ,top_p=1,n = 1,openai_api_key=user_api_key
    )
    
few_shot_block_code = build_few_shot_Other(Others_store, user_query, k=3, t='Code')
    
code_gen_template = """
You are an expert Gurobi programmer.
Your task is to strictly follow the User Query, CSV Schema, and "Abstract Model Plan" to translate them into a single, complete, executable Gurobi Python code block.
The code must start with ```python and end with ```.
The code must include all necessary imports (`gurobipy`, `pandas`, `numpy`, `sys`, `re`).
The code must include a `try...except` block to handle errors.
The code must robustly read '{dataset_address}' (if it's multiple paths, read the first).
The code must *fully* implement all variables, objectives, and constraints from the "Abstract Model Plan".

[Examples of the Full Process]
{few_shot_examples}

[User Query]
{query}

[CSV Schema]
{schema}

[Abstract Model Plan]
{abstract_plan}

[Your Gurobi Code]
"""

code_gen_prompt = PromptTemplate(template=code_gen_template, input_variables=["query", "schema", "abstract_plan", "dataset_address", "few_shot_examples"])
        
code_gen_chain = LLMChain(llm, prompt=code_gen_prompt)

final_code = code_gen_chain.run(
    query=user_query,
    schema=schema,
    abstract_plan=abstract_model_plan,
    dataset_address=dataset_address,
    few_shot_examples=few_shot_block_code 
)
\end{lstlisting}

\section{Example for Structurally Incorrect Model with a Correct Optimal Solution}\label{sec:em_obj}
In this section, we present several representative examples from \testdata where different models produce correct optimal solutions despite
formulating the underlying optimization problems incorrectly. Table~\ref{tab:by-family-transposed_em} summarizes the modeling accuracy (EM) of different models on \testdata. The cases collected in this section complement
the quantitative results by showing how structurally incorrect models may still return the same optimal solutions as the ground-truth formulations.

\begin{tcolorbox}[colback=white, colframe=boxagent2, coltitle=white, coltext=black,
title=Others (Generated by ORLM), breakable]
\small
\textbf{Query.}  
You are a nutritionist tasked with designing a one-day meal plan using various available foods to minimize the total cost of the menu while meeting specific nutritional guidelines. Per-serving nutrient and price information is provided in cost.csv, which lists Calories (kcal), Protein (g), Fat (g), Vitamin C (mg), and Cost (USD) for each food. Your plan must supply at least 2000 Calories, at least 50 g of protein, at least 60 mg of vitamin C, and no more than 70 g of fat. Portions may be fractional (e.g., 0.5 serving). Determine the optimal number of servings of each food that satisfies these requirements at the lowest possible cost, and clearly formulate the corresponding optimization model.

\begin{minipage}[t]{0.48\linewidth}
\textbf{Label (Ground-truth Mathematical Model).}

\textbf{Decision variables.}  

The labeled model introduces one continuous decision variable for \emph{every} food item
appearing in the dataset:
\[
x_i \ge 0, \quad \forall i \in \mathcal{F}.
\]

The full set $\mathcal{F}$ consists of multiple categories:
\begin{itemize}
\item Grains: 50 items
\item Dairy: 50 items
\item Meat: 50 items
\item Fruit: 50 items
\item Vegetables: 36 items
\item Staples (e.g., Oatmeal, Milk, Egg, Banana): 4 items
\end{itemize}
yielding a total of
\[
|\mathcal{F}| = 240 \text{ decision variables.}
\]

\textbf{Objective:}
\[
\min \sum_{i\in\mathcal{F}} c_i x_i
\]

\textbf{Constraints:}
\begin{align*}
&\sum_{i\in\mathcal{F}} a_i^{\text{cal}} x_i \ge 2000,\quad &\sum_{i\in\mathcal{F}} a_i^{\text{pro}} x_i \ge 50,\\
&\sum_{i\in\mathcal{F}} a_i^{\text{vitC}} x_i \ge 60,\quad
&\sum_{i\in\mathcal{F}} a_i^{\text{fat}} x_i \le 70.\\
\end{align*}
\end{minipage}
\hfill
\begin{minipage}[t]{0.48\linewidth}

\textbf{Model Generated by ORLM.}

\textbf{Decision variables.}  

The ORLM-generated formulation introduces decision variables for multiple food categories
present in \texttt{cost.csv}. However, for each category, only a restricted subset of items
is instantiated.

Concretely, the generated model includes:
\begin{itemize}
\item Grains: 25 items (out of 50),
\item Dairy: 25 items (out of 50),
\item Meat: 25 items (out of 50),
\item Fruit: 25 items (out of 50),
\item Vegetables: 25 items (out of 36),
\item Staples (Oatmeal, Milk, Egg, Banana): 4 items.
\end{itemize}

As a result, the effective decision set
\(
\widehat{\mathcal{F}} \subset \mathcal{F}
\)
satisfies
\[
|\widehat{\mathcal{F}}| \approx 129,
\]
which is substantially smaller than the full labeled model with
\(
|\mathcal{F}| = 240.
\)

\textbf{Objective and constraints.}  
The objective function and nutritional constraints are formulated in the same algebraic
form as the label model, but are evaluated only over the reduced set
\(
\widehat{\mathcal{F}}
\),
rather than the complete food universe.

\end{minipage}

\textbf{Key Difference.}  

The labeled formulation optimizes over the full food universe $\mathcal{F}$, whereas the ORLM
model restricts decisions to a much smaller subset $\widehat{\mathcal{F}} \subset \mathcal{F}$.
This dimensionality reduction fundamentally alters the feasible region and the optimization
problem itself.

Although the two models are not structurally equivalent, their optimal solutions may coincide under particular parameter realizations—for instance, when the omitted food items are not selected in any cost-minimizing feasible solution. Such agreement is instance-specific and does not imply correctness of the generated formulation.

\end{tcolorbox}

\begin{tcolorbox}[
  colback=white, colframe=boxagent2, coltitle=white, coltext=black,
  title=NRM (Generated by Gemini 3 Pro),
  breakable
]
\small

\textbf{Query.}
The supermarket offers a variety of products, with associated revenue data provided in the Revenue column. Each product has an independent demand stream. The retailer aims to maximize total expected revenue by focusing on the initial inventory of all the products. The initial inventory levels for the products are provided in the Initial Inventory column. During the sales horizon, no restocking is allowed, and there are no in-transit inventories. Throughout the sales horizon, customer inquiries for products arrive sequentially, following a Poisson process, with demand information specified in the Demand column. The decision variables, represented by $x_i$, indicate the number of customer requests the company plans to fulfill for each product $i$, with each $x_i$ being a positive integer.

\begin{minipage}[t]{0.52\linewidth}
\textbf{Label (Ground-truth Mathematical Model).}

\textbf{Decision variables.}
\[
x_i \in \mathbb{Z}_{\ge 0}, \quad i=1,\dots,109.
\]

\textbf{Objective:}
\[
\max \sum_{i=1}^{109} R_i x_i .
\]

\textbf{Constraints:}
\begin{align*}
& x_i \le I_i, \quad i=1,\dots,109,\\
& x_i \le D_i, \quad i=1,\dots,109.
\end{align*}

\textbf{Implied feasible region.}
\[
0 \le x_i \le \min\{I_i,D_i\}, \quad i=1,\dots,109.
\]

\end{minipage}
\hfill
\begin{minipage}[t]{0.45\linewidth}

\textbf{Model Implied by Gemini 3 Pro.}

\textbf{Decision variables.}
\[
x_i \in \mathbb{Z}_{\ge 0}, \quad i \in \mathcal{P},
\]
where $\mathcal{P}$ is the product list read from the CSV (one variable per row; in this instance it matches 109 products).

\textbf{Objective:}

Gemini 3 Pro replaces $\sum_i R_i x_i$ with a stochastic expected-sales objective:
\[
\max \sum_{i\in\mathcal{P}} R_i \cdot \mathbb{E}\!\left[\min\{x_i,\ \mathrm{Pois}(\mu_i)\}\right],
\]

\textbf{Constraints:}
\begin{align*}
&0 \le x_i \le I_i, \quad i\in\mathcal{P},\\
&\sum_{i\in\mathcal{P}} x_i \le C,
\end{align*}
with a hard-coded total capacity $C=150000$.

\end{minipage}

\textbf{Key Differences.}
\begin{itemize}
\item \textbf{Objective mismatch.} the label model is linear in $x_i$, while the generated code optimizes a Poisson expected-sales function using a piecewise-linear approximation.
\item \textbf{Missing demand upper bounds.} the label includes $(\mathrm{Dem})\ x_i \le D_i$ for every product, but the generated code does not enforce these per-product demand limits as constraints.
\item \textbf{Extra global capacity constraint.} the generated code adds $(\mathrm{Cap})\ \sum_i x_i \le C$ with $C$ fixed to 150000, which is not present in the label formulation.
\end{itemize}

The two formulations are not structurally equivalent. Nevertheless, their optimal solutions can coincide for certain datasets—for example, when the optimizer under the generated model happens to satisfy $x_i^\star \le D_i$ for all $i$ and the added capacity constraint is non-binding.

\end{tcolorbox}

\begin{tcolorbox}[colback=white, colframe=boxagent2, coltitle=white, coltext=black,
title= NRM (Generated by \agentabbrv(GPT-4.1)),breakable]
\small

\textbf{Query:}  
The mobile device retailer offers a portfolio of products with heterogeneous regional demand.
Each product $i$ yields revenue $R_i$ and has a fixed initial inventory level $I_i$, with no
restocking allowed during the sales horizon. Customer orders arrive stochastically, and the
retailer determines an integer fulfillment quantity $x_i$ for each product in order to maximize
total expected revenue. Demand constraints impose an upper bound $D_i$ on feasible fulfillment.
The decision problem is to optimally allocate limited inventory across products under both
inventory and demand limits.

\begin{minipage}[t]{0.48\linewidth}
\textbf{Label (Ground-Truth Mathematical Model).}

\textbf{Objective:}
\[
\max \sum_{i=1}^{71} R_i x_i
\]

\textbf{Constraints:}
\begin{align*}
& x_i \le I_i, \qquad & i = 1,\dots,71, \\
& x_i \le D_i, \qquad & i = 1,\dots,71, \\
&x_i \in \mathbb{Z}^{+}.
\end{align*}

\end{minipage}
\hfill
\begin{minipage}[t]{0.48\linewidth}
\textbf{Model Generated by \agentabbrv\ (GPT-4.1).}

\textbf{Objective:}
\[
\max \sum_{i=1}^{71} R_i x_i
\]

\textbf{Constraints:}
\begin{align*}
& x_i \le I_i, \qquad & i = 1,\dots,71, \\
& x_i \in \mathbb{Z}^{+}, \qquad & i = 1,\dots,71.
\end{align*}
\end{minipage}

\textbf{Key Difference.}  
The generated model omits the demand constraints $x_i \le D_i$, resulting in a strictly larger
feasible region. However, if the optimal solution $x^\star$ of the generated model satisfies
\[
x_i^\star \le D_i, \quad \forall i,
\]
then $x^\star$ also satisfies
\[
x_i^\star \le \min\{I_i, D_i\}.
\]
Therefore, although the numerical optimal solution coincides with that of the true formulation for this instance, the agreement is incidental. 
\end{tcolorbox}

\section{Example Problem Instance in \sindataca}\label{sec:case_supp}
An example query in \sindataca is shown below.
\begin{tcolorbox}[colback=white, colframe=boxgrey, coltitle=white, coltext=black,title=Query,breakable]
\small
\label{CA1}
Based on all flight ticket choices in \texttt{flight.csv} and \texttt{od\_demand.csv} with attraction values in v1 and shadow attraction value ratios in v2, develop the SBLP (sales-based linear programming) formulation for among flights (OD = (`A', `C') AND Departure Time=`23:00'), (OD = (`A', `C') AND Departure Time=`19:05'), (OD = (`B', `A') AND Departure Time=`15:40'), (OD = (`B', `A') AND Departure Time=`18:50'), (OD = (`C', `A') AND Departure Time=`16:55'), (OD = (`B', `A') AND Departure Time=`09:05'), (OD = (`C', `A') AND Departure Time=`07:40'),  that maximize the total revenue of flight ticket sales. The SBLP should include decision variables, objective function, balance constraints, scale constraints, nonnegative constraints and selection constraints. Please consider that each Eco-flexi ticket consumes 2 units of flight capacity and each Eco-lite ticket consumes 1 unit of capacity.
\end{tcolorbox}

\end{appendices}
\end{document}